\documentclass[10pt,twocolumn,letterpaper]{article}

\usepackage[pagenumbers]{cvpr} 

\usepackage{graphicx}
\usepackage{amsmath}
\usepackage{amssymb}
\usepackage{amsfonts}       
\usepackage{tabularx}
\usepackage{multicol}
\usepackage{multirow}
\usepackage{wrapfig}
\usepackage{siunitx}
\usepackage{xcolor}
\usepackage{bm}
\newcommand{\KG}{s}
\newcommand{\MC}{\kappa}
\newcommand{\QC}{Q}
\definecolor{best}{RGB}{10, 150, 10}
\definecolor{close}{RGB}{255, 140, 0}
\DeclareMathAlphabet{\mathsfit}{\encodingdefault}{\sfdefault}{m}{sl}
\SetMathAlphabet{\mathsfit}{bold}{\encodingdefault}{\sfdefault}{bx}{n}

\newtheorem{theorem}{Theorem}
\newtheorem{proof}{Proof}
\newtheorem{proposition}{Proposition}
\definecolor{mypink1}{rgb}{0.858, 0.188, 0.478}
\definecolor{myblue}{rgb}{0.2, 0, 1}
\newcommand{\AlgName}{RMSGD}
\DeclareCaptionLabelFormat{andtable1}{Table 1 \& Table 2}
\DeclareCaptionLabelFormat{andtable}{Table 3 \& Table 4}

\usepackage{amsmath,amsfonts,bm}

\newcommand{\trace}[1]{\text{Tr}\left\{#1\right\}}

\newcommand{\Parameters}{\ensuremath{\bm w}}

\newcommand{\MomentumRate}{\ensuremath{\alpha}}
\newcommand{\GlobalLR}{\ensuremath{\eta}}

\def\eqref#1{equation~\ref{#1}}

\def\1{\bm{1}}

\DeclareMathAlphabet{\mathsfit}{\encodingdefault}{\sfdefault}{m}{sl}
\SetMathAlphabet{\mathsfit}{bold}{\encodingdefault}{\sfdefault}{bx}{n}
\newcommand{\tens}[1]{\bm{\mathsfit{#1}}}

\def\tE{{\tens{E}}}

\def\tW{{\tens{W}}}

\def\sR{{\mathbb{R}}}

\usepackage[ruled, vlined]{algorithm2e}
\usepackage{algorithmic}
\usepackage{url}            
\usepackage{booktabs}       
\usepackage{caption}
\usepackage{nicefrac}       
\usepackage{microtype}      

\usepackage[pagebackref,breaklinks,colorlinks]{hyperref}

\usepackage[capitalize]{cleveref}
\crefname{section}{Sec.}{Secs.}
\Crefname{section}{Section}{Sections}
\Crefname{table}{Table}{Tables}
\crefname{table}{Tab.}{Tabs.}

\def\cvprPaperID{00778} 
\def\confName{CVPR}
\def\confYear{2022}

\begin{document}

\title{Exploiting Explainable Metrics for Augmented SGD}

\author{Mahdi S. Hosseini$^{1}$\thanks{Equally major contribution}~~~~~~ Mathieu Tuli$^{2}$\footnotemark[1]~~~~~~Konstantinos N. Plataniotis$^2$\\
$^1$University of New Brunswick~~~~$^2$University of Toronto\\
{\small\tt{Code:}\color{purple}\url{https://github.com/mahdihosseini/RMSGD}
}}
\maketitle

\begin{abstract}
Explaining the generalization characteristics of deep learning is an emerging topic in advanced machine learning. There are several unanswered questions about how learning under stochastic optimization really works and why certain strategies are better than others. In this paper, we address the following question: \textit{can we probe intermediate layers of a deep neural network to identify and quantify the learning quality of each layer?} With this question in mind, we propose new explainability metrics that measure the redundant information in a network's layers using a low-rank factorization framework and quantify a complexity measure that is highly correlated with the generalization performance of a given optimizer, network, and dataset. We subsequently exploit these metrics to augment the Stochastic Gradient Descent (SGD) optimizer by adaptively adjusting the learning rate in each layer to improve in generalization performance. Our augmented SGD -- dubbed RMSGD -- introduces minimal computational overhead compared to SOTA methods and outperforms them by exhibiting strong generalization characteristics across application, architecture, and dataset.
\end{abstract}

\section{Introduction}\label{sec:introduction}
The task of predicting network generalization performance using some measure of complexity based on training data is an emerging topic in the field of machine learning. Development of such ``explainability'' metrics is vitally important in order to understand and better explain the learning mechanisms involved in training of a given optimizer, network, and dataset. Identifying the causal relationship between some metric and generalization gap (or even testing accuracy directly) in order to select optimal network topologies or tune hyper-parameters is an important problem and actively researched today \cite{NIPS2017_10ce03a1, jiang2018predicting, Liu2020Understanding, Jiang2020Fantastic, NEURIPS2020_86d7c8a0}.  

While the field of metric development for predicting generalization performance is growing (see related works in \autoref{sec:related_works}), our interest in this work is to exploit such explainability metrics to augment the training of deep neural networks (DNNS). We achieve this by defining new metrics -- \textit{stable rank, condition number, and a quality measure} -- derived from the intermediate layers of DNNS during training to access the true knowledge in the underlying weights and express how well the network layers are functioning as high-quality autoencoders for knowledge representation. Using these metrics, we exploit their ability to quantify learning in order to augment stochastic gradient descent (SGD) by dynamically adjusting the learning rate. 

From a different lens, this work also sheds light on the behaviour of commonly practiced hyper-parameter tuning techniques like learning rate scheduling through decay methods \cite{goodfellow2016deep, loshchilov2016sgdr, luo2018adaptive, zhuang2020adabelief, heo2021adamp, foret2021sharpnessaware} or functional methods \cite{loshchilov2016sgdr, smith2017cyclical, smith2019super, he2019bag}. Little is understood about such methods and why they really work and their use becomes more like \textit{alchemy} rather than analytical/empirical reasoning. We highlight how our metrics provide reasonable explanation to these strategies and also how they can be used in a simple optimization framework to augment SGD and gain performance. Our main contributions are as follows:

\textbf{[C1]} We introduce new explainability metrics derived using training data that quantify layer-level learning of neural networks and are robust to the randomness of initialization.

\textbf{[C2]} We use these metrics to explain the training mechanisms of neural networks for various optimizers and datasets, and predict generalization characteristics in deep learning.

\textbf{[C3]} We exploit these metrics to augment SGD and introduce our new RMSGD optimizer, which gains considerable performance improvements at minimal computational cost and generalizes well across experimental configuration.
\section{Related Works}\label{sec:related_works}
A variety of complexity measures have been recently introduced for predicting network generalization such as $\ell_p$-norm signatures from network weights in \cite{NIPS2017_10ce03a1}, margin distribution by measuring the distance between network training and decision bounds \cite{jiang2018predicting}, and a gradient signal-to-noise-ratio (GSNR) measure from the evolution of training weights \cite{Liu2020Understanding}. A more comprehensive analysis of related measures and exploring their dependencies to a variety of topological structures and datasets can be found in \cite{Jiang2020Fantastic, NEURIPS2020_86d7c8a0}. A few other works also take further steps to train an estimator on pairs of signature and test accuracy using a regression model or fully-connected layers for high accuracy prediction of generalization gap \cite{yak2019towards, corneanu2020computing} and test accuracy \cite{DBLP:journals/corr/abs-2002-11448}. A variant of such studies also utilizes a neural complexity measure as an additional regularizer to the loss function to accelerate training and attempts at improving the generalization gap \cite{NEURIPS2020_6e17a5fd}. While the above-mentioned metrics show strength in predicting the network generalization, they are either (a) defined as a function of overall network performance that cannot be probed in intermediate layers of deep network for performance measurement; or (b) are sensitive to the noise perturbations of the weight structures which can potentially lead to inconsistent behaviours as well as low correlation accuracy measures. Our proposed methods do not suffer from these drawbacks.

The field of stochastic optimization has grown considerably for training DNNS. Families of stochastic gradient descent (SGD) based optimizers have been introduced in \cite{polyakmomentum1964, bottou2010large, bottou2012stochastic, sutskever2013importance, yuan2016influence, loizou2020momentum}. More advanced methods to increase the generalization of SGD are also studied in \cite{foret2021sharpnessaware, heo2021adamp}. Despite good generalization characteristics of the SGD based optimizers, tuning their associated hyper-parameters (such as learning rate) are the main bottlenecks to their use in practice. A variety of adaptive optimization algorithms are also introduced to leverage an adaptive stochastic minimization framework such as Adam \cite{kingma2014adam}, AdaBound \cite{luo2018adaptive} and AdamP \cite{heo2021adamp}. While adaptive based optimizers are shown to work well across different applications such as computer vision (CV) and natural language processing, they generalize poorly to test data in CV applications \cite{wilson2017marginal}. Improvements are made in \cite{dozat2016, Liu2020On, luo2018adaptive, gunes2018online, vaswani2019painless} to overcome this issue however, they still lack in performance when compared to SGD-based methods. Our proposed work builds on top of SGD and inherits its performance.
\section{On Explainability Metrics}\label{sec:metric}
\subsection{Low-Rank Factorization of Matrix Weights}
We argue that it is useful to decompose the weight matrices of the network being studied by low-rank factorization. This will allow us to analyze the underlying information that is learned during training. We first note that the weight matrices to be decomposed can be obtained in one of two forms: In the first form, if the weight matrix is part of a convolution layer (i.e. it is $4$-dimensional), similar to \cite{lebedev2015speeding}, it can be unfolded as ${\tW}_{\text{4D}}\in{\sR}^{{h}\times{w}\times{n_i}\times{n_o}}\xrightarrow{\text{unfold}}{\tW}\in{\sR}^{m\times{n}}$, where $w,h$ are the width and height of the convolution kernel and $n_i,n_o$ correspond to the number of input and output channels, respectively. We unfold the tensor either on mode-3 (input channel) as ${\tW}\in{\sR}^{whn_o\times{n_i}}$ or mode-4 (output channel) as ${\tW}\in{\sR}^{whn_i\times{n_o}}$. In the second form, if the layer is a simple linear layer, we can directly utilize the $2$D tensor $\tW\in{\sR}^{m\times{n}}$ of linear layers weights (e.g. fully-connected layer, Transformer weights, etc). Note in the case of linear layers, bias would be ignored. In both forms we assume $n \leq m$. We then obtain the low-rank structure by factorizing
\begin{equation}
\tW_{\ell}\texttt{[Weight]} \xrightarrow{\text{\tiny fac.}}\widehat{\tW}_{\ell}\texttt{[Low-Rank]} + \tE_{\ell}\texttt{[Noise]}
\label{eq:lowrank_factorization}
\end{equation}
where, $\widehat{\tW}_{\ell}$ is the low-rank matrix containing limited non-zero singular values i.e. $\widehat{\tW}_{\ell} = U \Lambda V^{\top}$, where $\Lambda=\text{diag}\{\sigma_1, \sigma_2, \hdots \sigma_{n^{\prime}}\}$ and $n^{\prime} = \text{rank}~{\widehat{\tW}_{\ell}}$. Here, $n^{\prime}<\min(m,n)$ due to the low-rank property. For our experiments, we employ the Variational Baysian Matrix Factorization (VBMF) method \cite{nakajima2013global} to perform the low-rank factorization. This method provides a global analytical solution and avoids an iterative algorithm by solving a quadratic minimization problem, meaning it is computationally efficient and can be easily applied to multiple layers of arbitrary size with minimal overhead (see \autoref{fig:curves} for example).

\begin{figure}[htp]
    \centering
    \begin{subfigure}{\linewidth}
    \includegraphics[width=0.96\textwidth]{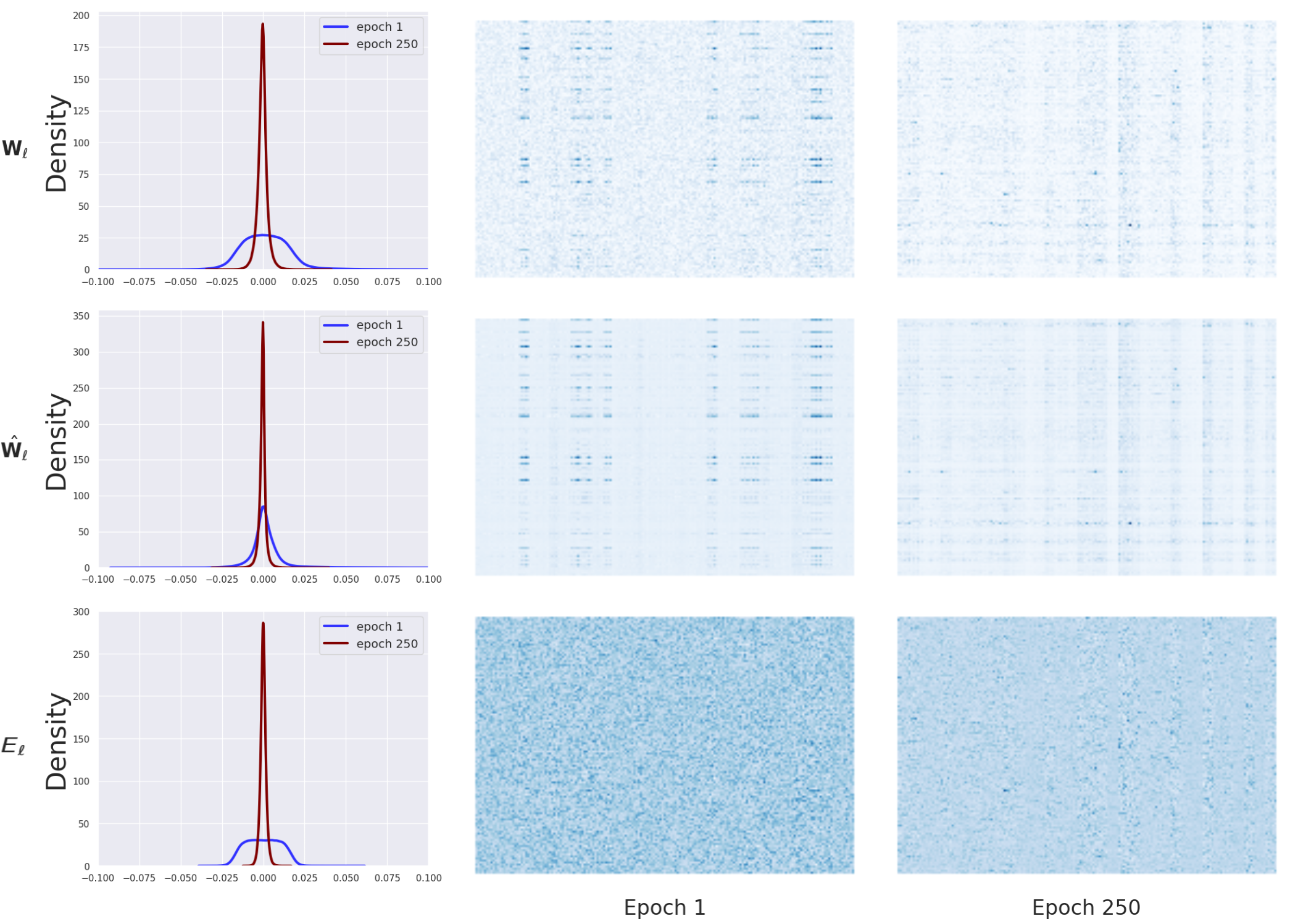}
    \caption{Low-Rank decomposition on layer $\ell=25$ from ResNet34/CIFAR10}
    \label{fig:lowrank_concept}
    \end{subfigure}
    \begin{subfigure}{\linewidth}
    \includegraphics[width=0.96\textwidth]{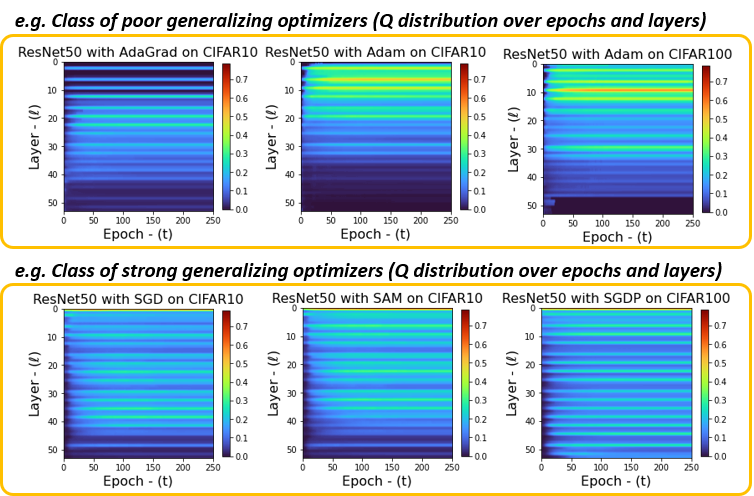}
    \caption{Quality (Q) Measure}
    \label{fig:quality_concept}
    \end{subfigure}
    \caption{(a) Implication of low-rank factorization on weight matrix taken from a particular layer of ResNet34; (b) Quality measure ($\QC$) on ResNet50 trained with different optimizers.}
    \label{fig:lowrank_Q_figures}
\end{figure}

The use of VBMF low-rank factorization allows our analysis to be robust to the randomness introduced by different initialization methods employed in training. This factorization is essential, and using techniques such as SVD directly would be heavily influenced by the presence of noise and prevent proper analysis. In the context of our application to augmenting SGD, performance would be seriously degraded.

Initially, the low-rank component of the weight matrix has an empty structure (i.e. $\widehat{\tW}_{\ell}=\oslash$) as the randomness of the initialized weights is fully captured in the noise perturbing component $E$. As training progresses, the low-rank component becomes non-empty and start learning to develop meaningful mapping structure. \autoref{fig:lowrank_concept} demonstrates an intuitive example of this development from a ResNet34 layer. Notice how the low-rank structure is preserved during epoch training while the noise fades away. This highlights how training reduces the perturbing noise within the layers of a neural network and strengthens the useful information embedded in the low-rank structure. We state that this leads to a stabilized encoding layer.

\subsection{Probing Metrics}\label{sec:probing_metric}
Given this notion of low-rank factorization, we wish to now quantify how well a network layer encodes and propagates information. To this aim, we borrow two metrics from linear algebra matrix analysis: \textit{stable rank}  \cite{rudelson2007sampling, cohen2016optimal} and \textit{condition number} \cite{horn2012matrix}. We apply these concepts to the low-rank factorized weight matrices discussed earlier.

The stable rank is the  norm energy of the singular values of a given matrix. We propose a modified definition of the stable rank on the low-rank structure $\widehat{\tW}_{\ell}\in{\sR}^{m\times{n}}$ as
\begin{align}
    \KG(\widehat{\tW}_{\ell}) = \frac{1}{n}\frac{\parallel \widehat{\tW}_{\ell}\parallel_*}{\parallel \widehat{\tW}_{\ell}\parallel_2}=
    \frac{1}{n\sigma^2_1(\widehat{\tW}_{\ell})}\sum^{n^{\prime}}_{i=1}\sigma^2_i(\widehat{\tW}_{\ell}),
    \label{eq:stable_rank}
\end{align}
where, $\sigma_1\geq\sigma_2\geq\cdots\geq\sigma_{n^{\prime}}$ are the low-rank singular values in descending order and $\parallel \cdot\parallel_*$ stands for nuclear norm (also known as the Schatten norm). This metric encodes the significance of low-rank span in the output (feature) mapping space. A higher measure indicates a better encoder and stronger carriage of information through the layer's weight matrix. Note that we normalize the stable rank by the smaller input dimension $n$ of the input matrix (since we assume $n\leq m$) to bound $\KG(\widehat{\tW}_{\ell})\in[0,1]$.

The condition number is also defined as a relative ratio of the highest and lowest singular values. We modify this definition on the low-rank structure $\widehat{\tW}_{\ell}\in{\sR}^{m\times{n}}$ as
\begin{align}
    \MC(\widehat{\tW}_{\ell}) = 1 - {\sigma_{n^{\prime}}(\widehat{\tW}_{\ell})}/{\sigma_{1}(\widehat{\tW}_{\ell})}.
    \label{eq:lowrank_condition}
\end{align}
Note that $\MC(\widehat{\tW}_{\ell})\in[0,1]$. This metric indicates the numerical sensitivity of the weight matrix's mapping with respect to input noise perturbations. Lower condition indicates higher robustness to noise and better input-output mapping. A concept example is shown in \autoref{fig:figure_Optimizer_behaviour} for the evolution of these metrics across training epochs of a layer in ResNet18. Note also that our metric development does not consider components such as skip connections. We argue intuitively that the influence of components such as skip connections gets captured in the weight matrices of nearby layers as it learns, and by thus analyzing the matrices themselves is a sufficient task, which we show holds in our experiments. 

\subsection{On the Meaning of Probing Metrics}
\label{sec:optimal_metrics}
Given the definitions of stable rank in \autoref{eq:stable_rank} and condition number in \autoref{eq:lowrank_condition}, we argue that a stable rank of $1$ and condition number of $0$ indicate a perfectly learned network. Specifically, for the stable-rank, higher value $\KG(\widehat{\tW}_{\ell}) \rightarrow 1$ indicates that most singular values are non-zero (i.e. $\sigma_i^2(\widehat{\tW}_{\ell}) > 0 \forall i \in [1, \hdots, n']$ where $n' \rightarrow n$. This creates a subspace spanned by a set of independent vectors corresponding to the non-zero singular values mentioned above. In other words, $\KG(\widehat{\tW}_{\ell}) \rightarrow 1$ corresponds to a many-to-many mapping but not a many-to-low (i.e. rank-deficient) mapping. Also, note that the stable rank is measured on the low-rank and not the raw measure of the weights. So the higher value indicates that the \textit{learned} weight matrix contains more non-empty structure which can be interpreted as a sign of meaningful learning.

For the condition number, we note that this metric also defines the numerical sensitivity of the inverse matrix $\widehat{\tW}_{\ell}$ toward minor input pertubations. Note that the error residual reconstruction under the linear system $y = \widehat{\tW}_{\ell}x$ will be bounded by $\parallel x - \hat{x} \parallel / \parallel x \parallel < c \sigma_1(\widehat{\tW}_{\ell}) / \sigma_{n'}(\widehat{\tW}_{\ell})$, where $x$ and $\hat{x}$ are the underlying and recovered signals, respectively, and $c$ is some constant measure from the linear system. We defer to \cite{horn2012matrix} for more information. Thus, the mapping condition has a direct impact on the residual reconstruction. This is specifically important during DNN training, where gradients are back-propagated and the matrix weights are involved in adjoint form for parameters updates. If the condition number is low (i.e. $\MC(\widehat{\tW}_{\ell}) \rightarrow 0$) then the noise perturbations from gradients will be accumulated during the iterative training phase and accordingly it yields a poor learned weight matrix for space mapping (i.e. the encoding).

Given this understanding, we the wish to incorporate both metrics into a single probing measure to quantify quality of learning. We tackle this in the following section.

\subsection{Quality Measure on Learned Network}
\begin{figure*}[!t]
    \centering
    \begin{subfigure}{0.78\textwidth}
    \centering
    \includegraphics[width=0.93\textwidth]{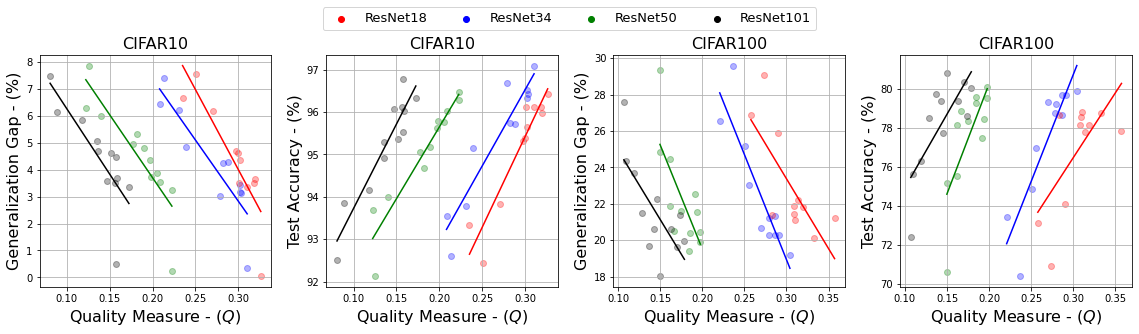}
    \scriptsize
        \begin{tabular}{c|c|c|c|c||c|c|c|c}

        \toprule
            Network&\multicolumn{4}{c||}{CIFAR10}&\multicolumn{4}{c}{CIFAR100}\\
            \cline{2-9}
            &\multicolumn{2}{c|}{PLCC (\%)}&\multicolumn{2}{c||}{ROCC (\%)}&\multicolumn{2}{c|}{PLCC (\%)}&\multicolumn{2}{c}{ROCC (\%)}\\
            \cline{2-9}
            &Gen. Gap&Test Acc.&Gen. Gap&Test Acc.&Gen. Gap&Test Acc.&Gen. Gap&Test Acc.\\
            \specialrule{2pt}{1pt}{1pt}
             ResNet18&$88.18$&$88.18$&$74.54$&$74.54$&$69.09$&$47.27$&$52.73$&$34.55$  \\
             ResNet34&$83.63$&$81.82$&$70.90$&$67.27$&$91.82$&$90.00$&$78.18$&$74.54$  \\
             ResNet50&$97.90$&$97.90$&$90.90$&$90.90$&$72.03$&$79.02$&$57.57$&$66.66$  \\
             ResNet101&$88.81$&$88.81$&$78.79$&$78.79$&$67.13$&$74.13$&$51.52$&$57.58$  \\
             \bottomrule
        \end{tabular}
        
    \caption{$Q$ Metric on ResNets}
    \label{fig:qc_metrics}
     \end{subfigure}
     \begin{subfigure}{0.2\textwidth}
     \centering
     \includegraphics[width=0.8\textwidth]{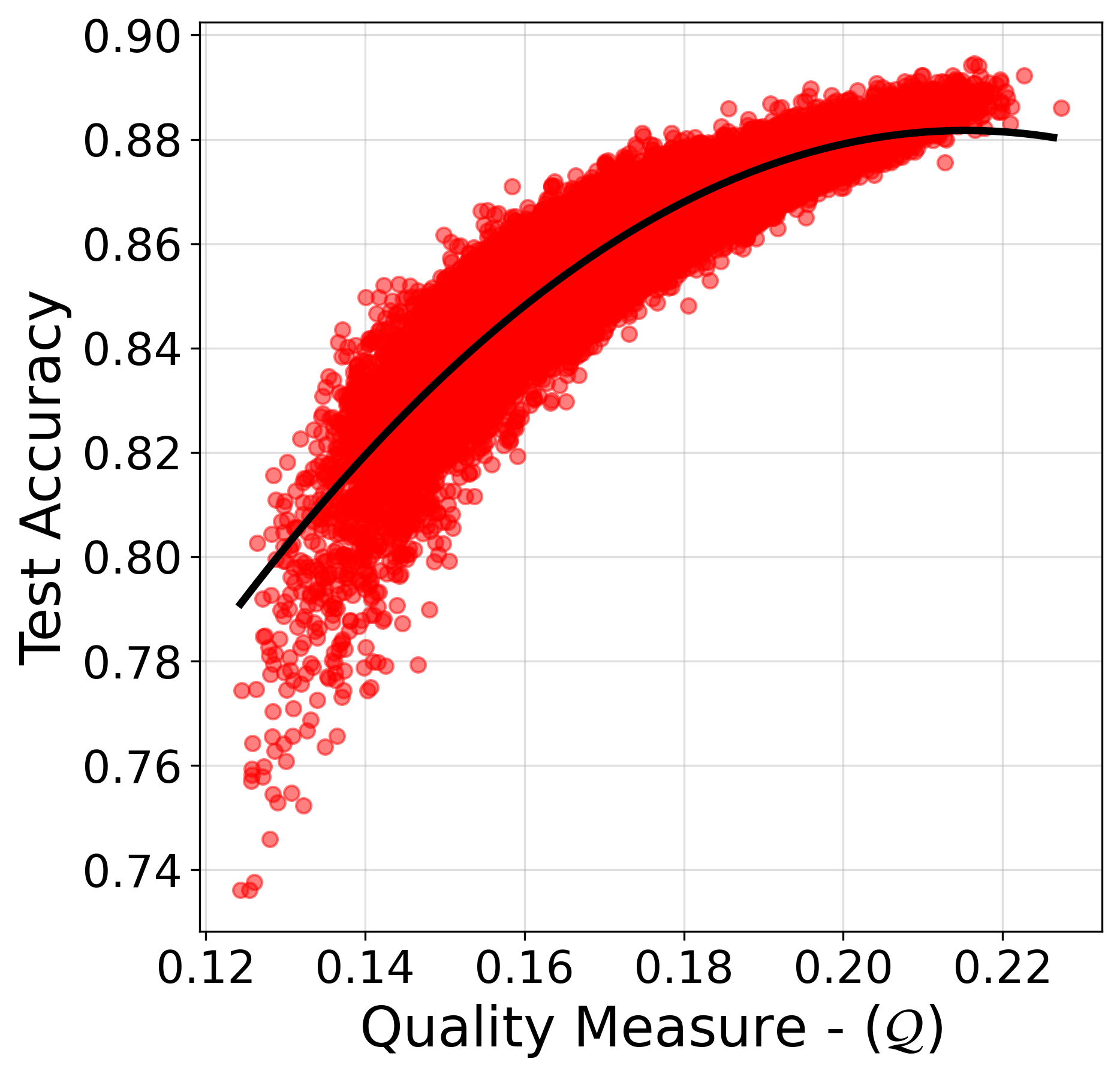}\\
     \includegraphics[width=0.8\textwidth]{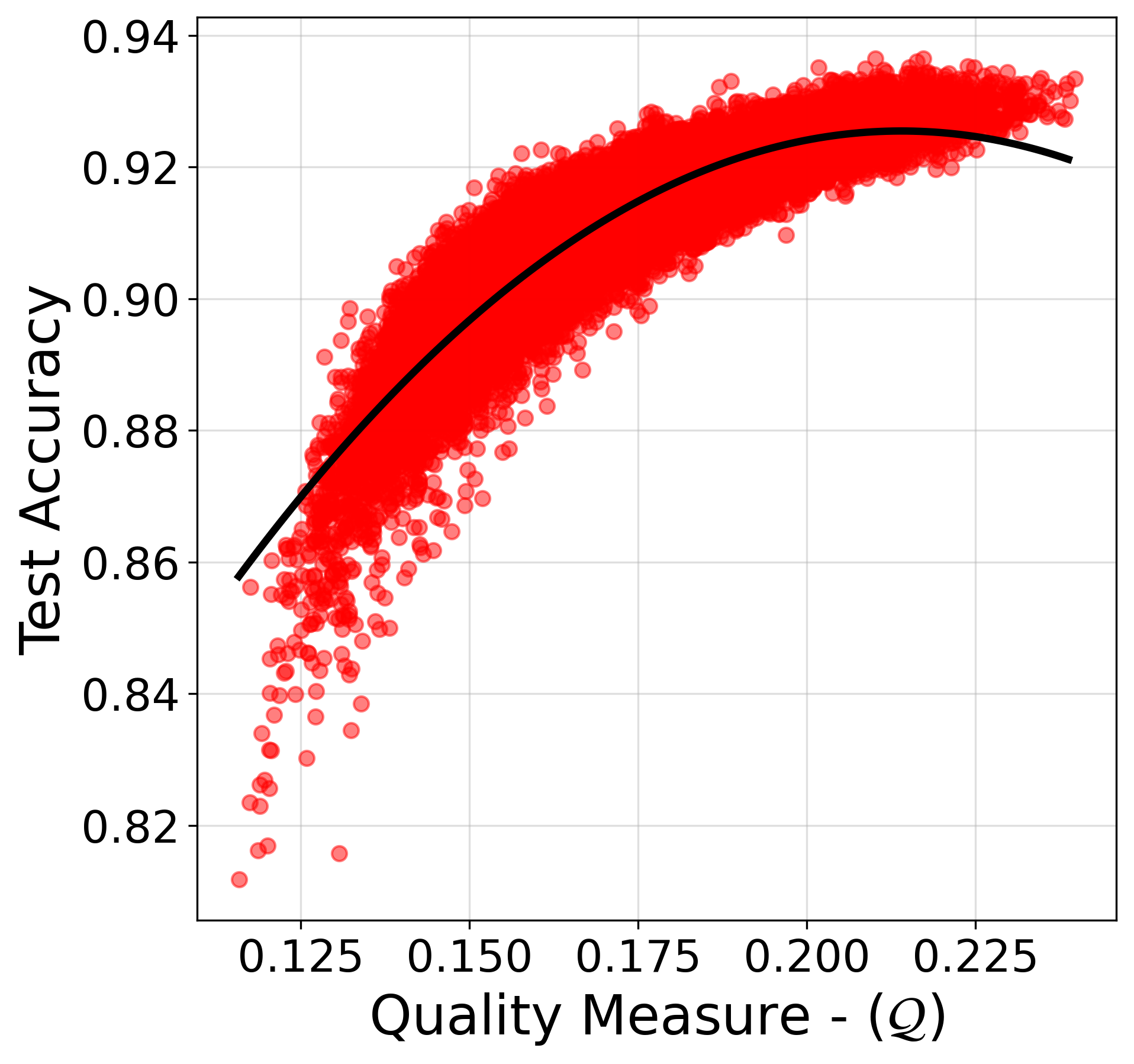}\\
     \caption{$Q$ Metric on NATS Benchmark}
    \label{fig:nats}
     \end{subfigure}
     \caption{Our proposed $\QC$ measure against generalization gap and test accuracy. (a) Applied on numerous optimizers and hyper-parameter settings (see Table \ref{table_cifar}) (dots represent a different experimental setup) on ResNets, with associated PLCC and ROCC correlations for our $\QC$ measure and generalization gap (Gen. Gap) and test accuracy (Test Acc.). We consider $\QC$ at the last epoch of training. (b) Applied on the NATS benchmark \cite{dong2021nats} on CIFAR10, where the top figure visualizes epoch $12$ and the bottom visualizes epoch $90$ of training.}
\end{figure*} 
We now aim to combine stable rank and condition number to develop a new quality measure that can help us quantify the quality of a learned network and describe its generalization characteristics. We know from \autoref{sec:optimal_metrics} that a stable rank of $1$ and condition number of $0$ would indicate a perfectly learned network. We propose the following quality measure to capture these properties:
\begin{align}
    q(\widehat{\tW}_{\ell}) = \arctan{\KG(\widehat{\tW}_{\ell})}/{\MC(\widehat{\tW}_{\ell})},~\text{where}~\ell\in[L]
\end{align}
where $\arctan(\cdot)$ is the element-wise arctangent, and $q$ is bound by $[0, \frac{\pi}{2}]$. Further, $q$ is maximized  when $\KG\rightarrow{1}$ and $\MC\rightarrow{0}$. \autoref{fig:quality_concept} visualizes the evolution of this quality measure on different conv layers of ResNet34 trained by different optimizers. The figure highlights the behaviour of this metric in response to poor and strong generalization performances. For instance, the Adam optimizer is known to provide poor generalization performance in CV applications, while the family of SGD-based optimizers are known to yield better performance \cite{wilson2017marginal, heo2021adamp, foret2021sharpnessaware}. We argue that this lack of performance from adaptive optimizers is realized by the low quality measure on latter layers of the ResNet34 model; this is visualized in \autoref{fig:quality_concept} by the darkness in the bottom of the figure. In contrast, the well performing SGD-based optimizers result in higher quality measure more consistently throughout the network, indicating better learned weights at all stages in the network.

We can define the overall quality measure of the network by aggregating all layers using an $L_2$-norm
\begin{align}
    Q = {1}/{\sqrt{L}} \parallel \underline{\bf{q}}\parallel_2^2,~\text{where}~\underline{\bf{q}}=[q(\widehat{\tW}_{1}),\cdots,q(\widehat{\tW}_{L})].
\end{align}
Normalizing by the square root of the number of layers accounts for the summation over layers within the $\ell_2$-norm space. We found this normalization method to behave better than a simple mean. 

We perform a small study on our quality metric against the popular NATS Benchmark \cite{dong2021nats} on CIFAR10 as well on a collection of ResNets on CIFAR10 and CIFAR100. We visualize our metric and associated correlation coefficients in \autoref{fig:qc_metrics} \& \autoref{fig:nats}. The NATS Benchmark is a Neural Architecture Search benchmark that provides model checkpoints at epoch 12 and epoch 90 in training, and whose goal is to predict performance/characterize generalization. These model checkpoints are of the 32,768 different topological models generated in the benchmark, and each involve a range of model sizes/complexities. As for the ResNets study, we employed a similar technique as Google's generalization prediction using margin distributions \cite{jiang2018predicting} whereby we trained various ResNets with various optimizers and computed our measure for each. The plots and table highlight how this metric is a strong indicator for both generalization performance as well as test performance, and most networks have very strong correlation scores.

Finally, we highlight that this explainability metric is (a) derived using only training data, meaning it can be used to predict the network's generalization to unseen data; (b) can quantify the contribution and strength of each layer in the network, not just the layer as a whole as in previous methods \cite{NIPS2017_10ce03a1, jiang2018predicting, Liu2020Understanding}; and (c) is unaffected by random initialization due to low-rank factorization. Note that these points also apply inherently to stable rank and condition number. This can potentially shed light on new layer-based optimization methods, one of which we explore in \autoref{sec_algorithm}. 
\section{Augmenting SGD by Probing Metrics}
\label{sec_algorithm}

Our goal in this section is to exploit the probing metrics developed in \autoref{sec:probing_metric} and develop an augmented SGD optimization algorithm that improves model performance.

\subsection{New Updating Mechanism on Vanilla SGD} \label{sec:vanilla_SGD}
Recall the SGD training objective to minimize an associated loss function given a training dataset $f({\tW}_{\ell};(X)^{\texttt{train}})$ \cite{polyakmomentum1964, bottou2010large, bottou2012stochastic, yuan2016influence, loizou2020momentum}. The update rule is then given by
\begin{equation}
{\tW}^{k}_{\ell}\longleftarrow{\tW}^{k-1}_{\ell} - \eta_k\overline{\nabla{f}}_k({\tW}^{k-1}_{\ell})
\label{eq_AdaS_1}
\end{equation}
for $k\in\{(t-1){K}+1,\cdots,t{K}\}$ where $t$ and $K$ correspond to epoch number and number of mini-batches, respectively, $\overline{\nabla{f}}_k({\tW}^{k-1}_{\ell})=1/|\Omega_k|\sum_{i\in\Omega_k}{\nabla f_i({\tW}^{k-1}_{\ell})}$ is the average stochastic gradients on $k$th mini-batch that are randomly selected from a batch of $n$-samples $\Omega_k\subset\{1,\cdots,n\}$, and $\eta_k$ defines the step-size taken toward the opposite direction of average gradients (i.e. the learning rate).

We aim to update the learning rate independently for each network layer after every epoch and therefore the step-size will be a function of epoch index and layer i.e. $\eta_k\equiv\eta_{\ell}(t)$. We now setup our problem by accumulating all observed gradients throughout $K$ mini-batch updates in one epoch as
\begin{equation}
{\tW}^{t}_{\ell}={\tW}^{t-1}_{\ell} - \eta_{\ell}(t-1)\overline{\nabla{f}}^t_{\ell},
\label{eq_AdaS_2}
\end{equation}
where, $\overline{\nabla{f}}^t_{\ell} = \sum^{t{K}}_{k=(t-1){K}+1}\overline{\nabla{f}}_k({\tW}^{k-1}_{\ell})$ corresponds to the total accumulated gradients in one training epoch. The idea here is to select a learning rate $\eta_{\ell}(t)$ such that the stable rank is increased over each epoch i.e. $\psi=\{\eta_{\ell}(t):~\KG(\widehat{\tW}^{t}_{\ell})\geq\KG(\widehat{\tW}^{t-1}_{\ell})\}$ in order to learn better encoding layers.

\begin{theorem}{(Increasing Stable Rank for Vanilla SGD).}\label{theorem_stable_rank_SGD}
Let the stable rank to be defined by \autoref{eq:stable_rank}. Starting with an initial learning rate $\eta_{\ell}(0)>0$ and setting the step-size of vanilla Stochastic Gradient Descent (SGD) proportional to
\begin{equation}
\eta_{\ell}(t) \triangleq~\zeta\left[\KG(\widehat{\tW}^{t}_{\ell}) - \KG(\widehat{\tW}^{t-1}_{\ell})\right]
\label{eq_AdaS_3}
\end{equation}
will guarantee the monotonic increase of the stable rank for the next epoch update $\KG(\widehat{\tW}^{t+1}_{\ell})\geq\KG(\widehat{\tW}^{t}_{\ell})$ for some existing lower bound $\eta(t)\geq{\eta_0}$ and $\zeta\geq{0}$.
\end{theorem}
The proof of \autoref{theorem_stable_rank_SGD} is provided in the supplementary material (Appendix-A).

With the start of a positive initial learning rate $\eta_{\ell}(0)>0$ and following the update rule in \autoref{eq_AdaS_3}, \autoref{theorem_stable_rank_SGD} guarantees the increase of the stable rank for the next epoch update for Vanilla SGD. Accordingly, learning rates using \autoref{eq_AdaS_3} will remain positive over consecutive epochs. For proof of demonstration, refer to \autoref{fig:figure_Optimizer_behaviour}. We note in \autoref{fig:figure_Optimizer_behaviour} that the SGD type training degrades in stable rank over time, as the network learns more, while in contrast our method RMSGD doesn’t. The idea is to have a high stable rank after training has completed, which we can see that SGD’s algorithm does not exhibit stable behaviour. Further, SGD has higher stable rank, but worse (i.e. higher) condition number. In contrast, RMSGD has a good stable rank, as well as a much better condition number (i.e. lower).

\subsection{\AlgName: Augmented SGD Algorithm}
The step-size defined in \autoref{sec:vanilla_SGD} for Vanilla SGD is measured only for two consecutive epochs. Due to the stochastic nature of gradients, the in-practice step-size can fluctuate. To reduce step-size oscillations, we employ a momentum algorithm for the historical accumulation of step-sizes and stable ranks over epoch updates. We augment the update rule for SGD by performing the following:
\begin{align}
\begin{array}{l}
\text{(i) revise learning rate by average momentum:} \\~~~~~~~\eta_{\ell}(t)\leftarrow \beta\eta_{\ell}(t-1) + \zeta[\KG(\widehat{\tW}^{t}_{\ell}) - \KG(\widehat{\tW}^{t-1}_{\ell})],\\
\text{(ii) apply gradients through an average momentum}\\~~~~~~~{\bm v}^{k}_{\ell} \leftarrow \alpha{\bm v}^{k-1}_{\ell} - \eta_{\ell}(t){{\bm g}^{k}_{\ell}},\\
\text{(iii) update network weights:}\\~~~~~~~{\bm w}^k_{\ell}\leftarrow {\bm w}^{k-1}_{\ell} + v_{\ell}^k,\nonumber
\end{array}
\end{align}
where, $k$, $t$ and $\ell$ correspond to the current mini-batch, current epoch, and network layer indexes, respectively. $\bm{v}$ is the velocity term, and $\bm{w}$ are the learnable parameters. Notice there are two associated momentum parameters: (1) SGD momentum fixed at $\alpha=0.9$; and (2) learning momentum fixed at $\beta=0.98$. We pick $\zeta=1$ and learning momentum $\beta<1$ to remain within the convergence bound of the unit circle associated with the update rule (i). Setting $\beta$ trades-off between faster convergence and increased stable rank, which eventually leads to higher performance. An ablative study of this parameter is shown in \autoref{fig:figure_Optimizer_behaviour}. We provide further study of this parameter in Appendix-B. We dub this optimizer \textbf{R}ank \textbf{M}omentum \textbf{SGD}: \textbf{RMSGD}\footnote{Code from \url{https://github.com/mahdihosseini/RMSGD}}, and show pseudo-code in Algorithm \ref{algorithm_AugSGD}. The condition number plays no role in this algorithm, but we introduced it previously as an additional metric to evaluate/quantify learning. We show here that just one of our metrics is sufficient to develop our algorithm, and future work might include combining both.

\begin{figure*}[htp]
    \centering
    \includegraphics[width=0.5\textwidth]{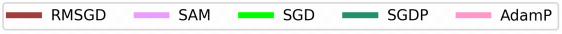}\\  
    \includegraphics[height=0.175\textwidth]{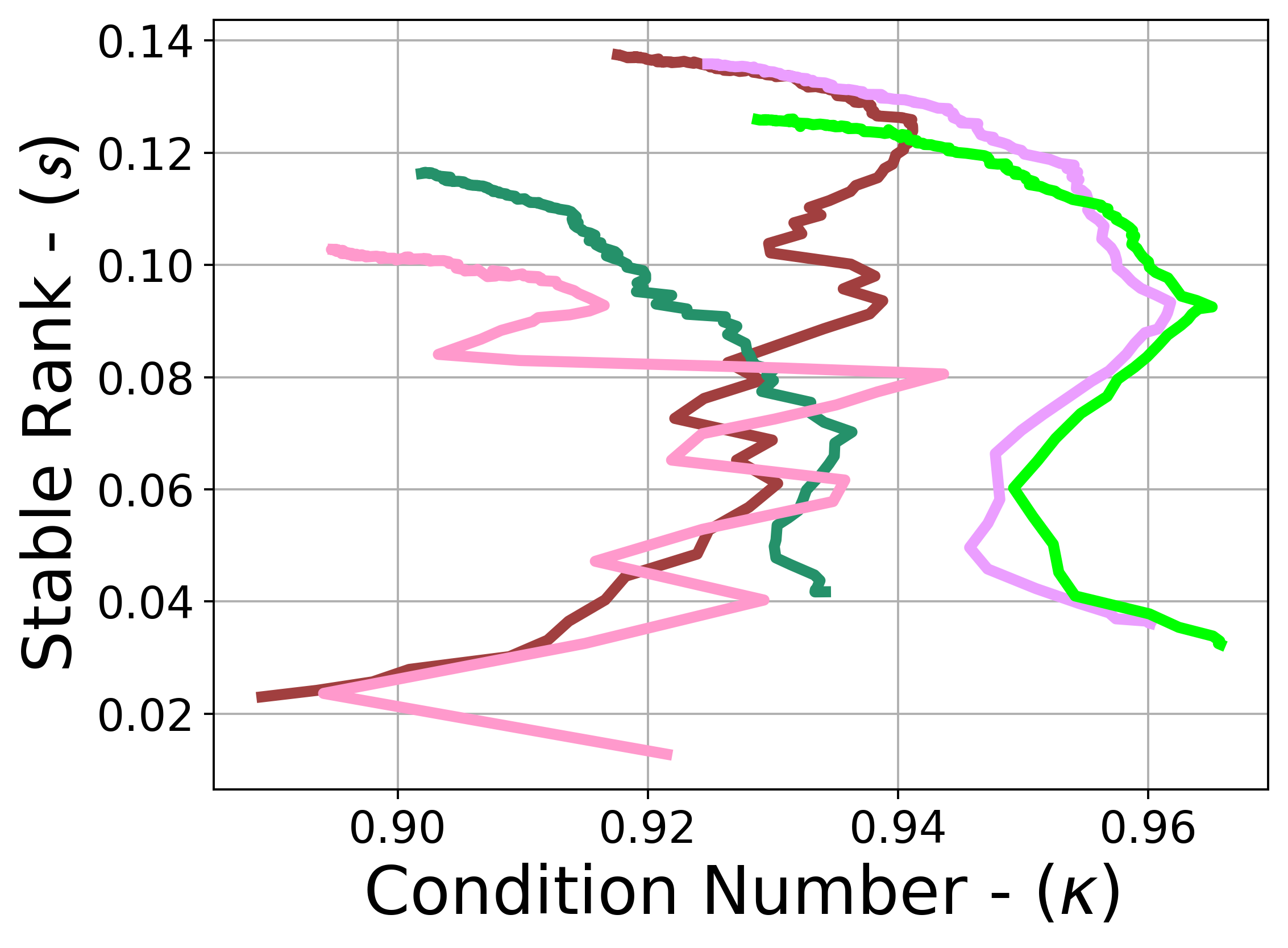}
    \includegraphics[height=0.175\textwidth]{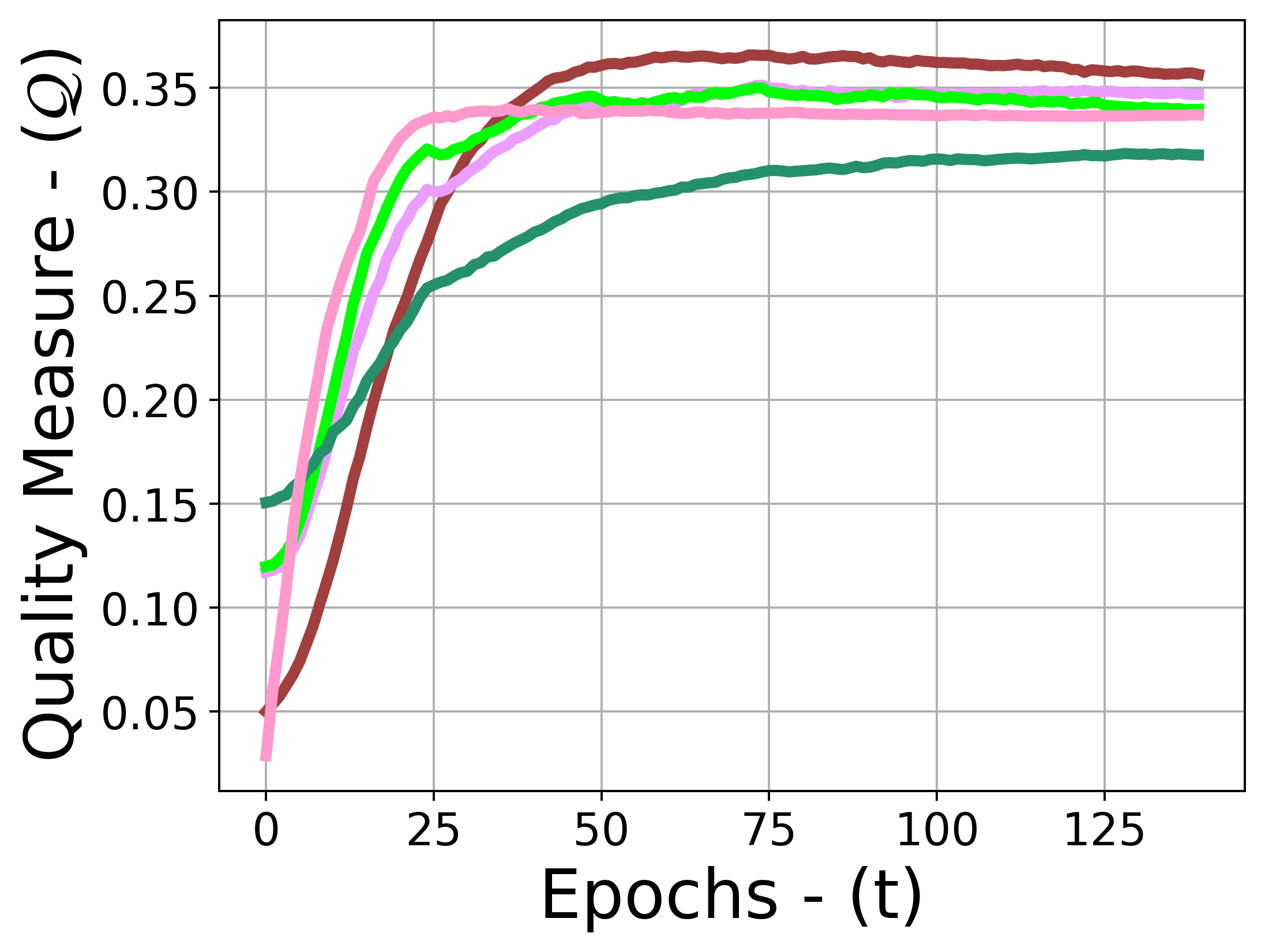}
    \includegraphics[height=0.175\textwidth]{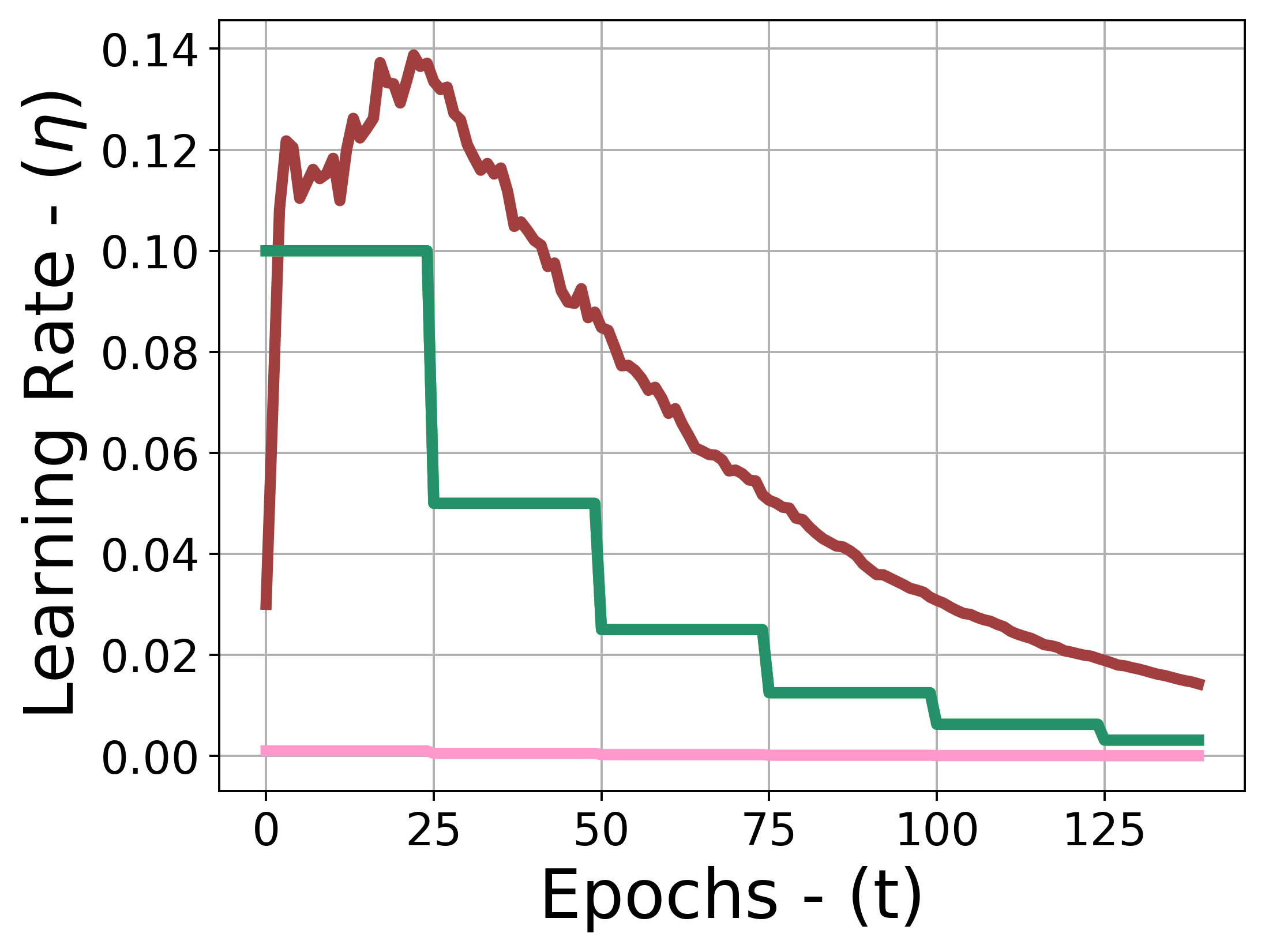}
    \includegraphics[height=0.175\textwidth]{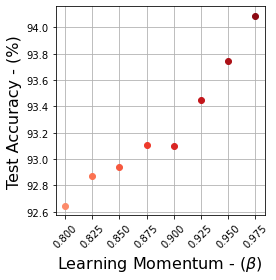}
    \caption{Comparison of (left) explainability metrics, (middle-left) Quality Measure, and (middle-right) learning rates for optimizers on ResNet50 applied to CIFAR10 over 140 epochs.  (right) Effect of varying the learning momentum factor $\beta$ on testing accuracy for ResNet34 on CIFAR10.}
    \label{fig:figure_Optimizer_behaviour}
\end{figure*}

\begin{algorithm}[htp]
\SetAlgoLined
\begin{algorithmic}
\footnotesize{
\REQUIRE batch size $|\Omega_k|$, \# epochs $T$, \# layers $L$, initial step-sizes $\left\{\eta_{\ell}(0)=0.03\right\}^{L}_{\ell=1}$, initial vectors $\left\{v_{\ell}^0, {\bm w}_{\ell}^0\right\}_{{\ell}=1}^L$, SGD momentum $\alpha=0.9$, learning momentum $\beta=0.98$, $\zeta=1$\\
\FOR{$t$ = $1$ : $T$}
\STATE \textcolor{black}{\textit{Stage-I: SGD optimization augmented by adaptive learning rates}} \\
- generate $K$ mini-batches: \(\{\Omega_k\subset\{1,\cdots,n\}\}^K_{k=1}\)\\
\FOR{$k = (t-1){K}+1 : tK$}
\STATE 1. compute gradients: ${\bm g}^{k}_{\ell} \leftarrow 1/|\Omega_k|\sum_{i\in\Omega_k}{\nabla f_i({\bm{w}}^{k-1}_{\ell})}~;\ell\in\{1,\cdots,L\}$\\
\STATE 2. compute velocity terms: ${\bm v}^{k}_{\ell} \leftarrow \alpha{\bm v}^{k-1}_{\ell} - \eta_{\ell}(t){{\bm g}^{k}_{\ell}}$\\
\STATE 3. apply updates: ${\bm w}^{k}_{\ell} \leftarrow {\bm w}^{k-1}_{\ell} + {\bm v}^{k}_{\ell}$
\ENDFOR
\STATE \textcolor{black}{\textit{Stage-II: adaptive computation of learning rates}} \\
\FOR{$\ell$ = $1$ : $L$}
\STATE 1. Factorize matrix weights $\widehat{\tW}^{t}_{\ell}$ by \autoref{eq:lowrank_factorization} using \texttt{EVBMF}  \cite{nakajima2013global}\\
\STATE 2. compute stable rank $\KG(\widehat{\tW}_{\ell})$ by \autoref{eq:stable_rank}\\
\STATE 3. update LR: $\textcolor{black}{\eta_{\ell}(t)}\leftarrow \beta\eta_{\ell}(t-1) + \zeta[\KG(\widehat{\tW}^{t}_{\ell}) - \KG(\widehat{\tW}^{t-1}_{\ell})]$\\
\ENDFOR
\ENDFOR
}
\end{algorithmic}
\caption{Augmented SGD (\AlgName)}
\label{algorithm_AugSGD}
\end{algorithm}

Each layer in the network is assigned an index $\{\ell\}^{L}_{\ell=1}$ where all learnable parameters (e.g. conv, linear transform, biases, batch-norms, etc) are called using this index. The goal in {\AlgName} is first to callback the matrix weights of each layer, second to compute the stable rank $\KG(\widehat{\tW}^{t}_{\ell})$ of each layer using  \autoref{eq:stable_rank}, third to compute the difference gain using \autoref{eq_AdaS_3}, and finally accumulate this difference value in a decaying momentum fashion. This is performed to compute the learning rate of each layer independently for the current epoch $t$ to augment the SGD optimization framework. We note here that the initial stable rank is zero $\KG(\widehat{\tW}^{0}_{\ell})=0$ for all layers due to random initialization of weights. \autoref{fig:figure_Optimizer_behaviour} demonstrates an evolution example of the learning rate adapted by \AlgName. Interestingly, the behavior of learning rate coincides with scheduling techniques using cyclical learning in \cite{smith2017cyclical, smith2019super} and warm-up/cosine decay in \cite{he2019bag}. We believe our method explains the intuition behind such scheduling technique and why the learning rate is suggested to start from small value and increase with proceeding epochs (i.e. warm-up), and then decay to finalize training.

\begin{proposition}{(Convergence Guarantees of RMSGD). Given RMSGD only evolves the learning rate and ensures that $\eta_{\ell}(t) > 0~\forall t \in \{1, \hdots, T\}$, its convergence guarantee follows that of SGD \cite{zhou2021convergence, loizou2020momentum}.}
\end{proposition}
We further elaborate on the proof in Appendix-A.
\section{Empirical Evaluation}\label{sec:experiment}
\textbf{Setup.} We evaluate \AlgName\ against state-of-the-art adaptive and non-adaptive optimizers and conduct different studies including (a) image classification on CIFAR10/CIFAR100 \cite{krizhevsky2009learning}, ImageNet \cite{imagenet2015}, as well as on two computational pathology datasets (MHIST \cite{wei2021petri} and ADP \cite{hosseini2019atlas}). Hyper-parameters are tuned using ResNet18 on CIFAR10; (b) image classification with Cutout \cite{devries2017improved}; and (c) batch size robustness. Grid search is used for learning rate tuning. For full details on data augmentation, hyper-parameter tuning, and additional results, see Appendix-B.

\textbf{Hardware.} A single-GPU (RTX2080Ti) was used for each experiment, see Appendix-B for specifics.

\textbf{Note.} For all tables, {\color{best}green} shows the best result and {\color{close}orange} is within standard deviation from best.

\begin{table*}[ht]
    \centering
    \setlength{\tabcolsep}{1pt}
    \caption{Perfomance of various networks and optimizers on CIFAR10 and CIFAR100 (a) without Cutout and (b) with Cutout. Reported using wall clock time of $250$ SGD training epochs as the cutoff. Note ResNet$\{18,34,50,101\}$=R$\{18,34,50,101\}$ and ResNeXt=RNeXt.}
    \scriptsize
    \begin{subfigure}{\textwidth}
    \centering
    \caption{Without Cutout}
    \begin{tabular}{c|c|c|c|c|c|c|c|c|c|c||c}
    &Network&AdaBelief\textsuperscript{\cite{zhuang2020adabelief}}&AdaBound\textsuperscript{\cite{luo2018adaptive}}&AdaGrad\textsuperscript{\cite{duchi2011}}&Adam\textsuperscript{\cite{kingma2014adam}}&AdamP\textsuperscript{\cite{heo2021adamp}}&SLS\textsuperscript{\cite{vaswani2019painless}}&SAM\textsuperscript{\cite{foret2021sharpnessaware}}&SGD\textsuperscript{\cite{goodfellow2016deep}}&SGDP\textsuperscript{\cite{heo2021adamp}}&\AlgName\\
    \specialrule{2pt}{1pt}{1pt}
    \multirow{5}{*}{\rotatebox[origin=c]{90}{CIFAR10}}
    &R18\textsuperscript{\cite{he2016deep}}&$93.34_{0.10}$&$93.84_{0.09}$&$92.45_{0.24}$&$93.27_{0.10}$&$94.82_{0.10}$&$93.62_{0.10}$&${\color{close}\mathbf{95.58_{0.07}}}$&$95.32_{0.07}$&${\color{close}\mathbf{95.39_{0.16}}}$&${\color{best}\mathbf{95.66_{0.17}}}$\\
    &R34\textsuperscript{\cite{he2016deep}}&$93.55_{0.05}$&$93.79_{0.19}$&$92.59_{0.30}$&$93.47_{0.18}$&$95.14_{0.25}$&$93.45_{0.16}$&${\color{best}\mathbf{95.81_{0.16}}}$&${\color{close}\mathbf{95.56_{0.10}}}$&${\color{close}\mathbf{95.75_{0.14}}}$&${\color{close}\mathbf{95.71_{0.07}}}$\\
    &R50\textsuperscript{\cite{he2016deep}}&$93.70_{0.18}$&$94.00_{0.15}$&$92.12_{0.23}$&$92.67_{0.12}$&$94.69_{0.10}$&$92.70_{0.18}$&$95.20_{0.18}$&$95.05_{0.28}$&$95.19_{0.15}$&${\color{best}\mathbf{95.63_{0.05}}}$\\
    &R101\textsuperscript{\cite{he2016deep}}&$93.86_{0.20}$&$94.17_{0.13}$&$92.51_{0.22}$&$93.13_{0.08}$&$94.92_{0.24}$&$64.20_{20.98}$&${\color{close}\mathbf{95.40_{0.12}}}$&${\color{close}\mathbf{95.30_{0.13}}}$&${\color{close}\mathbf{95.36_{0.04}}}$&${\color{best}\mathbf{95.53_{0.14}}}$\\
    &RNeXt\textsuperscript{\cite{xie2017aggregated}}&$93.55_{0.05}$&$92.83_{0.14}$&$91.09_{0.19}$&$91.78_{0.16}$&$93.82_{0.10}$&$93.67_{0.09}$&$94.38_{0.09}$&$94.62_{0.09}$&$94.79_{0.24}$&${\color{best}\mathbf{95.49_{0.05}}}$\\
    \hline
    \hline
    
    \multirow{5}{*}{\rotatebox[origin=c]{90}{CIFAR100}}
    &R18&$73.11_{0.21}$&$74.09_{0.27}$&$70.92_{0.31}$&$72.45_{0.34}$&$76.81_{0.31}$&$73.59_{0.04}$&$77.16_{0.25}$&$77.80_{0.07}$&${\color{close}\mathbf{78.13_{0.16}}}$&${\color{best}\mathbf{78.63_{0.34}}}$\\
    &R34&$73.43_{0.14}$&$74.84_{0.18}$&$70.39_{0.57}$&$72.09_{0.50}$&$76.93_{0.40}$&$73.22_{0.11}$&$77.98_{0.39}$&$77.88_{0.39}$&$78.74_{0.12}$&${\color{best}\mathbf{79.32_{0.10}}}$\\
    &R50&$75.15_{0.45}$&$75.52_{0.37}$&$70.60_{0.91}$&$70.53_{0.36}$&$77.47_{0.16}$&$75.80_{0.23}$&$77.39_{0.66}$&$78.12_{0.42}$&$78.44_{0.24}$&${\color{best}\mathbf{79.59_{0.54}}}$\\
    &R101&$75.63_{0.10}$&$76.31_{0.41}$&$72.39_{0.84}$&$72.20_{0.68}$&$77.71_{0.16}$&$73.31_{0.84}$&$78.38_{0.48}$&$78.48_{0.45}$&${\color{close}\mathbf{78.60_{0.55}}}$&${\color{best}\mathbf{79.36_{0.26}}}$\\
    &RNeXt&$72.64_{0.49}$&$72.97_{0.38}$&$68.83_{0.43}$&$71.54_{0.41}$&$74.54_{0.40}$&$72.35_{0.42}$&$75.83_{0.30}$&$75.36_{0.33}$&${\color{close}\mathbf{76.56_{0.33}}}$&${\color{best}\mathbf{77.14_{0.31}}}$\\
    \end{tabular}
  
    \label{table_cifar}
    \end{subfigure}
    \begin{subfigure}{\textwidth}
    \centering
    \caption{With Cutout}
    \begin{tabular}{c|c|c|c||c|c|c|c||c}
    &\multicolumn{4}{c|}{CIFAR10}&\multicolumn{4}{c}{CIFAR100}\\
    \cline{2-9}
    Network&SAM\textsuperscript{C}&SGD\textsuperscript{C}&SGDP\textsuperscript{C}&\AlgName\textsuperscript{C}&SAM\textsuperscript{C}&SGD\textsuperscript{C}&SGDP\textsuperscript{C}&\AlgName\textsuperscript{C}\\
    \specialrule{2pt}{1pt}{1pt}
    ResNet18&${\color{close}\mathbf{95.96_{0.13}}}$&${\color{close}\mathbf{96.12_{0.13}}}$&${\color{close}\mathbf{96.13_{0.13}}}$&${\color{best}\mathbf{96.13_{0.08}}}$&$77.58_{0.11}$&$78.16_{0.21}$&${\color{best}\mathbf{78.82_{0.37}}}$&${\color{close}\mathbf{78.53_{0.22}}}$\\
    ResNet34&${\color{close}\mathbf{96.64_{0.09}}}$&${\color{close}\mathbf{96.53_{0.13}}}$&${\color{best}\mathbf{96.70_{0.10}}}$&$96.42_{0.08}$&$78.57_{0.19}$&$78.63_{0.55}$&${\color{close}\mathbf{79.67_{0.24}}}$&${\color{best}\mathbf{79.70_{0.19}}}$\\
    ResNet50&$95.79_{0.10}$&$95.78_{0.27}$&$96.03_{0.16}$&${\color{best}\mathbf{96.28_{0.07}}}$&$77.73_{0.28}$&$78.36_{0.67}$&${\color{close}\mathbf{79.52_{0.31}}}$&${\color{best}\mathbf{80.06_{0.45}}}$\\
    ResNet101&${\color{close}\mathbf{96.17_{0.08}}}$&$96.04_{0.16}$&$96.12_{0.05}$&${\color{best}\mathbf{96.33_{0.08}}}$&${\color{close}\mathbf{79.41_{0.66}}}$&$79.35_{0.62}$&${\color{close}\mathbf{80.03_{0.67}}}$&${\color{best}\mathbf{80.36_{0.35}}}$\\
    ResNeXt&$95.01_{0.18}$&$95.04_{0.18}$&$95.24_{0.15}$&${\color{best}\mathbf{95.62_{0.08}}}$&$76.34_{0.06}$&$75.91_{0.19}$&$77.14_{0.21}$&${\color{best}\mathbf{78.06_{0.28}}}$\\
    MobileNetV2\textsuperscript{\cite{sandler2018mobilenetv2}}&$94.65_{0.12}$&$94.53_{0.16}$&$94.07_{0.07}$&${\color{best}\mathbf{95.48_{0.11}}}$&$75.43_{0.18}$&$73.94_{0.21}$&$73.40_{0.07}$&${\color{best}\mathbf{76.36_{0.27}}}$\\
    SENet18\textsuperscript{\cite{hu2018squeeze}}&${\color{close}\mathbf{95.92_{0.16}}}$&${\color{close}\mathbf{95.99_{0.10}}}$&${\color{best}\mathbf{96.04_{0.08}}}$&$95.80_{0.06}$&${\color{close}\mathbf{77.64_{0.32}}}$&${\color{best}\mathbf{77.80_{0.23}}}$&${\color{close}\mathbf{77.70_{0.05}}}$&${\color{close}\mathbf{77.77_{0.15}}}$\\
    EfficientNetB0\textsuperscript{\cite{tan2019efficientnet}}&$90.44_{0.23}$&$91.70_{0.24}$&$92.09_{0.21}$&${\color{best}\mathbf{92.83_{0.14}}}$&${\color{close}\mathbf{69.10_{0.50}}}$&$68.42_{0.24}$&${\color{close}\mathbf{68.98_{0.44}}}$&${\color{best}\mathbf{69.87_{0.49}}}$\\
    ShuffleNetV2\textsuperscript{\cite{ma2018shufflenet}}&${\color{close}\mathbf{94.71_{0.15}}}$&$94.40_{0.21}$&$94.37_{0.12}$&${\color{best}\mathbf{95.01_{0.29}}}$&${\color{best}\mathbf{74.70_{0.19}}}$&$74.13_{0.35}$&${\color{close}\mathbf{74.44_{0.29}}}$&${\color{close}\mathbf{74.38_{0.36}}}$
    \label{table_cutout}
    \end{tabular}
    \end{subfigure}
\end{table*}
\subsection{Image Classification}
\textbf{Note on CIFAR experiments.} Since different optimizers exhibit widely different epoch times (see \autoref{fig:curves}), for a fair comparison, we limit training to the total wall clock time consumed by $250$ epochs using SGD. In terms of epochs, this amounts to $\sim250$ epochs for all optimizers except SAM, which consumes only $\sim128-133$ epochs due to its $2$ forward passes and twice as long epoch times.

\textbf{Computer Vision: CIFAR.} We report all results in \autoref{table_cifar} and some in \autoref{fig:curves}, with the other figures in Appendix-B. \AlgName\ performs consistently optimally, and shows strength with increasing dataset complexity (CIFAR10 $\rightarrow$ CIFAR100) and increasing network complexity (ResNet18 $\rightarrow$ ResNet34 $\rightarrow$ ResNet50 $\rightarrow$ ResNet101). We note that SGDP and SAM do remain within standard deviations of performance in many cases, which highlights the competitiveness of these optimizers. Note that for other adaptive optimizers, the competitiveness is non-existent.

We further highlight the performance to time results shown in \autoref{fig:curves}. Despite SAM's competitive performance, it consumes $\sim2\times$ as many seconds per epoch to train. We highlight how \AlgName\ is able to outperform while incurring very low computational overhead. SAM does exhibit better generalization by its lower train-test gap. Note that for scenarios where additional resources or time is available, we also report final performance of all optimizers at epoch-$250$ in Appendix-B.

\textbf{Computer Vision: CIFAR with Cutout.} We also compare \AlgName\ using cutout, as one may argue that cutout can be used to augment SGD and negate the need for adaptive optimizers. However, we can see from Table \ref{table_cutout} that \AlgName\ with cutout is still able to consistently outperform SAM, SGD, and SGDP, and incurs great improvements ($\sim1\%$) over its non-cutout performance.

\textbf{Computer Vision: ImageNet.} We report ImageNet results in \autoref{imagenet_table}, \autoref{imagenet_table2}, and \autoref{fig:curves}. The $\sim7\%$ top-1 test accuracy performance gap highlights \AlgName's strength over SGD and SAM, even with no ImageNet-specific hyper-parameter tuning. This highlights how \AlgName\ may be used to tackle larger datasets with smaller computational resources. SGD and \AlgName\ took $1$ week to train, while SAM took $2$. The performance differences from the original MobileNetV2 \cite{sandler2018mobilenetv2} relate to the smaller batch size of $128$.

\begin{table}
    \setlength{\tabcolsep}{1pt}
    \caption{Test accuracy results for MobileNetV2 on ImageNet using a batch size of $128$, trained a single GPU machine. }
    \vspace{-4mm}
    \scriptsize
    \begin{center}
    \begin{tabular}{c|c|c||c|c||c|c}
         \multirow{2}{*}{\rotatebox[origin=c]{0}{Epoch}}&\multicolumn{2}{c||}{SAM}&\multicolumn{2}{c||}{SGD}&\multicolumn{2}{c}{\AlgName}  \\
         \cline{2-7}
         &Top-1&Top-5&Top-1&Top-5&Top-1&Top-5\\
         \specialrule{2pt}{1pt}{1pt}
         \rotatebox[origin=c]{0}{150}&$61.41$&$83.86$&$59.80$&$82.56$&${\color{best}\mathbf{67.84}}$&${\color{best}\mathbf{88.32}}$\\
         \rotatebox[origin=c]{0}{250}&$63.43$&$85.24$&$62.16$&$84.21$&${\color{best}\mathbf{70.25}}$&${\color{best}\mathbf{89.66}}$
         \label{imagenet_table}
    \end{tabular}
    \end{center}
    \vspace{-6mm}
\end{table}
\begin{table}
    \setlength{\tabcolsep}{1pt}
    \caption{Test accuracy results for MobileNetV2 and ResNet50 on ImageNet using a batch size of $256$, trained a single GPU machine. }
    \vspace{-4mm}
    \scriptsize
    \begin{center}
    \begin{tabular}{c|c|c|c||c}
         Network&SAM&SGD&AdamP&\AlgName\\
         \specialrule{2pt}{1pt}{1pt}
         MobileNetV2&$63.38$&	$64.61$&	$69.40$&	${\color{best}\mathbf{71.24}}$\\
         ResNet50	&$75.51$&	$76.12$&	$75.85$&	${\color{best}\mathbf{76.42}}$
         \label{imagenet_table2}
    \end{tabular}
    \end{center}
    \vspace{-6mm}
\end{table}
\begin{table}
\setlength{\tabcolsep}{1pt}
\caption{Test accuracy results for computational pathology datasets. Note that ResNet18=R18, ResNet34=R34, and MobileNetV2=MV2}
\vspace{-5mm}
    \tiny
    \begin{center}
    \begin{tabular}[b]{c|c|c|c|c|c|c||c}
         &Net&Adam&AdamP&SAM&SGD&SGDP&\AlgName  \\
         \specialrule{2pt}{1pt}{1pt}
         \multirow{3}{*}{\rotatebox[origin=c]{90}{MHIST}}&
         R18&{\color{close}$\mathbf{79.92_{0.80}}$}&{\color{close}$\mathbf{80.57_{1.43}}$}&{\color{close}$\mathbf{80.78_{1.61}}$}&{\color{close}$\mathbf{80.76_{1.32}}$}&{\color{close}$\mathbf{80.53_{0.65}}$}&{\color{best}$\mathbf{81.80_{1.20}}$}\\
         &R34&{\color{best}$\mathbf{81.02_{0.95}}$}&{\color{close}$\mathbf{80.57_{0.88}}$}&{\color{close}$\mathbf{80.72_{1.12}}$}&{\color{close}$\mathbf{79.65_{1.98}}$}&{\color{close}$\mathbf{80.16_{0.85}}$}&{\color{close}$\mathbf{80.94_{0.85}}$}\\
         &MV2&{\color{close}$\mathbf{79.90_{0.95}}$}&{\color{close}$\mathbf{80.59_{0.52}}$}&{\color{best}$\mathbf{82.89_{0.93}}$}&{\color{close}$\mathbf{81.41_{0.81}}$}&{\color{close}$\mathbf{80.80_{1.58}}$}&{\color{close}$\mathbf{82.58_{0.64}}$}\\
         \hline\hline
         \multirow{3}{*}{\rotatebox[origin=c]{90}{ADP}}&
         R18&$92.75_{0.20}$&$94.04_{0.12}$&$93.28_{0.12}$&$93.22_{0.17}$&$93.48_{0.43}$&{\color{best}$\mathbf{94.27_{0.10}}$}\\
         &R34&$92.80_{0.06}$&{\color{close}$\mathbf{93.95_{0.08}}$}&$93.24_{0.11}$&$93.38_{0.14}$&$93.65_{0.12}$&{\color{best}$\mathbf{94.19_{0.21}}$}\\
         &MV2&$92.89_{0.17}$&{\color{close}$\mathbf{93.78_{0.09}}$}&$89.43_{1.46}$&$88.81_{1.32}$&$91.42_{0.53}$&{\color{best}$\mathbf{93.83_{0.28}}$}
         \label{mhist_table}
    \end{tabular}
    \end{center}
    \vspace{-8mm}
\end{table}

\textbf{Computational Pathology: MHIST \& ADP} We report our computational pathology results in \autoref{mhist_table} and some shown in  \autoref{fig:curves}. We show that \AlgName\ is able to outperform all optimizers on ADP. All optimizers tend to perform similarly on MHIST, likely due to its smaller size. Computational pathology dataset are included in the experiments as an additional (less conventional) dataset that pose additional challenges (e.g. difficulty and scarcity of labels, image size).

\begin{wrapfigure}[7]{r}{0.2\textwidth}
    \hspace{-4mm}
    \includegraphics[trim = {2mm 5mm 2mm 10mm}, width=0.2\textwidth]{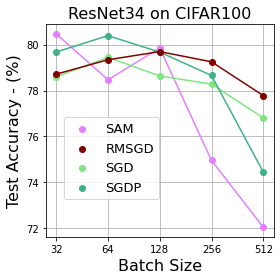}
    \label{fig_batch}
\end{wrapfigure}
\textbf{Batch Size.} We report our ablative batch size study in Figure 4. We highlight RMSGD’s greater robustness to varying batch sizes, particularly compared to SAM. SGD exhibits robustness but has inferior performance. 

\begin{figure*}[th]
    \centering
    \includegraphics[width=0.7\textwidth]{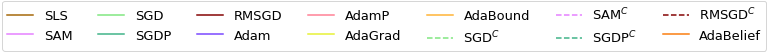}\\
    \centering
    \includegraphics[height=0.195\textwidth]{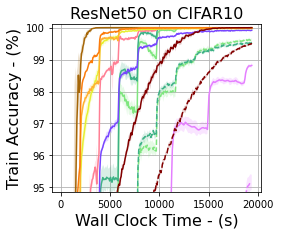}
    \includegraphics[height=0.195\textwidth]{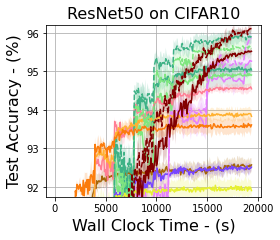}
    \includegraphics[height=0.195\textwidth]{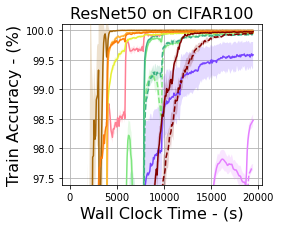}
    \includegraphics[height=0.195\textwidth]{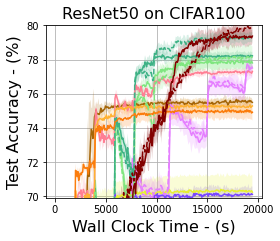}
    \centering
    \includegraphics[height=0.2\textwidth]{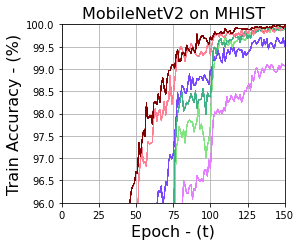}
    \includegraphics[height=0.2\textwidth]{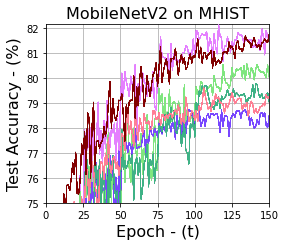}
    \includegraphics[height=0.2\textwidth]{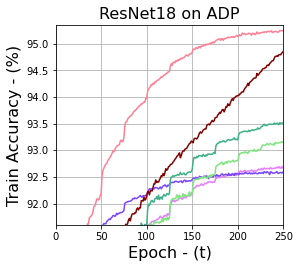}
    \includegraphics[height=0.2\textwidth]{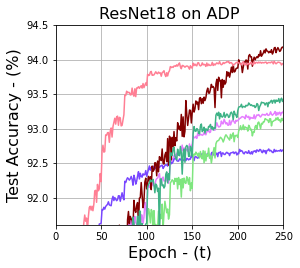}
    \centering
    \includegraphics[height=0.2\textwidth]{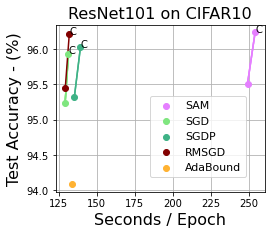}
    \includegraphics[height=0.2\textwidth]{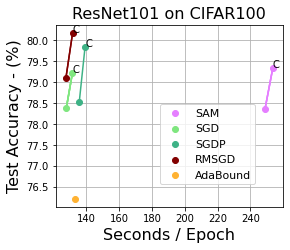}
    \includegraphics[height=0.2\textwidth]{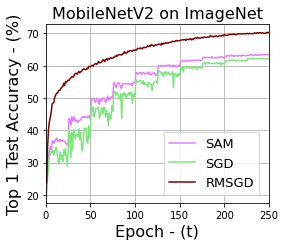}
    \includegraphics[height=0.2\textwidth]{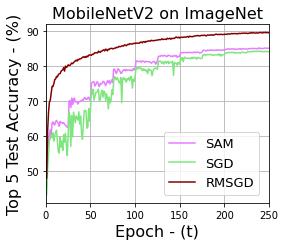}
    \caption{Test/train performance on different selections of dataset, network and optimizer. The cutout augmentation is annotated with a $C$. A single GPU is used for all experiments including for ImageNet results with batch size of $128$.}
    \label{fig:curves}
\end{figure*}
\section{Concluding remarks}\label{sec:conclusion}
In this work we introduced probing metrics (i.e. stable rank, condition number, and a quality measure) and demonstrated how they can be used to quantify learning of neural networks and be used as indicators for generalization performance. We demonstrated how these metrics can be exploited very simply to augment vanilla SGD and achieve significant performance gains with extremely low computational overhead ($<1$s per training epoch).

\textbf{Computational cost vs. performance benefits.} We highlight that \AlgName\ is able to remain extremely computationally efficient, while incurring performance benefits across many applications, network complexities, and dataset complexities. This is in contrast to optimizers like SGDP, who are superior to SGD but incur computational cost and slightly lower performance than \AlgName. SAM has also shown to result in performance improvements over SGD, however at a significant cost computationally (double that of SGD). We further highlight that SAM tends to also lose performance in scenarios where the batch size is small for ImageNet training. \AlgName\ is able to remain very performant in low computational environments. A comparative study is done in Appendix-B for epoch time analysis on various setup of network and dataset using different optimizers.

\textbf{Learning rate as regularization.} We emphasize \AlgName's robustness to varying batch sizes, and particularly its high performance levels on low batch sizes, as compared to SGD and SAM who lose performance (e.g. on ADP -- \autoref{mhist_table}, ImageNet -- \autoref{imagenet_table}). We hypothesize that \AlgName's ability to dynamically adjust its learning rate per layer regularizes training in a manner that larger batch sizes normally do \cite{keskar2016large, HofferHS17, l.2018dont, he2019bag}. With larger batch sizes, model gradient updates in SGD yield low variance and using a higher learning rate is possible in order to achieve better performance. However, such scaling is not consistent across optimizer selection. We found that scaling the batch size with SAM can in fact degrades performance. This is similar to what has been reported in \cite{you2017scaling, you2019large, liu2022sharpnessaware} where training stability when using larger batch sizes varies for each layer in a network. This finding led to the development of a new optimizer that used layer-wise learning rate scheduling and demonstrated improved performance in large batch size settings. The concept of RMSGD's per-layer learning rate adjustment, enabled by tracking the per-layer stable rank, also permits such a stablilization behaviour; but works well even when using small batch sizes for training. We argue that this explains why its performance levels remain consistently high in different batch size setups. 


\textbf{Societal impact.} Recent works \cite{fu2021reconsidering, strubell-etal-2019-energy} have highlighted the risks of AI on climate change, and being computationally efficient to  reduce carbon footprint is important. We showed that \AlgName\ is able to run on minimal hardware and achieve superior results, contributing towards this idea.


{\small
\bibliographystyle{ieee_fullname}
\bibliography{citations}
}

\clearpage
\appendix
\onecolumn
\tableofcontents
\clearpage
\section*{Appendix-A: Proof of Theorems and Remarks}\label{sec_appendix_proofs}
\section{Proofs for Theorem 1}
The following two proofs correspond to the proof of Theorem 1 for $p=2$ and $p=1$, respectively. Note we have omitted layer index $\ell$ for convenience.
\begin{proof}{(Theorem 1)}\label{proof_p2}
Recall that the summation of squared singular values of a matrix is equivalent to the {\em Frobenius} (norm) i.e. $||{{\tW}}^{t}||^2_F=\sum^{N^{\prime}_{d}}_{k=1}{\sigma^2_{k}({{\tW}}^{t})}=\trace{{\tW^t}^{T}{\tW}^{t}}$ \cite{horn2012matrix}. Using the definition of stable-rank in Equation (2) from main paper draft, Section 4.1, the stable rank of matrix ${\tW^{t+1}}$ (assumed to be a column matrix $m\leq n$) is expressed by
\begin{equation}
\KG({\tW^{t+1}}) = 
\frac{1}{n\sigma^2_{1}({\tW^{t+1}})}\sum^{n^{\prime}}_{i=1}{\sigma^2_{i}({\tW^{t+1}})} =
\frac{1}{n||{\tW^{t+1}}||^2_2}\trace{{\tW^{t+1}}^T{\tW^{t+1}}}.
\label{eq_RMSGD_4}
\end{equation}
An upper-bound of first singular value can be calculated by first recalling its equivalence to $\ell_2$-norm and then applying the Cauchy–Schwarz inequality
\begin{equation}
\sigma^2_{1}({\tW^{t+1}}) = ||{\tW^{t+1}}||^2_2 = ||{\tW^{t}}-\eta_{\ell}(t){\overline{\nabla{f}}_{t+1}}||^2_2 \leq ||{\tW^{t}}||^2_2 + \eta_{\ell}^2(t)||{\overline{\nabla{f}}_{t+1}}||^2_2 + 2\eta_{\ell}(t)||{\tW^{t}}||_2||{\overline{\nabla{f}}_{t+1}}||_2.
\label{eq_RMSGD_5}
\end{equation}

Note that $\eta_{\ell}(t)$ is given by previous epoch update and considered to be positive (we start with an initial learning rate $\eta(0)>0$). Therefore, the right-hand-side of the inequality in (\ref{eq_RMSGD_5}) will be positive and holds.

By substituting (\ref{eq_RMSGD_5}) in (\ref{eq_RMSGD_4}) and expanding the terms in trace, a lower bound of ${\tW^{t+1}}$ is given by
\begin{equation}
\KG({\tW^{t+1}}) \geq \frac{1}{N\gamma}\left[\trace{{\tW^{t}}^T{\tW^{t}}}-2\eta_{\ell}(t)\trace{{\tW^{t}}^T{\overline{\nabla{f}}_{t+1}}}+\eta_{\ell}^2(t)\trace{{\overline{\nabla{f}}_{t+1}}^T{\overline{\nabla{f}}_{t+1}}}\right],
\label{eq_RMSGD_6}
\end{equation}

where, $\gamma=||{\tW^{t}}||^2_2 + \eta_{\ell}^2(t)||{\overline{\nabla{f}}_{t+1}}||^2_2 + 2\eta_{\ell}(t)||{\tW^{t}}||_2||{\overline{\nabla{f}}_{t+1}}||_2$. The latter inequality can be revised to
\begin{align}
\begin{array}{l}
\KG({\tW^{t+1}})
\geq 
\frac{1}{N\gamma}
\left[\left(1-\frac{\gamma}{||{\tW^{t}}||^2_2}+\frac{\gamma}{||{\tW^{t}}||^2_2}\right)\trace{{\tW^{t}}^T{\tW^{t}}}\right. \\
~~~~~~~~~~~~~~~~~~~~~~~~~~~~~~~~~~~~~~~~~~~~~~~~~~~~~~~~~~~~~~~~~~~~~~~~~~~~~
\left. -2\eta_{\ell}(t)\trace{{\tW^{t}}^T{\overline{\nabla{f}}_{t+1}}}+\eta_{\ell}^2(t)\trace{{\overline{\nabla{f}}_{t+1}}^T{\overline{\nabla{f}}_{t+1}}}\right] \\
~~~~~=\frac{1}{N\gamma}
\left[\frac{\gamma}{||{\tW^{t}}||^2_2}\trace{{\tW^{t}}^T{\tW^{t}}}+\left(1-\frac{\gamma}{||{\tW^{t}}||^2_2}\right)\trace{{\tW^{t}}^T{\tW^{t}}}\right. \\
~~~~~~~~~~~~~~~~~~~~~~~~~~~~~~~~~~~~~~~~~~~~~~~~~~~~~~~~~~~~~~~~~~~~~~~~~~~~~
\left. -2\eta_{\ell}(t)\trace{{\tW^{t}}^T{\overline{\nabla{f}}_{t+1}}}+\eta_{\ell}^2(t)\trace{{\overline{\nabla{f}}_{t+1}}^T{\overline{\nabla{f}}_{t+1}}}\right]\\
~~~~~=\KG({\tW^{t}}) + \frac{1}{N\gamma}
\left[\underbrace{\left(1-\frac{\gamma}{||{\tW^{t}}||^2_2}\right)\trace{{\tW^{t}}^T{\tW^{t}}}-2\eta_{\ell}(t)\trace{{\tW^{t}}^T{\overline{\nabla{f}}_{t+1}}}+\eta_{\ell}^2(t)\trace{{\overline{\nabla{f}}_{t+1}}^T{\overline{\nabla{f}}_{t+1}}}}_{D}\right].
\end{array}
\label{eq_RMSGD_7}
\end{align}
Therefore, the bound in (\ref{eq_RMSGD_7}) is revised to
\begin{equation}
\KG({\tW^{t+1}}) - \KG({\tW^{t}}) \geq \frac{1}{N\gamma}D.
\label{eq_RMSGD_8}
\end{equation}
Since $\gamma\geq 0$, the monotonicity of the Equation (\ref{eq_RMSGD_8}) is guaranteed if $D\geq{0}$. The remaining term $D$ can be expressed as a quadratic function of $\eta$
\begin{align}
\begin{array}{l}
D(\eta) = \left[\trace{{\overline{\nabla{f}}_{t+1}}^T{\overline{\nabla{f}}_{t+1}}}-\frac{||{\overline{\nabla{f}}_{t+1}}||^2_2}{||{\tW^{t}}||^2_2}\trace{{\tW^{t}}^T{\tW^{t}}}\right]\eta_{\ell}^2(t) \\
~~~~~~~~~~~~~~~~~~~~~~~~~~~~~~~~~~~~~~~~~~~ - \left[2\trace{{\tW^{t}}^T{\overline{\nabla{f}}_{t+1}}} + 2\frac{||{\overline{\nabla{f}}_{t+1}}||_2}{||{\tW^{t}}||_2}\trace{{\tW^{t}}^T{\tW^{t}}}\right]\eta_{\ell}(t)
\end{array}
\label{eq_RMSGD_9}
\end{align}
where, the condition for $D(\eta)\geq{0}$ in (\ref{eq_RMSGD_9}) is
\begin{equation}
\eta\geq \max\left\{2\frac{
\trace{{\tW^{t}}^T{\overline{\nabla{f}}_{t+1}}} + \frac{||{\overline{\nabla{f}}_{t+1}}||_2}{||{\tW^{t}}||_2}\trace{{\tW^{t}}^T{\tW^{t}}}
}{
\trace{{\overline{\nabla{f}}_{t+1}}^T{\overline{\nabla{f}}_{t+1}}}-\frac{||{\overline{\nabla{f}}_{t+1}}||^2_2}{||{\tW^{t}}||^2_2}\trace{{\tW^{t}}^T{\tW^{t}}}
},~0\right\}.
\label{eq_RMSGD_10}
\end{equation}
The lower bound in (\ref{eq_RMSGD_10}) proves the existence of a lower bound for monotonicity condition.

Our final inspection is to check if the substitution of step-size (7) in (\ref{eq_RMSGD_8}) would still hold the inequality condition in (\ref{eq_RMSGD_8}). Followed by the substitution, the inequality should satisfy
\begin{equation}
\eta_{\ell}(t) \geq \zeta\frac{1}{N\gamma}D.
\label{eq_RMSGD_11}
\end{equation}
We have found that $D(\eta)\geq{0}$ for some lower bound in (\ref{eq_RMSGD_10}), where the inequality in (\ref{eq_RMSGD_11}) also holds from some $\zeta\geq{0}$ and the proof is done.
\end{proof}

\section{Proof for Proposition 1}\label{appendix_proof_RMSGD_Convergence}
Consider the RMSGD update rule for Stochastic Gradient Descent with Momentum
\begin{align}
\begin{array}{l}
{\bm v}_{\ell}^{k} = \MomentumRate v_{\ell}^{k-1} - \eta_{\ell}(t) {\bm g}_{\ell}^{k}, \\
\Parameters_{\ell}^{k+1} = \Parameters_{\ell}^{k} + {\bm v}_{\ell}^{k}.
\notag
\end{array}
\end{align}
It is evident that $\Parameters_{\ell}^{k} = \Parameters_{\ell}^{k-1} + {\bm v}_{\ell}^{k-1}$ and thus, ${\bm v}_{\ell}^{k-1} = \Parameters_{\ell}^{k} - \Parameters_{\ell}^{k-1}$. We can then write our parameter update rule as 
\begin{align}
\begin{array}{l}
\Parameters_{\ell}^{k+1} = \Parameters_{\ell}^{k} + \MomentumRate(\Parameters_{\ell}^{k} - \Parameters_{\ell}^{k-1}) - \eta_{\ell}(t) {\bm g}_{\ell}^{k} .
\notag
\end{array}
\end{align}
Note how for RMSGD, the parameter update rule is subject to the parametrization of $\GlobalLR_{\ell}(t)$ by $\ell$ (per layer) and $t$ (per epoch).

We highlight the convergence of SGD algorithm is given by the following theorem.

\begin{figure}[th]
    \centering
    \includegraphics[width=0.23\textwidth]{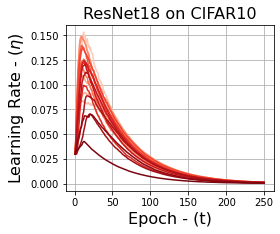}
    \includegraphics[width=0.23\textwidth]{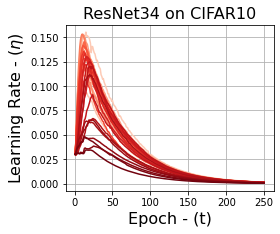}
    \includegraphics[width=0.23\textwidth]{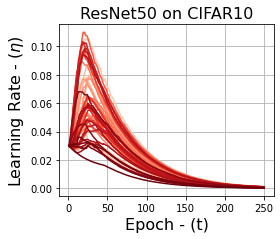}
    \includegraphics[width=0.23\textwidth]{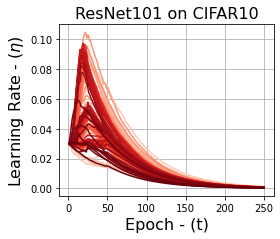}
    \centering
    \includegraphics[width=0.23\textwidth]{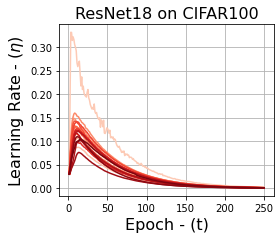}
    \includegraphics[width=0.23\textwidth]{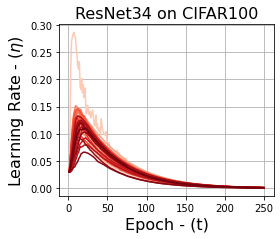}
    \includegraphics[width=0.23\textwidth]{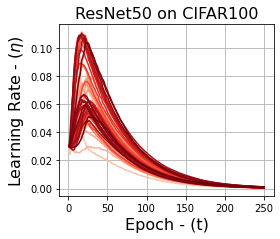}
    \includegraphics[width=0.23\textwidth]{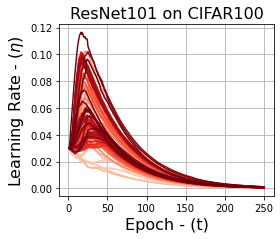}
    \centering
    \includegraphics[width=0.23\textwidth]{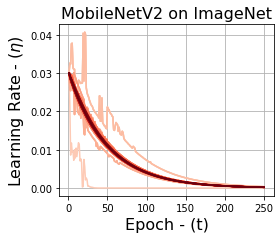}
		\includegraphics[width=0.23\textwidth]{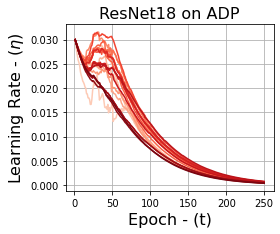}
    \includegraphics[width=0.23\textwidth]{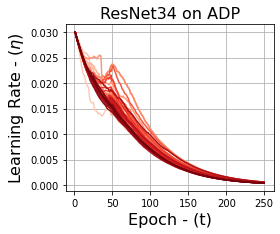}
		\includegraphics[width=0.23\textwidth]{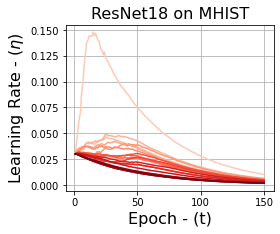}
    \caption{Learning rate evolution over full $250$ of trainig for various ResNets applied on CIFAR10 and CIFAR100, as well as some experiments on ImageNet, ADP, and MHSIT. Increasing darkness of lines indicates increasing layer index for each network. Note that each network starts at $\eta=0.03$, and we used a batch size of $128$ for CIFAR and ImageNet experiments, and $32$ for ADP and MHIST. These learning rates are reported as the average learning rate over $5$ trials, except for ImageNet, it's over $3$ trials.}
    \label{fig_learning_rates_bound}
\end{figure}

\begin{theorem}{(Convergence of Momentum SGD, Theorem 3.2 introduced in \cite{zhou2021convergence} (page 132)).}\label{theorem_mSGD_convergence}
Let the following assumptions hold
\begin{enumerate}
\item The loss function $f$ is convex (Assumption 1.2 in \cite{zhou2021convergence}, page 128) i.e. 
\begin{align}
f({\Parameters^{\prime}}^{k})\geq{f(\Parameters^{k}) + <\nabla{f(\Parameters^{k})}, {\Parameters^{\prime}}^{k} - \Parameters^{k}>},~~\text{for all}~\Parameters^{k},{\Parameters^{\prime}}^{k}
\end{align}
\item The loss function is continuously differentiable as well as the gradient of the loss function is Lipschitz continuous with Liptschitz constant $C>0$ (Assumption 1.3 in \cite{zhou2021convergence}, page 128)
\begin{align}
\|\nabla{f(\Parameters^{k})} - \nabla{f({\Parameters^{\prime}}^{k})}\| \leq C \|\Parameters^{k} - {\Parameters^{\prime}}^{k}\|,~~\text{for all}~\Parameters^{k},{\Parameters^{\prime}}^{k}
\label{eq:Lipschitz}
\end{align}
\item The normalized variance of the gradient of the loss function is bounded by (Assumptions 1.4-1.5 in \cite{zhou2021convergence}, page 128)
\begin{align}
\frac{Var(\hat{{\bm g}}^{k})}{\|{\bm g}^{k}\|^2}\leq M,~~\text{for all}~k\in\mathbb{N}
\label{eq:upper}
\end{align}
where $\hat{{\bm g}}^{k}$ is an unbiased estimate of the gradient of loss function and $M>0$ is a positive scalar.
\end{enumerate}
Furthermore, the step-size of momentum SGD satisfies
\begin{align}
\eta \leq \frac{1-\alpha}{L(M+1)}.
\label{eq:convergence}
\end{align}
Then, the convergence of loss function is guaranteed i.e. 
$\lim_{k\rightarrow\infty}\text{E}[f(\Parameters^{k}) - f(\Parameters^{*})]\rightarrow{0}$.
\end{theorem}
Note the above theorem is deduced from \cite{zhou2021convergence} only for the case of momentum SGD where we ignored the Nestrov accelerated version here i.e. $\beta=0$ in Equation 2.4 in \cite{zhou2021convergence}, page 130. Given RMSGD, where only the learning rates are evolved given different epoch and layer indexes, then the Liptschitz constant in \autoref{eq:Lipschitz} and the upper bound value in \autoref{eq:upper} should be revised by the taking the supremum (least upper bound) across all epochs and layers
\begin{align}
\begin{tabular}{ccc}
$C^{\prime}\leftarrow \sup\limits_{t,\ell}C_{\ell}^{t}$ & \text{and} &
$M^{\prime}\leftarrow \sup\limits_{t,\ell}M_{\ell}^{t}$.
\end{tabular}
\end{align}
The convergence of RMSGD is now guaranteed by revising the upper bound in \autoref{eq:convergence} to satisfy
\begin{align}
\eta_{\ell}(t) \leq \frac{1-\alpha}{C^{\prime}(M^{\prime}+1)}.
\label{eq:RMSGD_convergence}
\end{align}
The condition in \autoref{eq:RMSGD_convergence} induces a higher bound on the learning rates across all epochs and layers. This can be generally satisfied assuming the Liptschitz continuity of gradients in \autoref{eq:Lipschitz} and bounded variance of the gradients  in \autoref{eq:upper} hold. We empirically evaluate this in Figure \ref{fig_learning_rates_bound} on different scenarios given different dataset and network for training. As it shows, the learning rates are fairly bounded and do not explode in practice.

\clearpage
\section*{Appendix-B: Additional Experiments, Hyper-Parameter Tuning, and More}
\section{Learning Momentum ($\beta$) Ablative Study}
\begin{figure}[!h]
    \centering
    \includegraphics[width=0.9\textwidth]{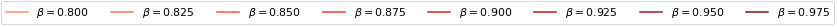}\\
    \centering
    \includegraphics[height=0.23\textwidth]{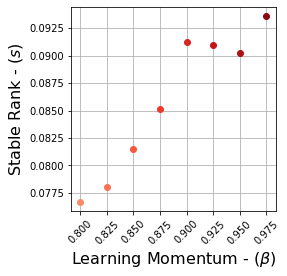}
    \includegraphics[height=0.23\textwidth]{figures/ablative/betas_test_acc_cifar10.png}
    \includegraphics[height=0.23\textwidth]{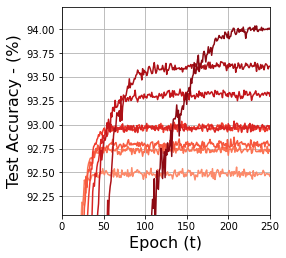}\\
		\centering
		\includegraphics[height=0.23\textwidth]{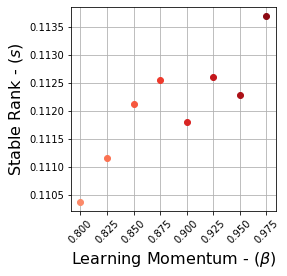}
		\includegraphics[height=0.22\textwidth]{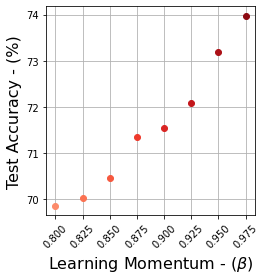}
    \includegraphics[height=0.23\textwidth]{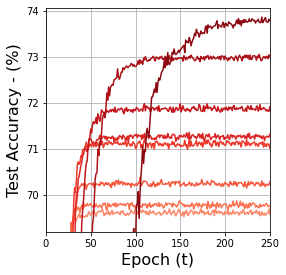}
    \caption{Learning momentum ablative study for VGG16 applied to CIFAR10 (top-row) and CIFAR100 (bottom-row), with a batch size of $128$. We show stable rank behaviour and corresponding test accuracy plots, to visualize the connection between stable rank and resulting performance, since stable rank is utilized in \AlgName's update algorithm. Notice the correlated positive trends with increasing $\beta$.}
    \label{ablative}
\end{figure}

We present the rest of our learning momentum ablative study in Figure \ref{ablative}. Recall that we performed our ablative study using VGG16 on CIFAR10 and CIFAR100, with a batch size of $128$. Note that Figure \ref{ablative} highlights how the $\beta$ parameter may be used to tradeoff between faster convergence speed (lower $\beta$) and better performance (high $\beta$) -- but at the cost of longer convergence speeds.

\section{A Note on Quality Measure Results in the Main Draft}
We provide here some additional details on the experimental details pertaining to Figure 2 in the main draft, which shows our quality measure on CIFAR10/CIFAR100. Specifically, each scatter point represents one of AdaBound, AdaGrad, Adam, AdamP, SAM, SLS, SGD, SGDP, or \AlgName\ on ResNet18, ResNet34, ResNet50 or ResNet101. Naturally, each permutation of these networks and optimizers are then applied to CIFAR10 and CIFAR100. We used a batch size of $128$. For hyper-parameter details, see \autoref{image_hp}.


\section{Experiments}
\subsection{Hardware}
In total, we had a server with $4$ RTX2080Ti GPUS, an Intel Xeon Gold 6246 processor, and $256$ gigabytes of RAM available for all experiments. All experiments reported here were performed on an instance with access to a single Nvidia RTX2080Ti GPU, 4 cores of the Intel Xeon Gold 6246 processor, and $64$ gigabytes of RAM. No experiments were performed using any GPU parallelism.

\subsection{Image Classification}
We present additional results and experimental details from image classification experiments here. 

We perform $5$ randomly initialized trials of each experiment except ImageNet, for which we performed $3$ trials, and we report the mean and standard deviation of results.

\textbf{Note on training time.} Since different optimizers exhibit widely different epoch times, for a fair comparison, we limit training to the total wall clock time consumed by $250$ epochs using SGD. In terms of epochs, this amounts to $\sim250$ epochs for all optimizers except SAM, which consumes only $\sim128-133$ epochs due to its $2$ forward passes. 

\subsubsection{Hyper-Parameters}
\label{image_hp}
For all image classification tasks, we tuned each optimizer on ResNet18 applied to CIFAR10, with a batch size of $128$. We used a step decay method of step size $25$ and gamma $0.5$ for all optimizers, except \AlgName\ (who had no learning rate decay). We found this setup to be optimal, as also reported in \cite{wilson2017marginal}. We used a consistent weight decay of $\num{5e-4}$ for all optimizers, momentum rate of $0.9$ for SGD, SGDP, SAM, and \AlgName, and kept all other hyper-parameters as default. For learning rates, for \AlgName\ we consider a grid search over,  $\{0.01, 0.02, 0.025, \mathbf{0.03}, 0.035, 0.04, 0.05\}$, with the selected learning rate bold. For AdaBelief, AdamP, SLS, SAM, and SGDP, we follow the CIFAR-specific hyper-parameter search ranges presented in their original work. For Adam, AdaGrad, and RMSProp, we followed the ranges presented in \cite{wilson2017marginal} and confirmed in our experiments that they yield optimal performance over the author-suggested values. For SGD, we considered a range of $\{0.05, 0.08, 0.09, \mathbf{0.1}, 0.11, 0.12\}$. We tabulate the final selected learning rate for each method in  \ref{learning_rate_grids}. In general, the selected learning rate results to the suggested learning rate provided in each optimizer's original work. We highlight once again that these learning rates and hyper-parameters were carried forward for all other experiments, including all other networks on CIFAR10/CIFAR100, cutout experiments on CIFAR10/CIFAR100, ImageNet, and both histopathology pathology datasets.
\begin{table}[!h]
    \centering
    \caption{Final selected learning rates for each optimizer, tuned using ResNet18 on CIFAR10 using a batch size of $128$. We tuned by completing a full $250$ epoch training cycle, and selected based on final validation top-1 accuracy.}
    \begin{tabular}{c|c|c|c|c|c|c|c|c}
        AdaBound&AdaGrad&Adam&AdamP&SLS&SAM&SGD&SGDP&\AlgName\\
        \hline
        $0.01$&$0.01$&$0.0003$&$0.01$&$1.0$&$1.0$&$0.1$&$0.1$&$0.03$
    \end{tabular}
    \label{learning_rate_grids}
\end{table}

\subsubsection{Dataset Details}

\textbf{CIFAR.} We consider a batch size of $128$. For CIFAR-related experiments, we perform $32\times32$ random-resize cropping and random horizontal flipping as data augmentations.

\textbf{ImageNet.} We perform $3$ randomly initialized trials and report the mean and standard deviation of results. We consider a batch size of $128$, as we performed experiments on a single GPU instance without any GPU parallelism. We follow \cite{he2015deep} and perform random resized cropping to $224\times244$ and random horizontal flipping on the train set and $256\times256$ resizing with $224\times224$ center cropping on the test set.

\textbf{Histopathology Datasets}. The histopathlogy datasets provide a study of optimizers on a low class and small sized dataset, with MHIST having $2,175$ training examples across binary class with $224\times 224$ image size, and ADP having $14,134$ examples across $22$ multi-labeled classes with $272\times 272$ image size. We consider a batch size of $32$ for both datasets following the recommendations in \cite{wei2021petri, hosseini2019atlas}. We consider random cropping $224\times224$ and random flipping as data augmentations for MHIST, and only random flipping for ADP. See MHIST \cite{wei2021petri} and ADP \cite{hosseini2019atlas} references for instructions on how to download the datasets.

\subsubsection{Results}
\begin{figure}[!ht]
    \centering
    \includegraphics[width=0.7\textwidth]{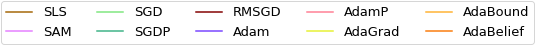}\\
    \centering
		\includegraphics[width=0.2\textwidth]{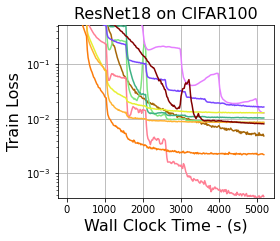}
		\includegraphics[width=0.2\textwidth]{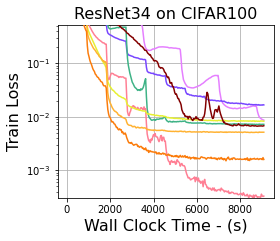}
    \includegraphics[width=0.2\textwidth]{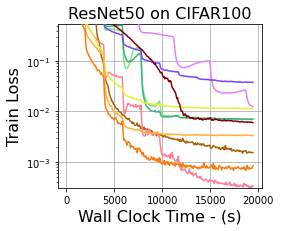}
    \includegraphics[width=0.2\textwidth]{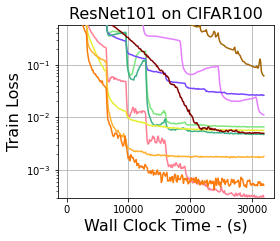}
    \includegraphics[width=0.2\textwidth]{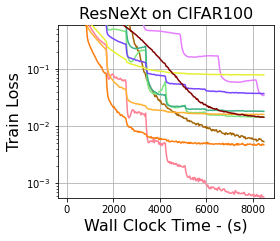}
    \centering
    \includegraphics[width=0.2\textwidth]{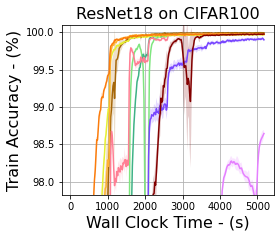}
    \includegraphics[width=0.2\textwidth]{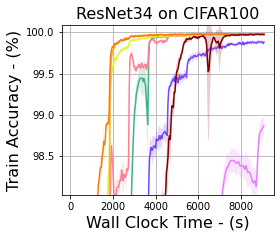}
    \includegraphics[width=0.2\textwidth]{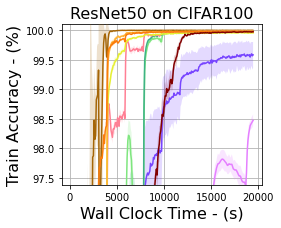}
    \includegraphics[width=0.2\textwidth]{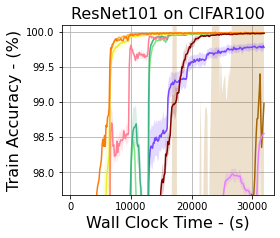}
    \includegraphics[width=0.2\textwidth]{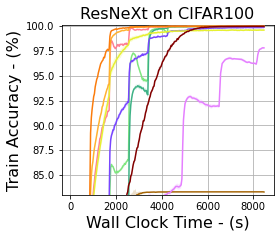}
    \centering
		\includegraphics[width=0.2\textwidth]{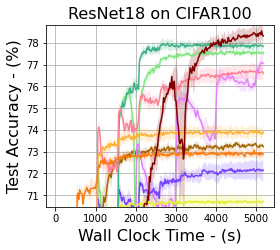}
    \includegraphics[width=0.2\textwidth]{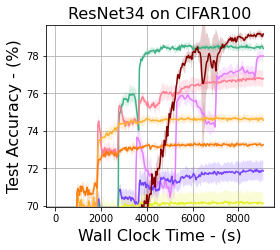}
    \includegraphics[width=0.2\textwidth]{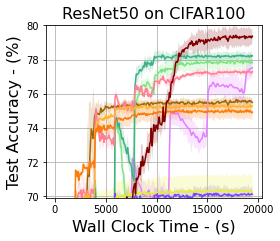}
    \includegraphics[width=0.2\textwidth]{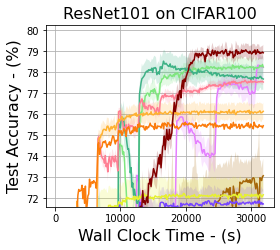}
    \includegraphics[width=0.2\textwidth]{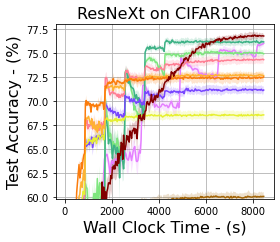}    
    \caption{Train accuracy, test accuracy, and train loss for ResNets and ResNeXt on CIFAR100 experiments, without Cutout. A batch size of $128$ was used, and we report the mean over $5$ trials for each experiment, with translucent bands to indicate the standard deviation of each experiment. All networks were tuned using ResNet18 applied on CIFAR10.}
    \label{cifar_resnets_c100}
\end{figure}

\begin{figure}[!ht]
    \centering
    \includegraphics[width=0.7\textwidth]{figures/cifar-supp/legend.png}\\
    \centering
    \includegraphics[width=0.2\textwidth]{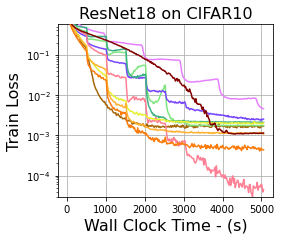}
    \includegraphics[width=0.2\textwidth]{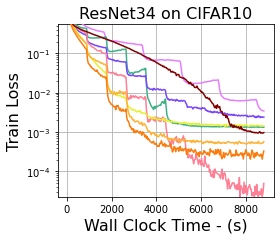}
    \includegraphics[width=0.2\textwidth]{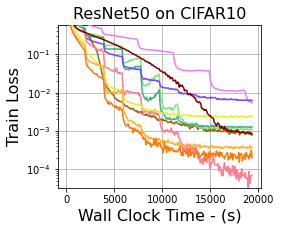}
    \includegraphics[width=0.2\textwidth]{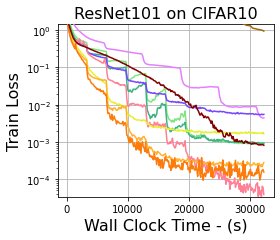}
    \includegraphics[width=0.2\textwidth]{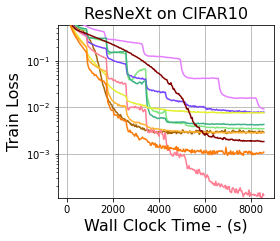}
    \centering
		\includegraphics[width=0.2\textwidth]{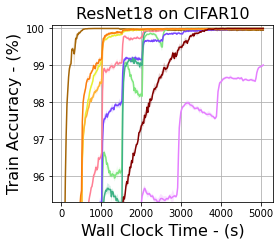}
    \includegraphics[width=0.2\textwidth]{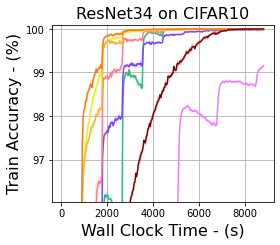}
    \includegraphics[width=0.2\textwidth]{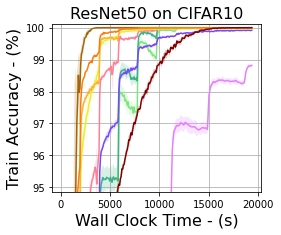}
    \includegraphics[width=0.2\textwidth]{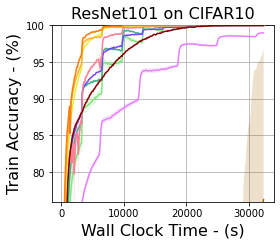}
    \includegraphics[width=0.2\textwidth]{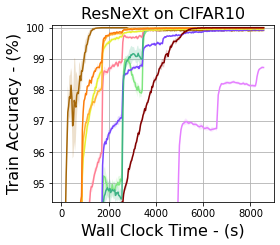}
    \centering
		\includegraphics[width=0.2\textwidth]{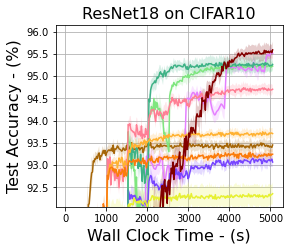}
    \includegraphics[width=0.2\textwidth]{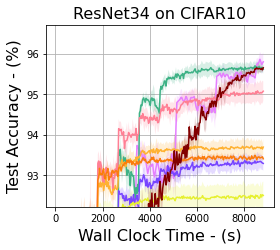}
    \includegraphics[width=0.2\textwidth]{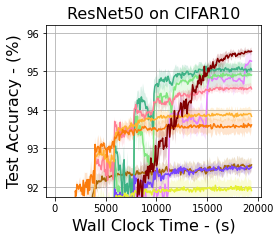}
    \includegraphics[width=0.2\textwidth]{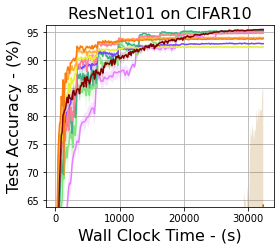}
    \includegraphics[width=0.2\textwidth]{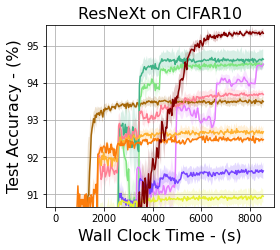}
    \caption{Train accuracy, test accuracy, and train loss for ResNets and ResNeXt on CIFAR10 experiments, without Cutout. A batch size of $128$ was used, and we report the mean over $5$ trials for each experiment, with translucent bands to indicate the standard deviation of each experiment. All networks were tuned using ResNet18 applied on CIFAR10.}
    \label{cifar_resnets_c10}
\end{figure}

\begin{figure}[!ht]
    \centering
    \includegraphics[width=0.7\textwidth]{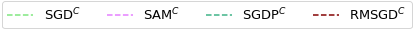}\\
    \centering
		\includegraphics[width=0.2\textwidth]{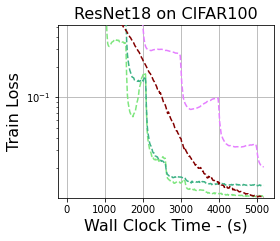}
    \includegraphics[width=0.2\textwidth]{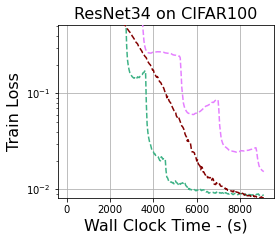}
    \includegraphics[width=0.2\textwidth]{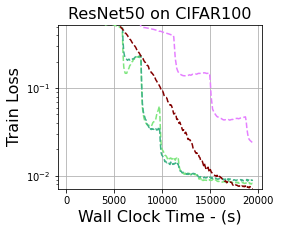}
    \includegraphics[width=0.2\textwidth]{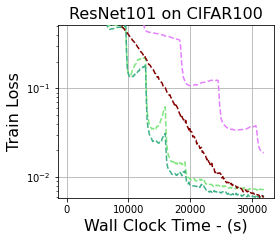}
    \includegraphics[width=0.2\textwidth]{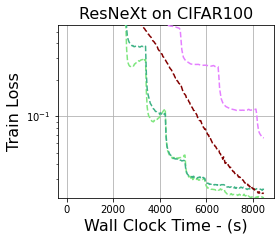}
    \centering
		\includegraphics[width=0.2\textwidth]{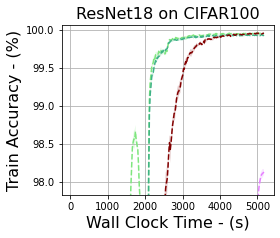}
    \includegraphics[width=0.2\textwidth]{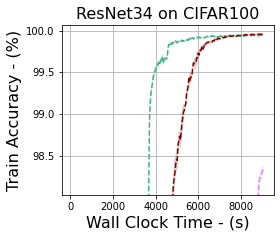}
    \includegraphics[width=0.2\textwidth]{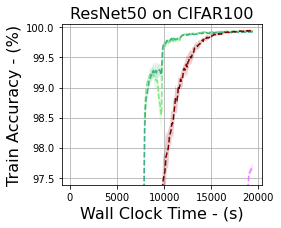}
    \includegraphics[width=0.2\textwidth]{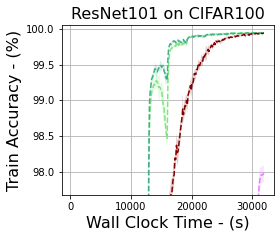}
    \includegraphics[width=0.2\textwidth]{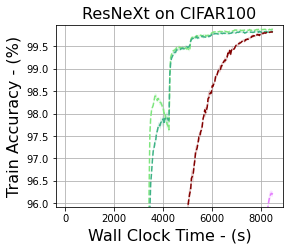}
    \centering
		\includegraphics[width=0.2\textwidth]{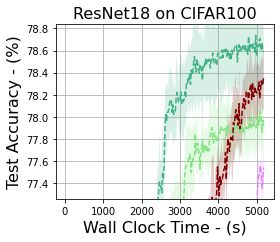}
    \includegraphics[width=0.2\textwidth]{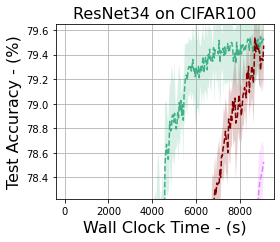}
    \includegraphics[width=0.2\textwidth]{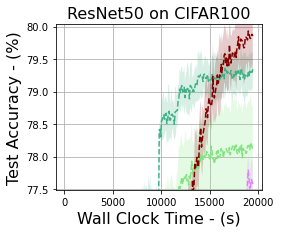}
    \includegraphics[width=0.2\textwidth]{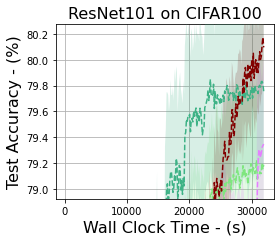}
    \includegraphics[width=0.2\textwidth]{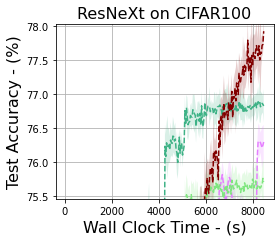}
    \caption{Train accuracy, test accuracy, and train loss for ResNets and ResNeXt on CIFAR100 experiments, with Cutout. A batch size of $128$ was used, and we report the mean over $5$ trials for each experiment, with translucent bands to indicate the standard deviation of each experiment. All networks were tuned using ResNet18 applied on CIFAR10.}
    \label{cifar_resnets_c100_cutout}
\end{figure}

\begin{figure}[!ht]
    \centering
    \includegraphics[width=0.7\textwidth]{figures/cifar-supp/cutout_legend.png}\\
    \centering
		\includegraphics[width=0.2\textwidth]{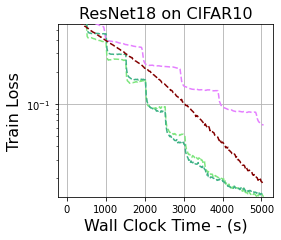}
    \includegraphics[width=0.2\textwidth]{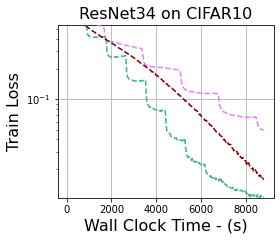}
    \includegraphics[width=0.2\textwidth]{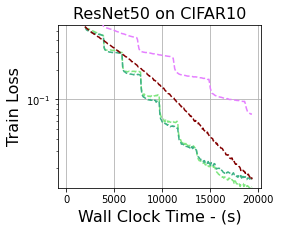}
    \includegraphics[width=0.2\textwidth]{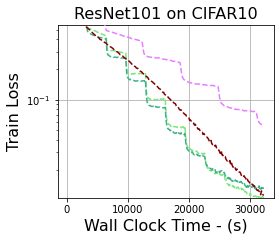}
    \includegraphics[width=0.2\textwidth]{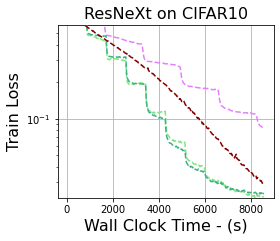}
    \centering
		\includegraphics[width=0.2\textwidth]{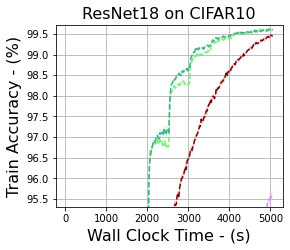}
    \includegraphics[width=0.2\textwidth]{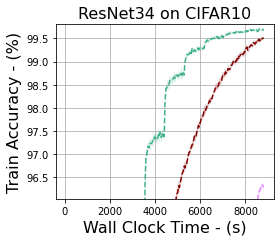}
    \includegraphics[width=0.2\textwidth]{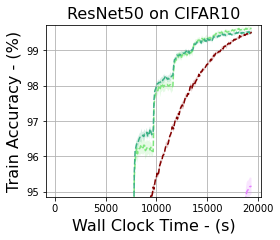}
    \includegraphics[width=0.2\textwidth]{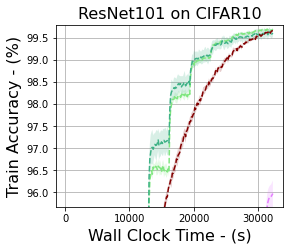}
    \includegraphics[width=0.2\textwidth]{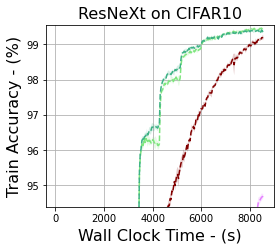}
    \centering
		\includegraphics[width=0.2\textwidth]{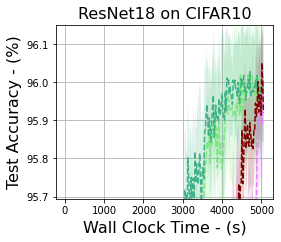}
    \includegraphics[width=0.2\textwidth]{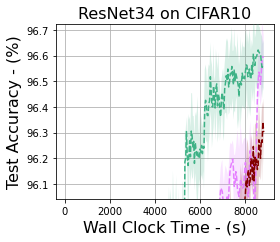}
    \includegraphics[width=0.2\textwidth]{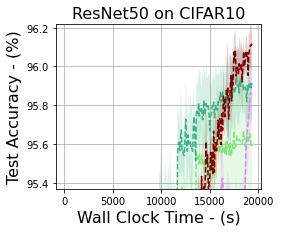}
    \includegraphics[width=0.2\textwidth]{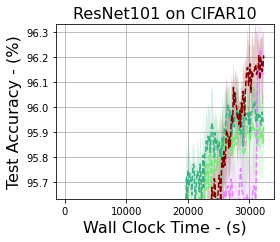}
    \includegraphics[width=0.2\textwidth]{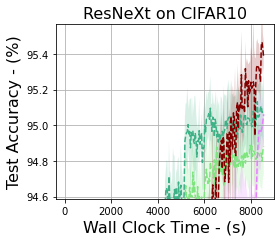}
    \caption{Train accuracy, test accuracy, and train loss for ResNets and ResNeXt on CIFAR10 experiments, with Cutout. A batch size of $128$ was used, and we report the mean over $5$ trials for each experiment, with translucent bands to indicate the standard deviation of each experiment. All networks were tuned using ResNet18 applied on CIFAR10.}
    \label{cifar_resnets_c10_cutout}
\end{figure}

\begin{figure}[!ht]
    \centering
    \includegraphics[width=0.7\textwidth]{figures/cifar-supp/cutout_legend.png}\\
    \centering
		\includegraphics[width=0.2\textwidth]{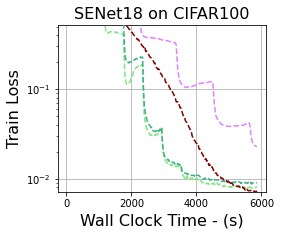}
    \includegraphics[width=0.2\textwidth]{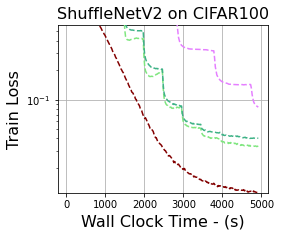}
    \includegraphics[width=0.2\textwidth]{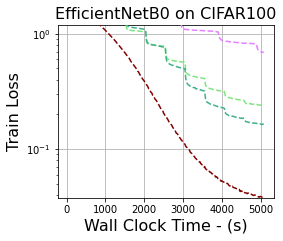}
    \includegraphics[width=0.2\textwidth]{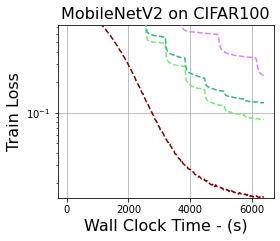}
    \centering
		\includegraphics[width=0.2\textwidth]{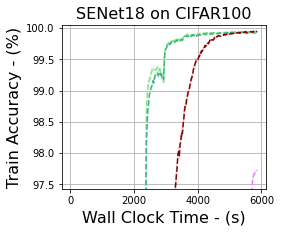}
    \includegraphics[width=0.2\textwidth]{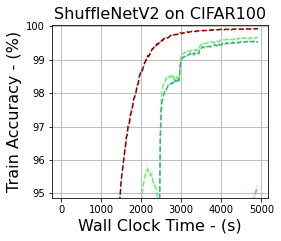}
    \includegraphics[width=0.2\textwidth]{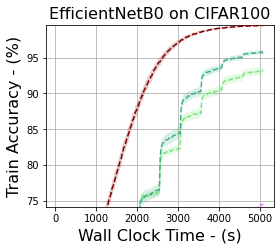}
    \includegraphics[width=0.2\textwidth]{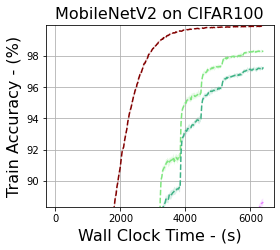}
    \centering
		\includegraphics[width=0.2\textwidth]{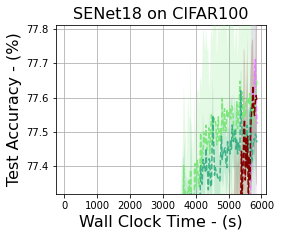}
    \includegraphics[width=0.2\textwidth]{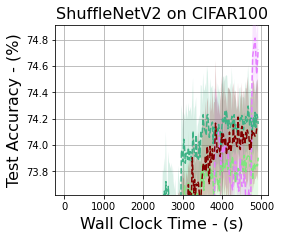}
    \includegraphics[width=0.2\textwidth]{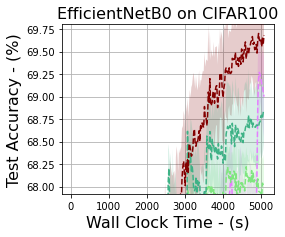}
    \includegraphics[width=0.2\textwidth]{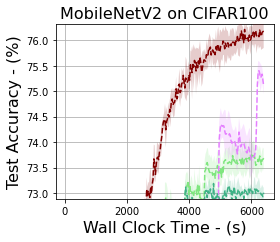}
    \caption{Train accuracy, test accuracy, and train loss for other networks (SeNet18, ShuffleNetV2, EfficientNetB0, MobileNetv2) on CIFAR100 experiments. These plots include only cutout experiments. A batch size of $128$ was used, and we report the mean over $5$ trials for each experiment, with translucent bands to indicate the standard deviation of each experiment. All networks were tuned using ResNet18 applied on CIFAR10.}
    \label{cifar_other_c100}
\end{figure}

\begin{figure}[!ht]
    \centering
    \includegraphics[width=0.7\textwidth]{figures/cifar-supp/cutout_legend.png}\\
    \centering
		\includegraphics[width=0.2\textwidth]{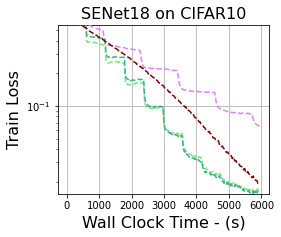}
    \includegraphics[width=0.2\textwidth]{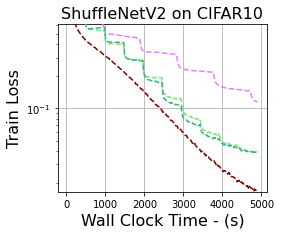}
    \includegraphics[width=0.2\textwidth]{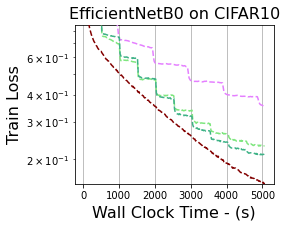}
    \includegraphics[width=0.2\textwidth]{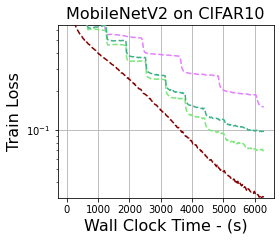}
    \centering
		\includegraphics[width=0.2\textwidth]{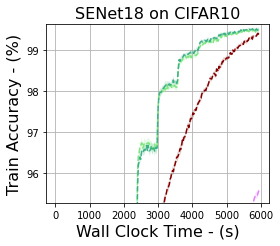}
    \includegraphics[width=0.2\textwidth]{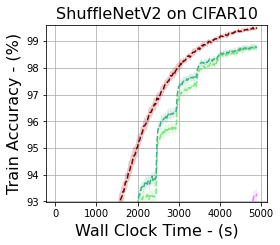}
    \includegraphics[width=0.2\textwidth]{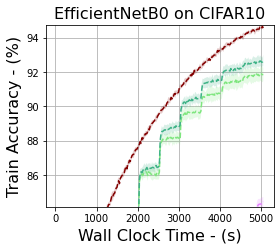}
    \includegraphics[width=0.2\textwidth]{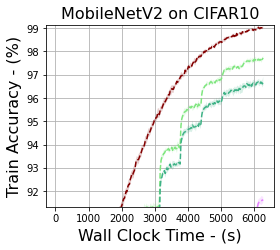}
    \centering
		\includegraphics[width=0.2\textwidth]{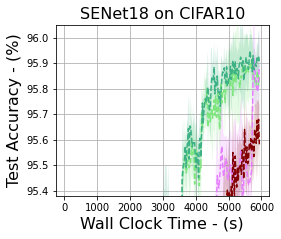}
    \includegraphics[width=0.2\textwidth]{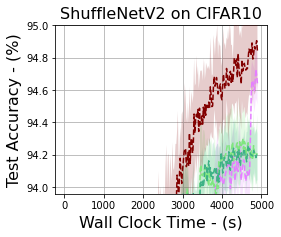}
    \includegraphics[width=0.2\textwidth]{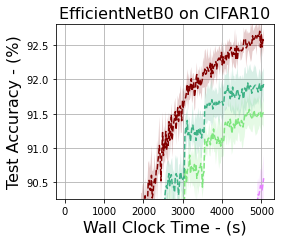}
    \includegraphics[width=0.2\textwidth]{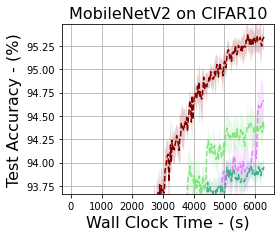}
    \caption{Train accuracy, test accuracy, and train loss for other networks (SeNet18, ShuffleNetV2, EfficientNetB0, MobileNetv2) on CIFAR10 experiments. These plots include only cutout experiments. A batch size of $128$ was used, and we report the mean over $5$ trials for each experiment, with translucent bands to indicate the standard deviation of each experiment. All networks were tuned using ResNet18 applied on CIFAR10.}
    \label{cifar_other_c10}
\end{figure}

Figures \ref{cifar_resnets_c100}, \ref{cifar_resnets_c10}, \ref{cifar_resnets_c100_cutout}, \ref{cifar_resnets_c10_cutout}, \ref{cifar_other_c10}, \ref{cifar_other_c100} visualize the train loss, test accuracy, and train accuracy for CIFAR experiments. Table \ref{cifar_no_cutout} shows the final epoch results of CIFAR experiments without cutout, and Table \ref{cifar_cutout} shows the final epoch results of CIFAR experiments with cutout. Note that SAM here has consumed twice as many seconds to train, and unfortunately we do not have the analogous $500$-epoch training results for the other optimizer, which makes it an unfair comparison to the other optimizers.

\begin{figure}
    \centering
    \includegraphics[width=0.7\textwidth]{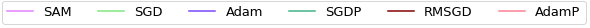}\\
    \centering
		\includegraphics[width=0.24\textwidth]{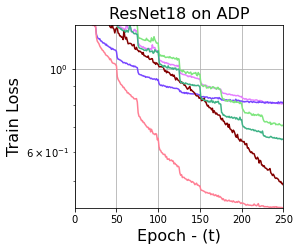}
    \includegraphics[width=0.24\textwidth]{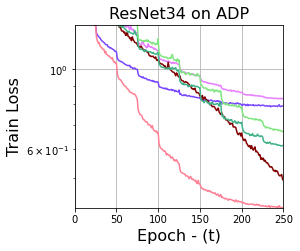}
    \includegraphics[width=0.24\textwidth]{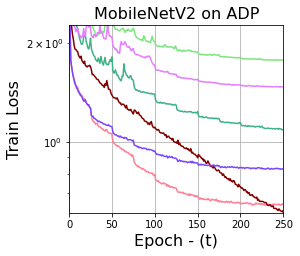}
    \centering
		\includegraphics[width=0.23\textwidth]{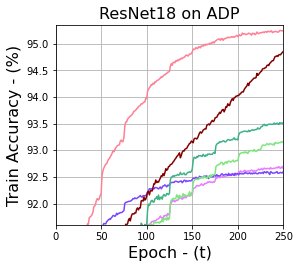}
    \includegraphics[width=0.23\textwidth]{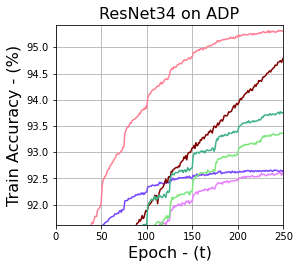}
    \includegraphics[width=0.23\textwidth]{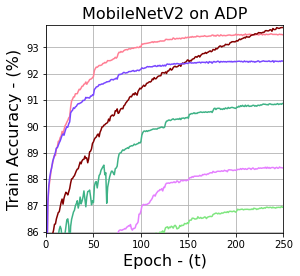}
    \centering
		\includegraphics[width=0.23\textwidth]{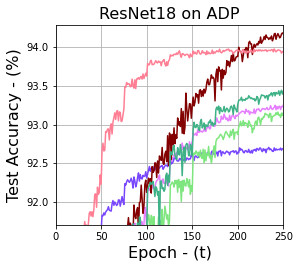}
    \includegraphics[width=0.23\textwidth]{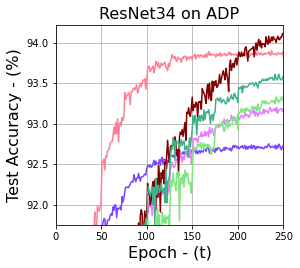}
    \includegraphics[width=0.23\textwidth]{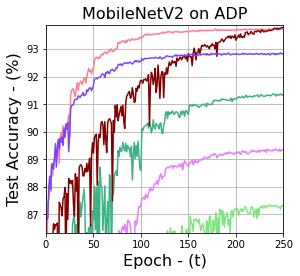}
    \caption{Train accuracy, test accuracy, and train loss for ResNet18, ResNet34, and MobileNetV2 on ADP. Each experiment was run with a batch size of $32$, and we report the mean over $5$ trials. All networks were tuned using ResNet18 applied on CIFAR10.}
    \label{adp_figures}
\end{figure}

\begin{figure}
    \centering
    \includegraphics[width=0.7\textwidth]{figures/mhist/legend.png}\\
    \centering
		\includegraphics[width=0.23\textwidth]{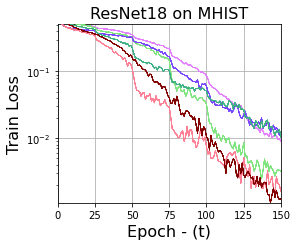}
    \includegraphics[width=0.23\textwidth]{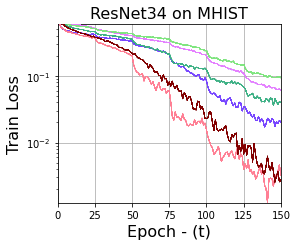}
    \includegraphics[width=0.23\textwidth]{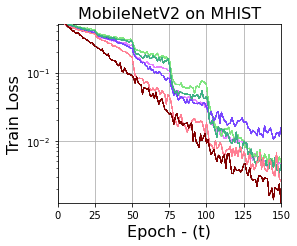}
    \centering
		\includegraphics[width=0.23\textwidth]{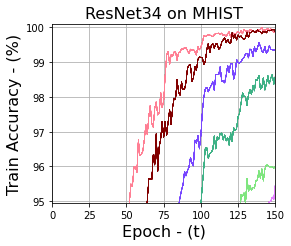}
    \includegraphics[width=0.23\textwidth]{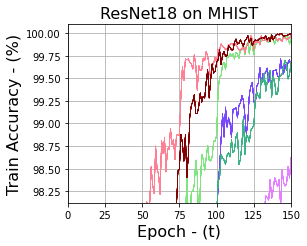}
    \includegraphics[width=0.23\textwidth]{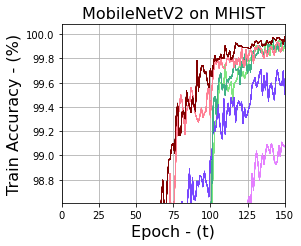}
    \centering
		\includegraphics[width=0.23\textwidth]{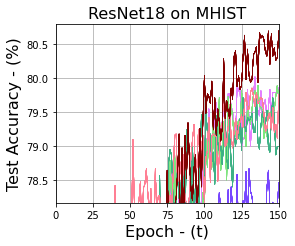}
    \includegraphics[width=0.23\textwidth]{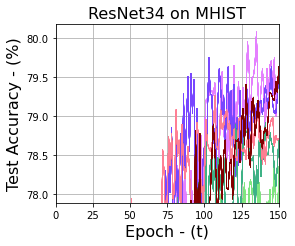}
    \includegraphics[width=0.23\textwidth]{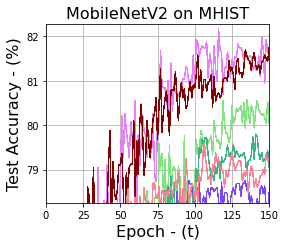}
    \caption{Train accuracy, test accuracy, and train loss for ResNet18, ResNet34, and MobileNetV2 on MHIST. Each experiment was run with a batch size of $32$, and we report the mean over $5$ trials. All networks were tuned using ResNet18 applied on CIFAR10.}
    \label{mhist_figures}
\end{figure}

Figure \ref{mhist_figures} visualizes the test and train accuracy, and the train loss for MHIST experiments. Figure \ref{adp_figures} shows the test and train accuracy, and the train loss for ADP experiments. Table \ref{mhist_table} tabulates the test area under the curve (AUC) results for MHIST experiments. 

\clearpage
\begin{table}[t]
    \centering
    \caption{Test area under the curve (AUC) results for experiments on MHIST. We reported the mean and standard deviation over $5$ trials. We used a batch size of $32$.}
    \tiny
    \begin{tabular}{c|c|c|c|c|c||c}
         Network&Adam&AdamP&SAM&SGD&SGDP&\AlgName  \\
         \specialrule{2pt}{1pt}{1pt}
         ResNet18&{\color{orange}${86.77_{0.23}}$}&{\color{orange}${87.33_{1.05}}$}&{\color{orange}${88.26_{1.43}}$}&{\color{orange}${87.04_{1.75}}$}&{\color{orange}${87.08_{0.53}}$}&{\color{best}${88.51_{0.63}}$}\\
         ResNet34&{\color{best}${87.90_{0.44}}$}&{\color{orange}${87.50_{0.65}}$}&{\color{orange}${87.79_{1.22}}$}&{\color{orange}${86.57_{1.98}}$}&{\color{orange}${86.47_{0.77}}$}&{\color{orange}${87.32_{0.98}}$}\\
         MobileNetV2&{\color{orange}${86.58_{0.83}}$}&{\color{orange}${86.77_{0.54}}$}&{\color{best}${89.34_{0.50}}$}&{\color{orange}${87.48_{0.63}}$}&{\color{orange}${87.47_{2.05}}$}&{\color{orange}${88.88_{0.57}}$}\\
    \end{tabular}
    \label{mhist_table}
\end{table}

\begin{table}[!h]
    \setlength{\tabcolsep}{2pt}
    \centering
    \caption{Perfomance of various networks and optimizers on CIFAR10 and CIFAR100 without Cutout. Results reported for $250$ epochs of training, rather than wall clock time. Note that SAM consumes twice as much time to train compared to all other optimizers. The best result is highlighted in {\color{best}green}, and for \AlgName\ results, {\color{close}orange} highlights when the results lie with the standard deviation from the best. We used a batch size of $128$, and all networks were tuned using ResNet18 on CIFAR10.}
    \tiny
    \begin{tabular}{c|c|c|c|c|c|c|c|c|c||c}
    Dataset&Network&AdaBound&AdaGrad&Adam&AdamP&SLS&SAM&SGD&SGDP&\AlgName\\
    \specialrule{2pt}{1pt}{1pt}
    \multirow{4}{*}{CIFAR10}
    &ResNet18&$93.84_{0.09}$&$92.45_{0.24}$&$93.27_{0.10}$&$94.82_{0.10}$&$93.62_{0.10}$&${\color{best}\mathbf{95.98_{0.07}}}$&$95.32_{0.07}$&$95.39_{0.16}$&$95.66_{0.17}$\\
    &ResNet34&$93.79_{0.19}$&$92.59_{0.30}$&$93.47_{0.18}$&$95.14_{0.25}$&$93.45_{0.16}$&${\color{best}\mathbf{96.34_{0.16}}}$&$95.56_{0.10}$&$95.75_{0.14}$&$95.71_{0.07}$\\
    &ResNet50&$94.00_{0.15}$&$92.12_{0.23}$&$92.67_{0.12}$&$94.69_{0.10}$&$92.70_{0.18}$&${\color{best}\mathbf{95.80_{0.18}}}$&$95.05_{0.28}$&$95.19_{0.15}$&${\color{close}\mathbf{95.63_{0.05}}}$\\
    &ResNet101&$94.17_{0.13}$&$92.51_{0.22}$&$93.13_{0.08}$&$94.92_{0.24}$&$64.20_{20.98}$&${\color{best}\mathbf{96.07_{0.12}}}$&$95.30_{0.13}$&$95.36_{0.04}$&$95.53_{0.14}$\\
    &ResNeXt&$92.83_{0.14}$&$91.09_{0.19}$&$91.78_{0.16}$&$93.82_{0.10}$&$93.67_{0.09}$&$95.01_{0.09}$&$94.62_{0.09}$&$94.79_{0.24}$&${\color{best}\mathbf{95.49_{0.05}}}$\\
    \hline
    \hline
    \multirow{5}{*}{CIFAR100}
    &ResNet18&$74.09_{0.27}$&$70.92_{0.31}$&$72.45_{0.34}$&$76.81_{0.31}$&$73.59_{0.04}$&$77.82_{0.25}$&$77.80_{0.07}$&${\color{close}\mathbf{78.13_{0.16}}}$&${\color{best}\mathbf{78.63_{0.34}}}$\\
    &ResNet34&$74.84_{0.18}$&$70.39_{0.57}$&$72.09_{0.50}$&$76.93_{0.40}$&$73.22_{0.11}$&${\color{close}\mathbf{79.22_{0.39}}}$&$77.88_{0.39}$&$78.74_{0.12}$&${\color{best}\mathbf{79.32_{0.10}}}$\\
    &ResNet50&$75.52_{0.37}$&$70.60_{0.91}$&$70.53_{0.36}$&$77.47_{0.16}$&$75.80_{0.23}$&${\color{close}\mathbf{78.85_{0.66}}}$&$78.12_{0.42}$&$78.44_{0.24}$&${\color{best}\mathbf{79.59_{0.54}}}$\\
    &ResNet101&$76.31_{0.41}$&$72.39_{0.84}$&$72.20_{0.68}$&$77.71_{0.16}$&$73.31_{0.84}$&${\color{best}\mathbf{79.71_{0.48}}}$&$78.48_{0.45}$&$78.60_{0.55}$&${\color{close}\mathbf{79.36_{0.26}}}$\\
    &ResNeXt&$72.97_{0.38}$&$68.83_{0.43}$&$71.54_{0.41}$&$74.54_{0.40}$&$72.35_{0.42}$&${\color{close}\mathbf{76.82_{0.30}}}$&$75.36_{0.33}$&${\color{close}\mathbf{76.56_{0.33}}}$&${\color{best}\mathbf{77.14_{0.31}}}$\\
    \end{tabular}
    \label{cifar_no_cutout}
\end{table}

\begin{table}[!h]
    \setlength{\tabcolsep}{2pt}
    \centering
    \caption{Perfomance of various networks and optimizers on CIFAR10 and CIFAR100 with Cutout. Results reported for $250$ epochs of training, rather than wall clock time. Note that SAM consumes twice as much time to train compared to all other optimizers. The best result is highlighted in {\color{best}green}, and for \AlgName\ results, {\color{close}orange} highlights when the results lie with the standard deviation from the best. We used a batch size of $128$, and all networks were tuned using ResNet18 on CIFAR10.}
    \tiny
    \begin{tabular}{c|c|c|c|c||c|c|c|c}
    &\multicolumn{4}{c||}{CIFAR10}&\multicolumn{4}{c}{CIFAR100}\\
    \cline{2-9}
    Network&SAM\textsuperscript{C}&SGD\textsuperscript{C}&SGDP\textsuperscript{C}&\AlgName\textsuperscript{C}&SAM\textsuperscript{C}&SGD\textsuperscript{C}&SGDP\textsuperscript{C}&\AlgName\textsuperscript{C}\\
    \specialrule{2pt}{1pt}{1pt}
    ResNet18&${\color{best}\mathbf{96.44_{0.13}}}$&$96.12_{0.13}$&$96.13_{0.13}$&$96.13_{0.08}$&${\color{close}\mathbf{78.77_{0.11}}}$&$78.16_{0.21}$&${\color{best}\mathbf{78.82_{0.37}}}$&${\color{close}\mathbf{78.53_{0.22}}}$\\
    ResNet34&${\color{best}\mathbf{97.10_{0.09}}}$&$96.53_{0.13}$&$96.70_{0.10}$&$96.42_{0.08}$&${\color{best}\mathbf{79.85_{0.19}}}$&$78.63_{0.55}$&${\color{close}\mathbf{79.67_{0.24}}}$&${\color{close}\mathbf{79.70_{0.19}}}$\\
    ResNet50&${\color{best}\mathbf{96.49_{0.10}}}$&$95.78_{0.27}$&$96.03_{0.16}$&$96.28_{0.07}$&$79.26_{0.28}$&$78.36_{0.67}$&${\color{close}\mathbf{79.52_{0.31}}}$&${\color{best}\mathbf{80.06_{0.45}}}$\\
    ResNet101&${\color{best}\mathbf{96.79_{0.08}}}$&$96.04_{0.16}$&$96.12_{0.05}$&$96.33_{0.08}$&${\color{best}\mathbf{80.82_{0.66}}}$&$79.35_{0.62}$&${\color{close}\mathbf{80.03_{0.67}}}$&${\color{close}\mathbf{80.36_{0.35}}}$\\
    ResNeXt&${\color{best}\mathbf{95.67_{0.18}}}$&$95.04_{0.18}$&$95.24_{0.15}$&${\color{close}\mathbf{95.62_{0.08}}}$&$77.41_{0.06}$&$75.91_{0.19}$&$77.14_{0.21}$&${\color{best}\mathbf{78.06_{0.28}}}$\\
    MobileNetV2&${\color{best}\mathbf{95.60_{0.12}}}$&$94.53_{0.16}$&$94.07_{0.07}$&${\color{close}\mathbf{95.48_{0.11}}}$&${\color{best}\mathbf{76.53_{0.18}}}$&$73.94_{0.21}$&$73.40_{0.07}$&${\color{close}\mathbf{76.36_{0.27}}}$\\
    SENet18&${\color{best}\mathbf{96.44_{0.16}}}$&$95.99_{0.10}$&$96.04_{0.08}$&$95.80_{0.06}$&${\color{best}\mathbf{78.85_{0.32}}}$&$77.80_{0.23}$&$77.70_{0.05}$&$77.77_{0.15}$\\
    EfficientNetB0&$91.83_{0.23}$&$91.70_{0.24}$&$92.09_{0.21}$&${\color{best}\mathbf{92.83_{0.14}}}$&${\color{best}\mathbf{70.47_{0.50}}}$&$68.42_{0.24}$&$68.98_{0.44}$&${\color{close}\mathbf{69.87_{0.49}}}$\\
    ShuffleNetV2&${\color{best}\mathbf{95.39_{0.15}}}$&$94.40_{0.21}$&$94.37_{0.12}$&${\color{close}\mathbf{95.01_{0.29}}}$&${\color{best}\mathbf{75.88_{0.19}}}$&$74.13_{0.35}$&$74.44_{0.29}$&$74.38_{0.36}$\\
    \end{tabular}
    \label{cifar_cutout}
\end{table}

\clearpage
\subsection{Epoch Times}
\begin{figure}[h]
    \centering{

\includegraphics[width=0.23\textwidth]{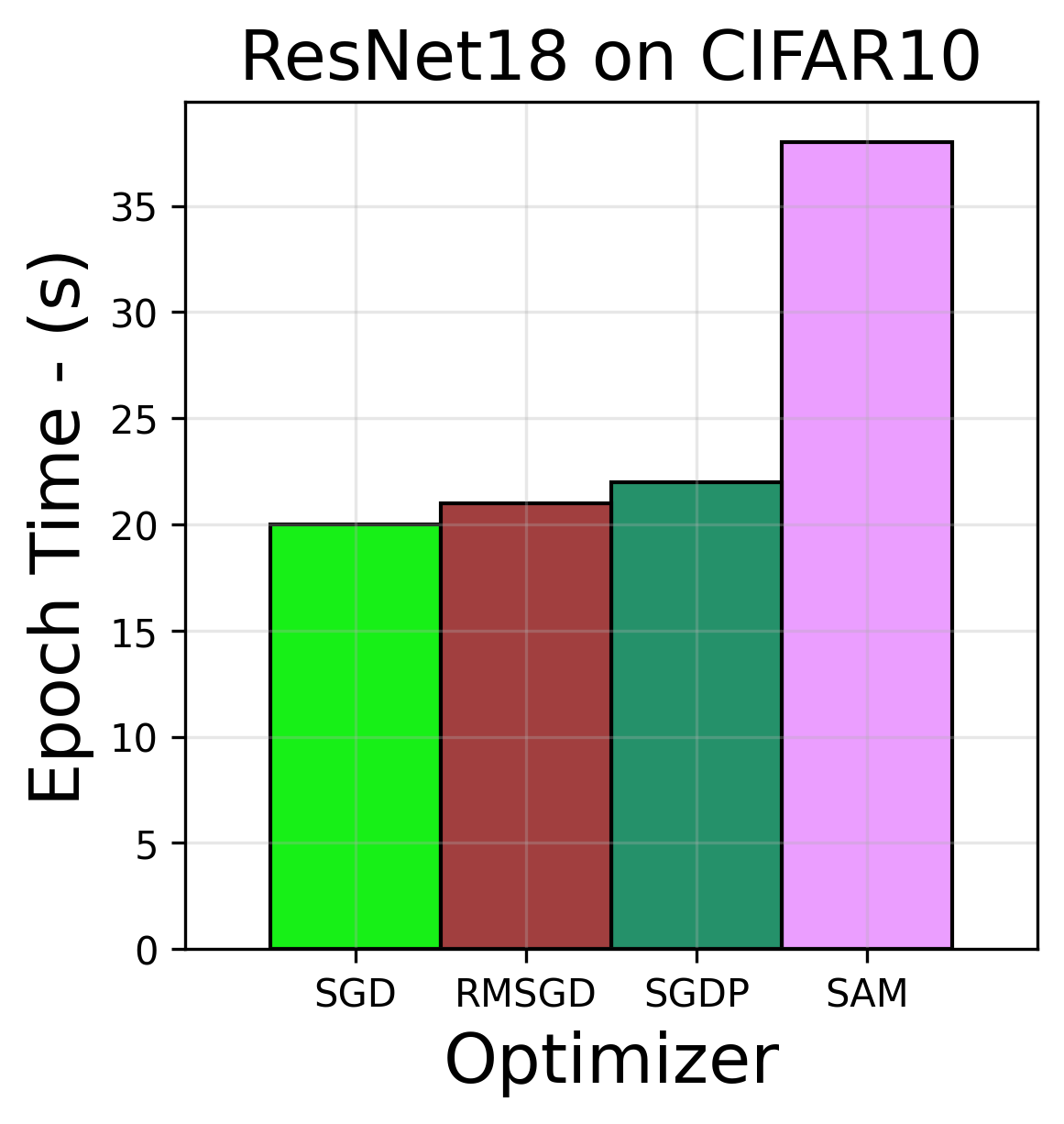}
\includegraphics[width=0.23\textwidth]{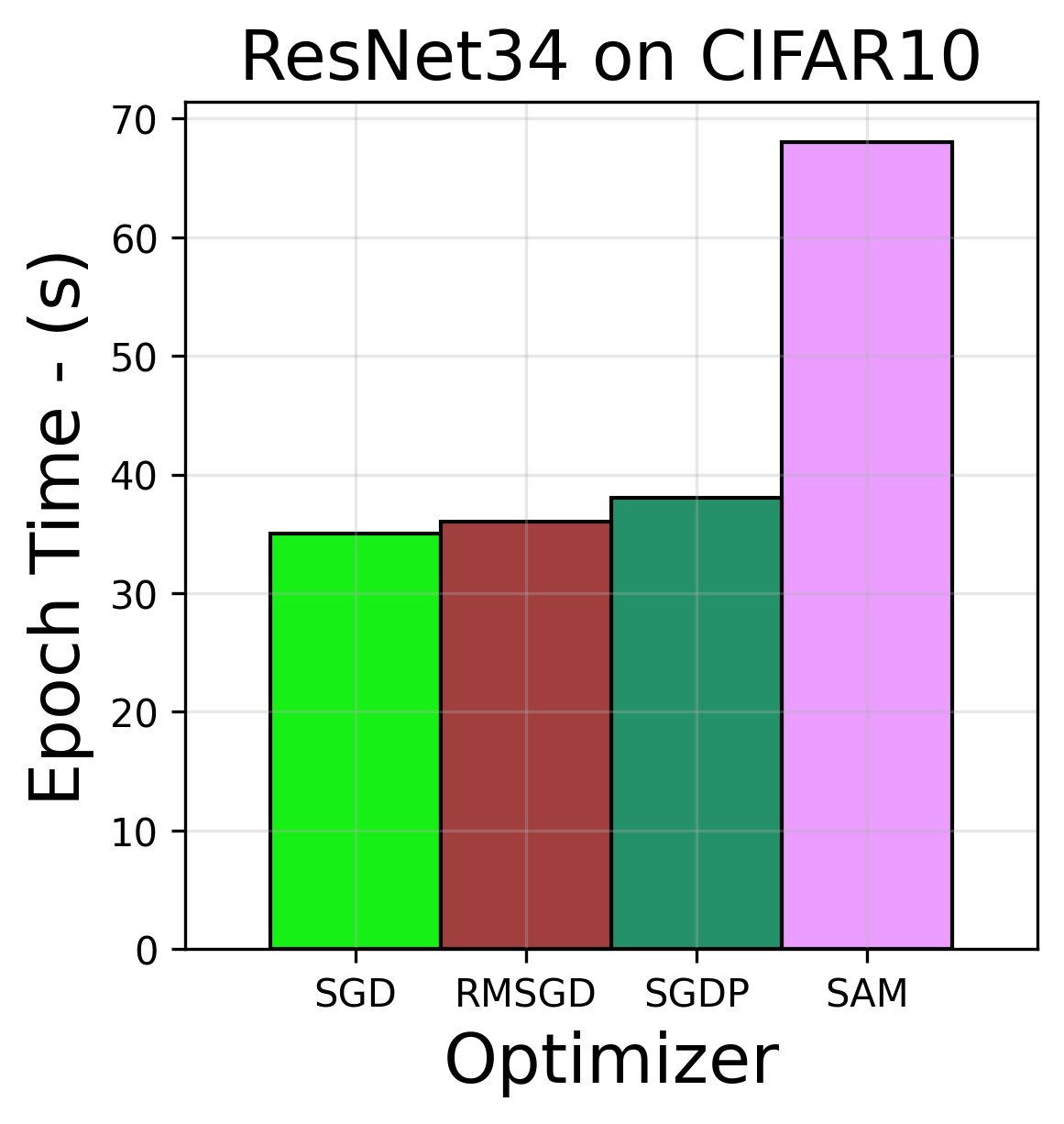}
\includegraphics[width=0.23\textwidth]{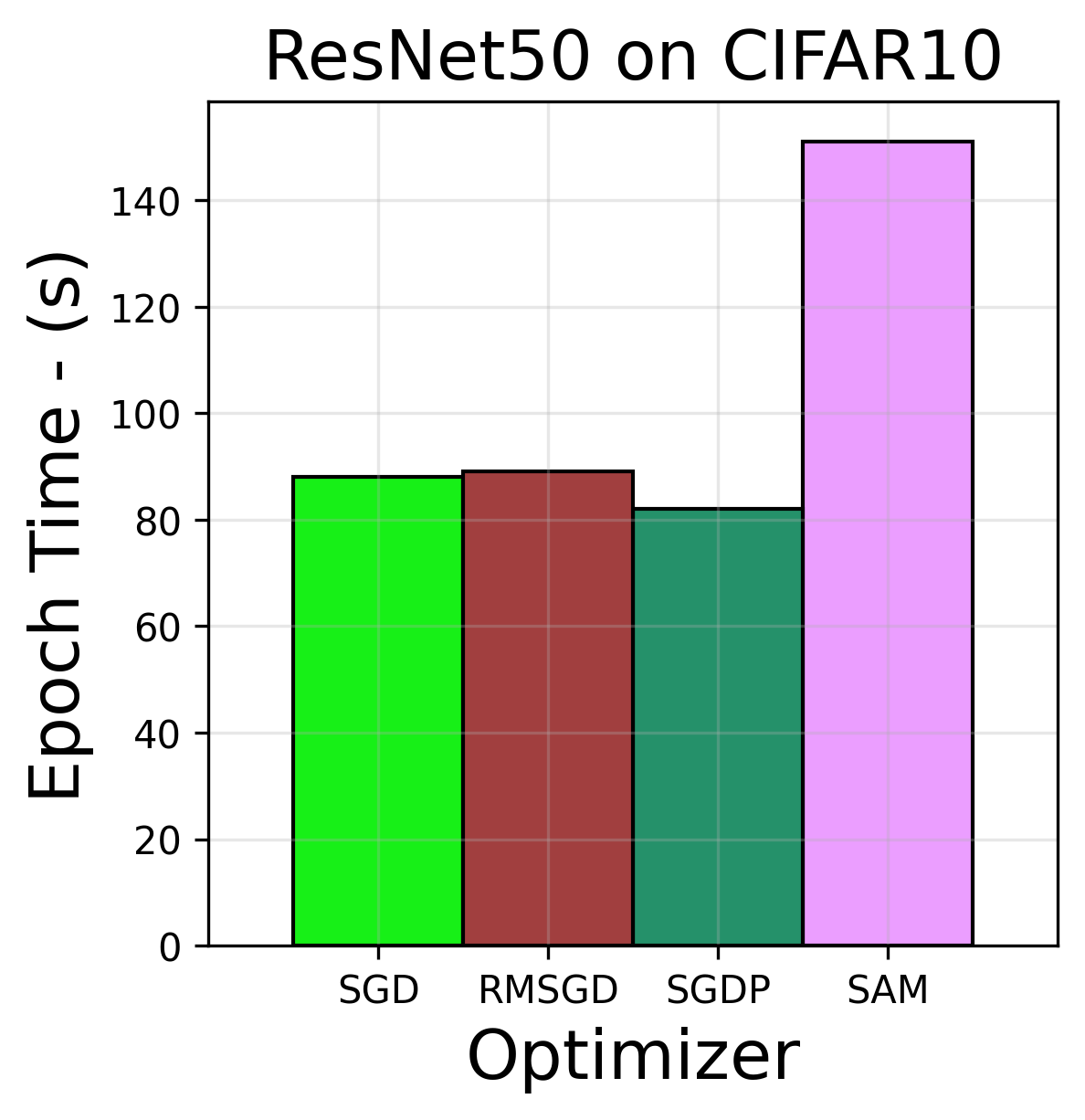}\\

\includegraphics[width=0.23\textwidth]{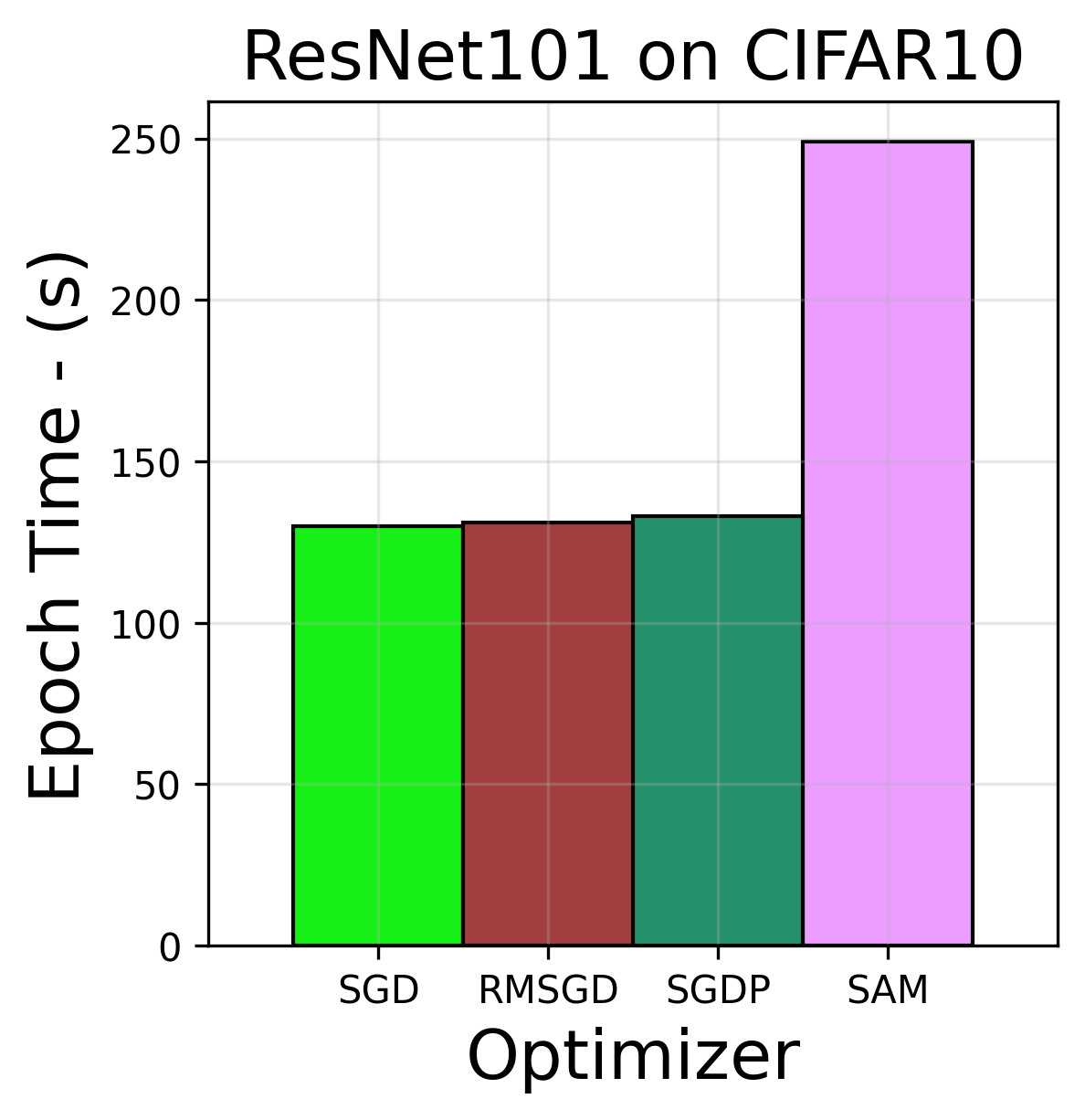}
\includegraphics[width=0.23\textwidth]{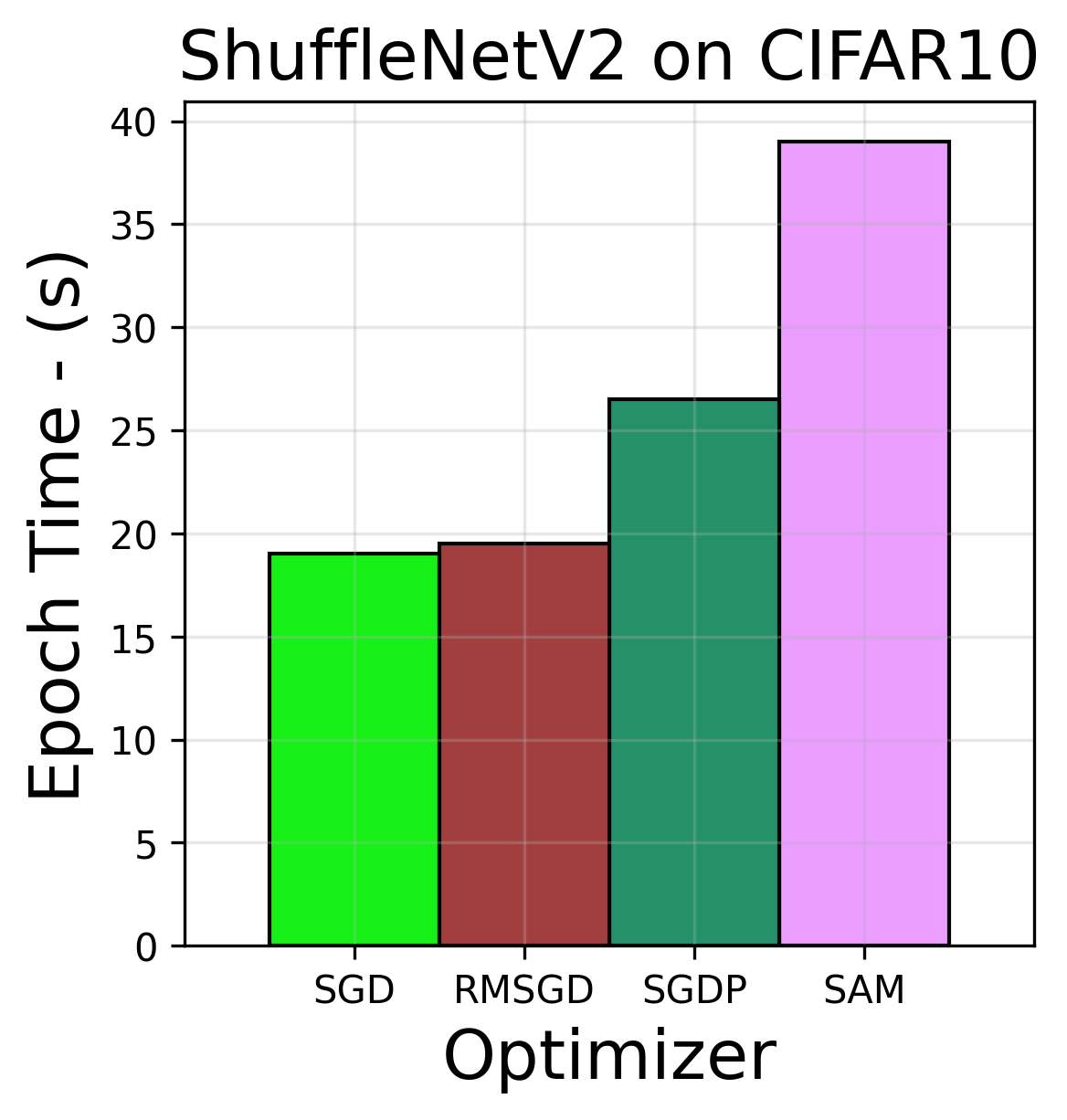}
\includegraphics[width=0.23\textwidth]{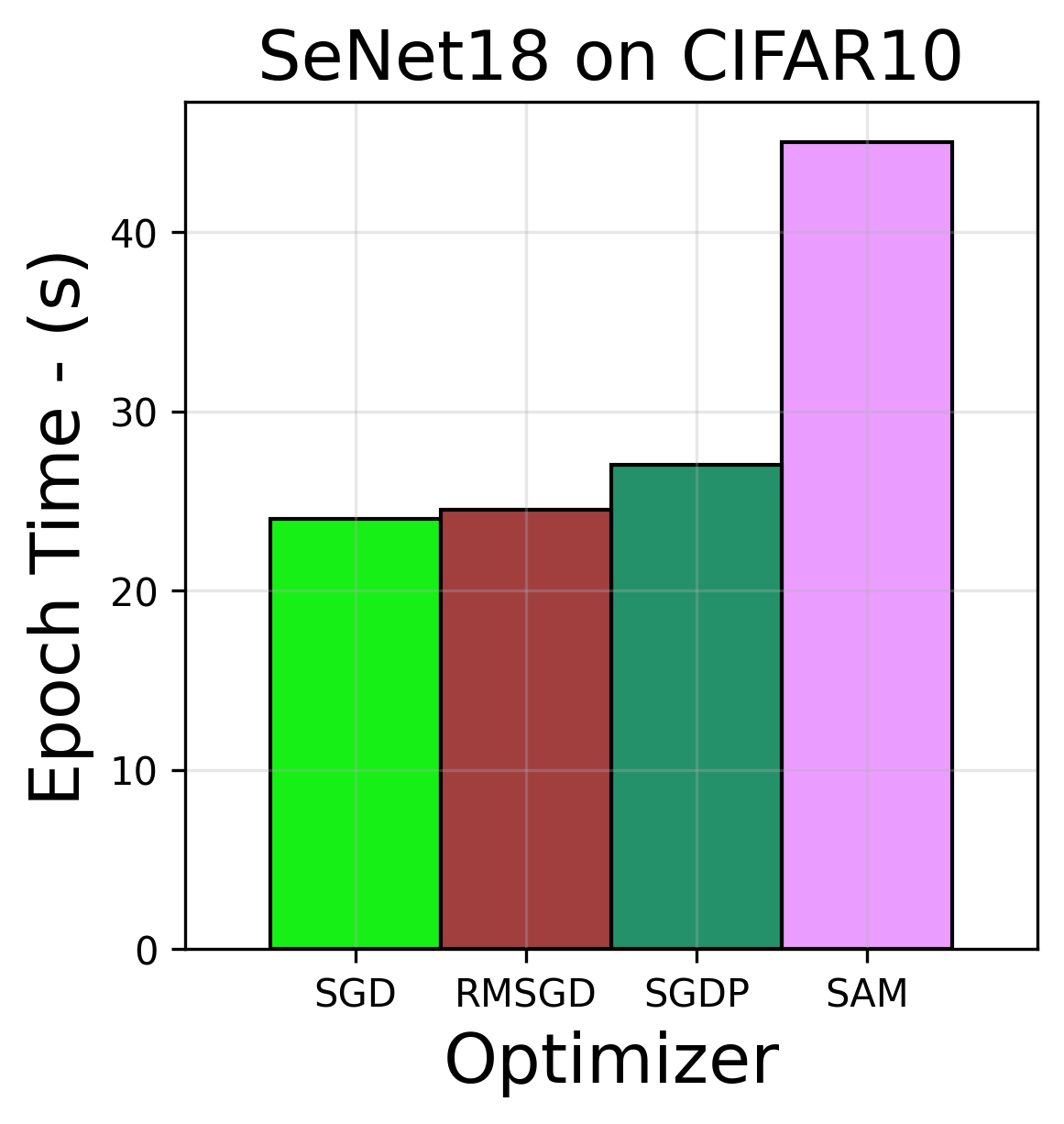}\\

\includegraphics[width=0.23\textwidth]{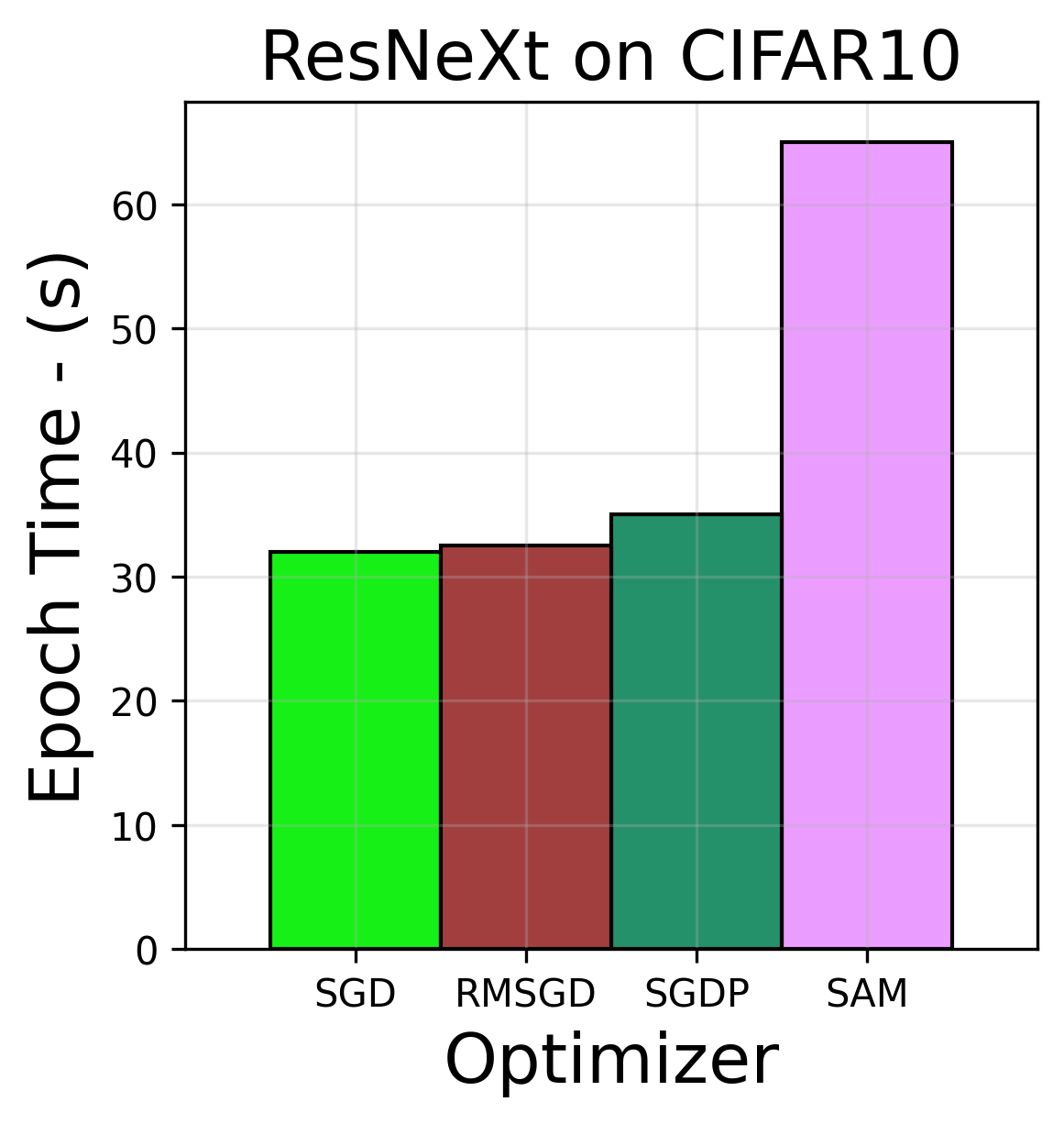}
\includegraphics[width=0.23\textwidth]{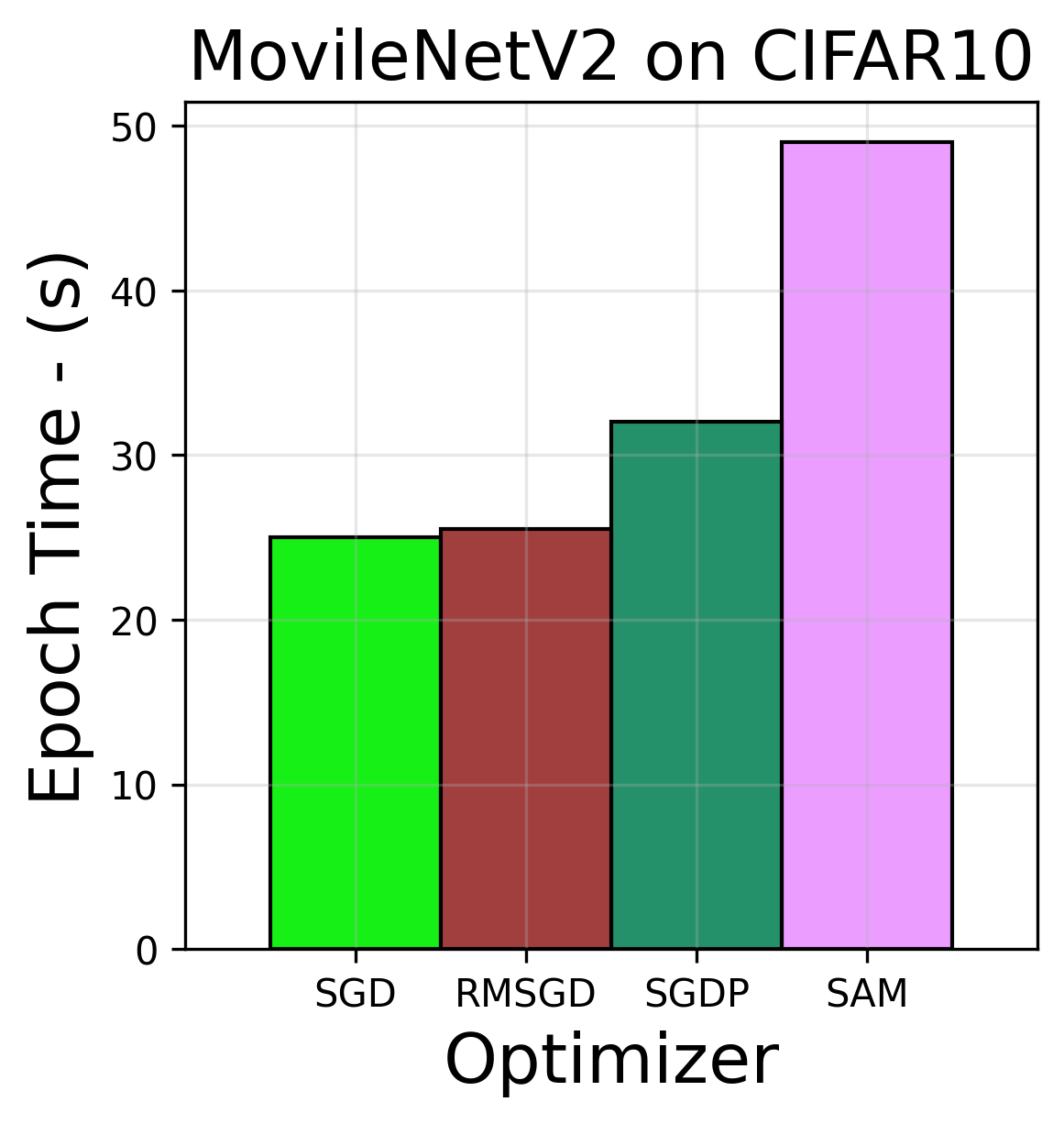}
\includegraphics[width=0.23\textwidth]{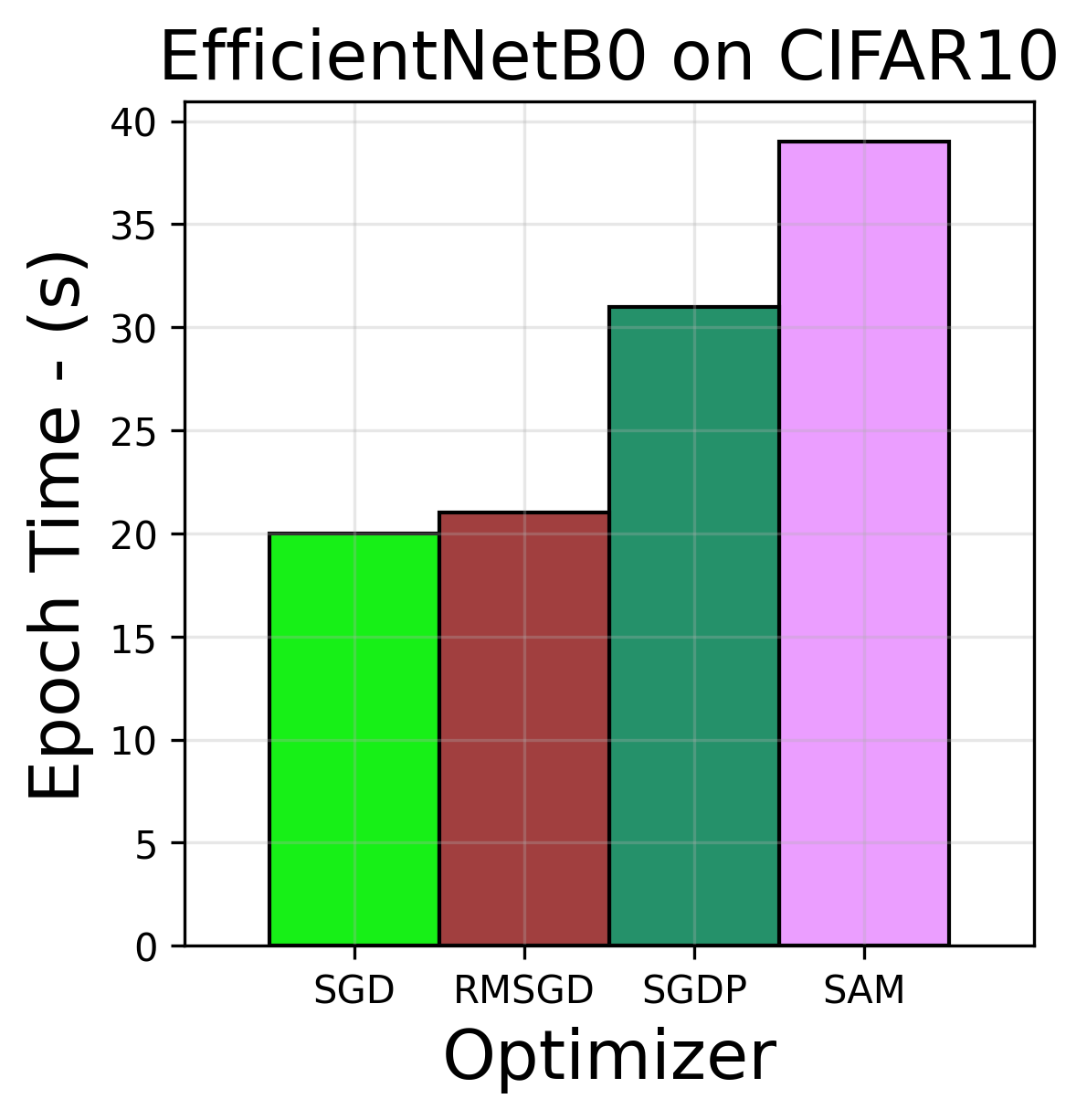}
}
    \caption{Epoch times for various networks on CIFAR10 using SDG, RMSGD, SGDP, and SAM.}
    \label{fig:epoch_times_c10}
\end{figure}
\begin{figure}[h]
\centering{
\includegraphics[width=0.23\textwidth]{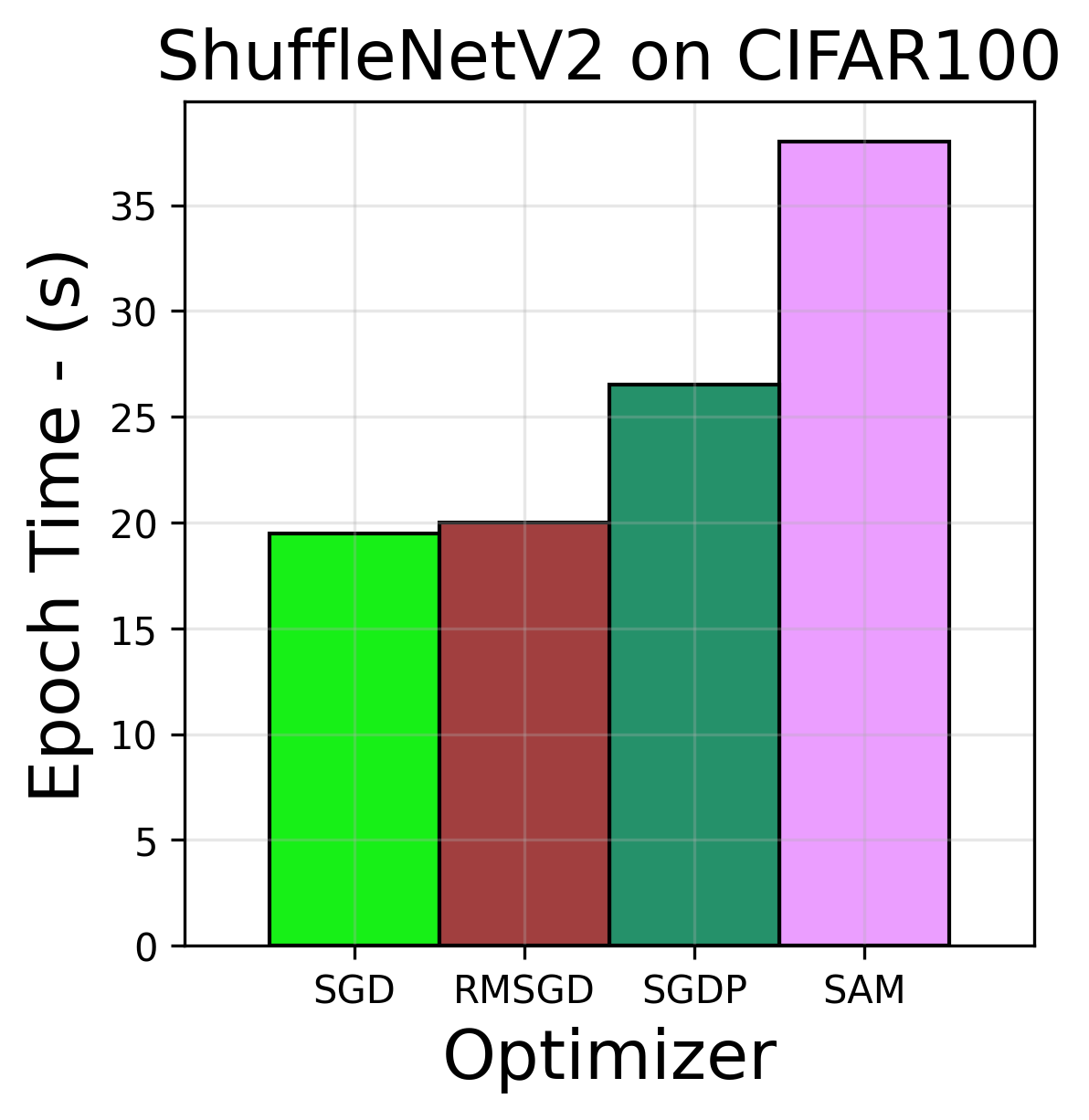}
\includegraphics[width=0.23\textwidth]{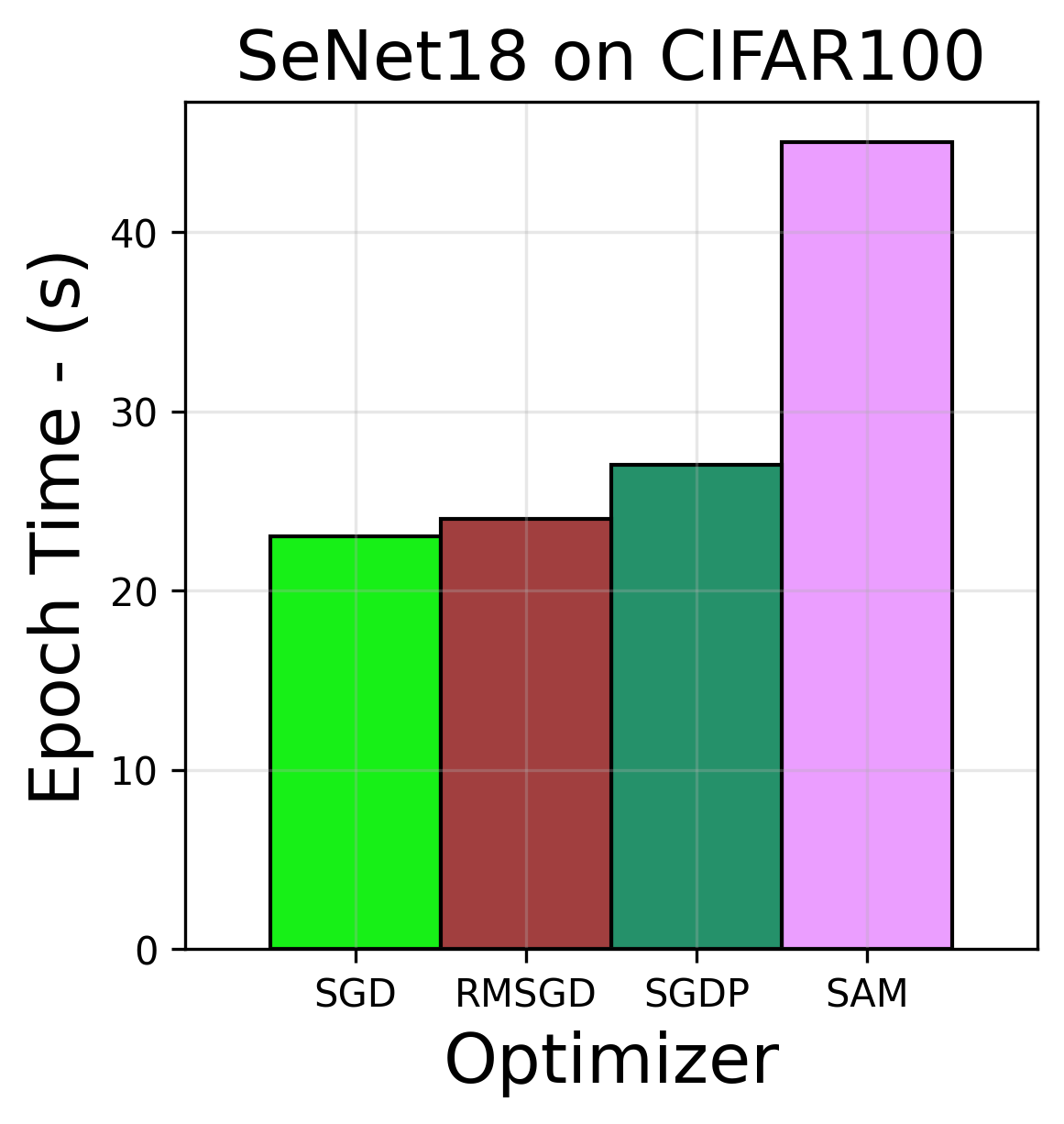}
\includegraphics[width=0.23\textwidth]{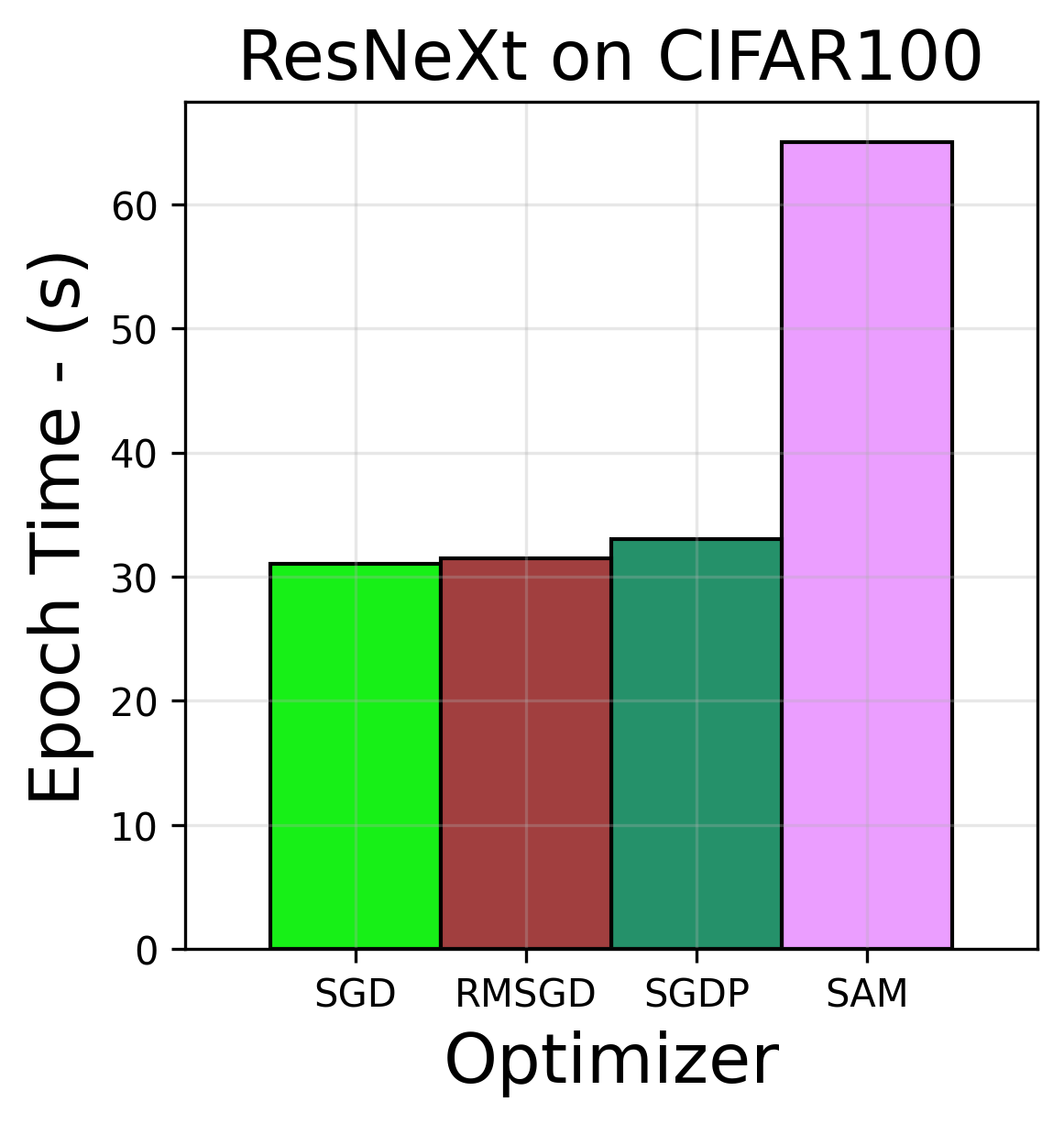}\\

\includegraphics[width=0.23\textwidth]{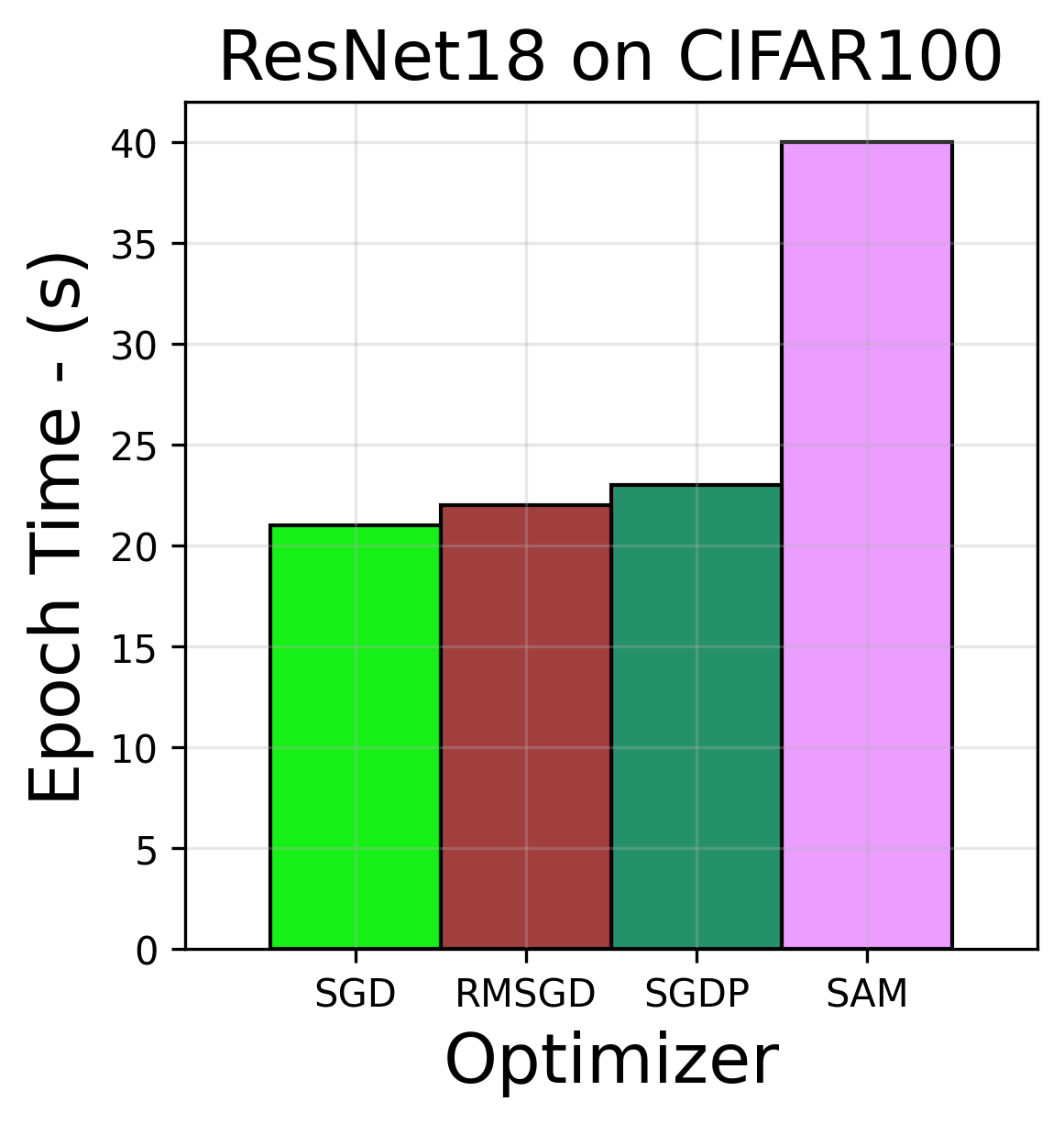}
\includegraphics[width=0.23\textwidth]{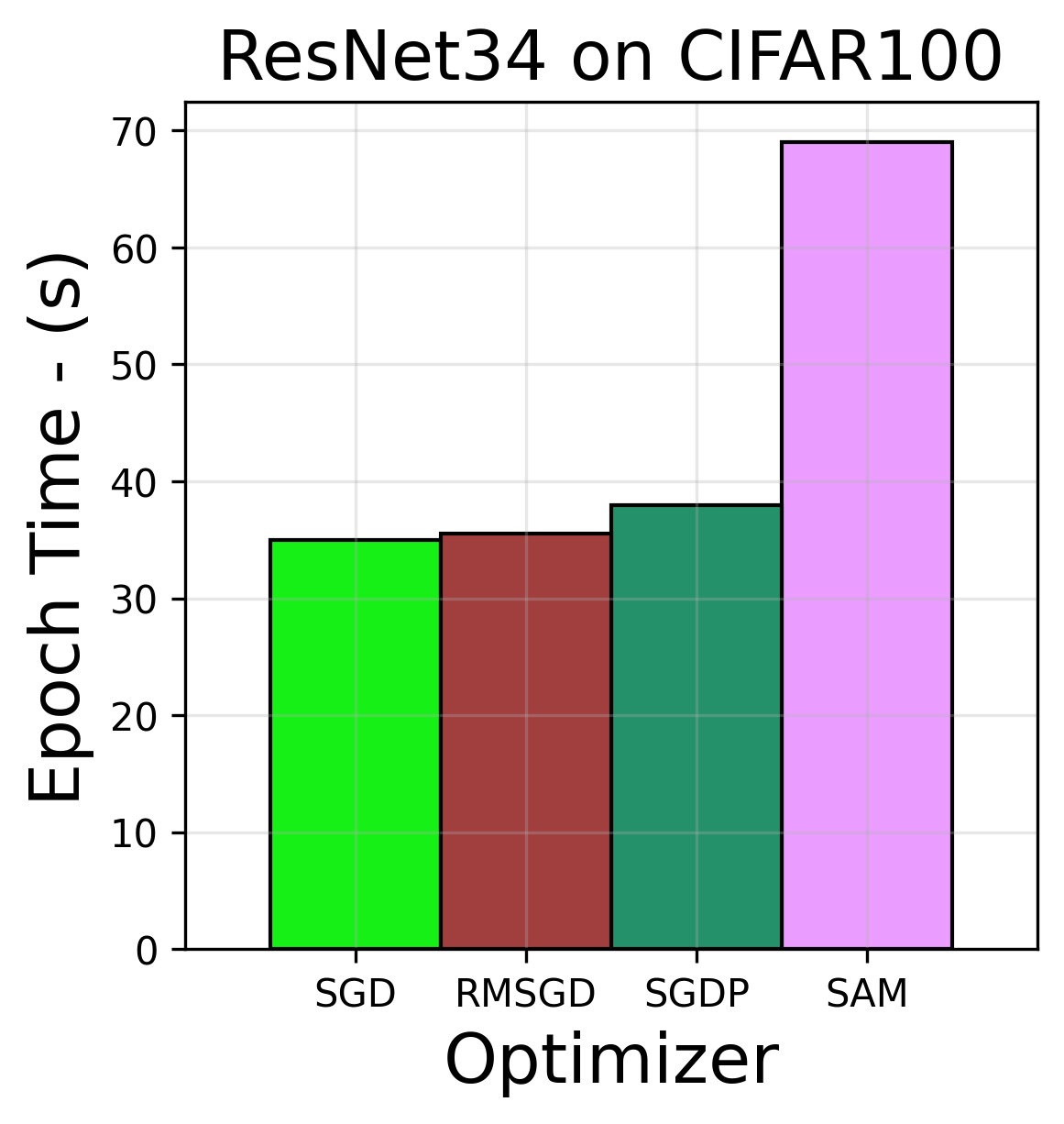}
\includegraphics[width=0.23\textwidth]{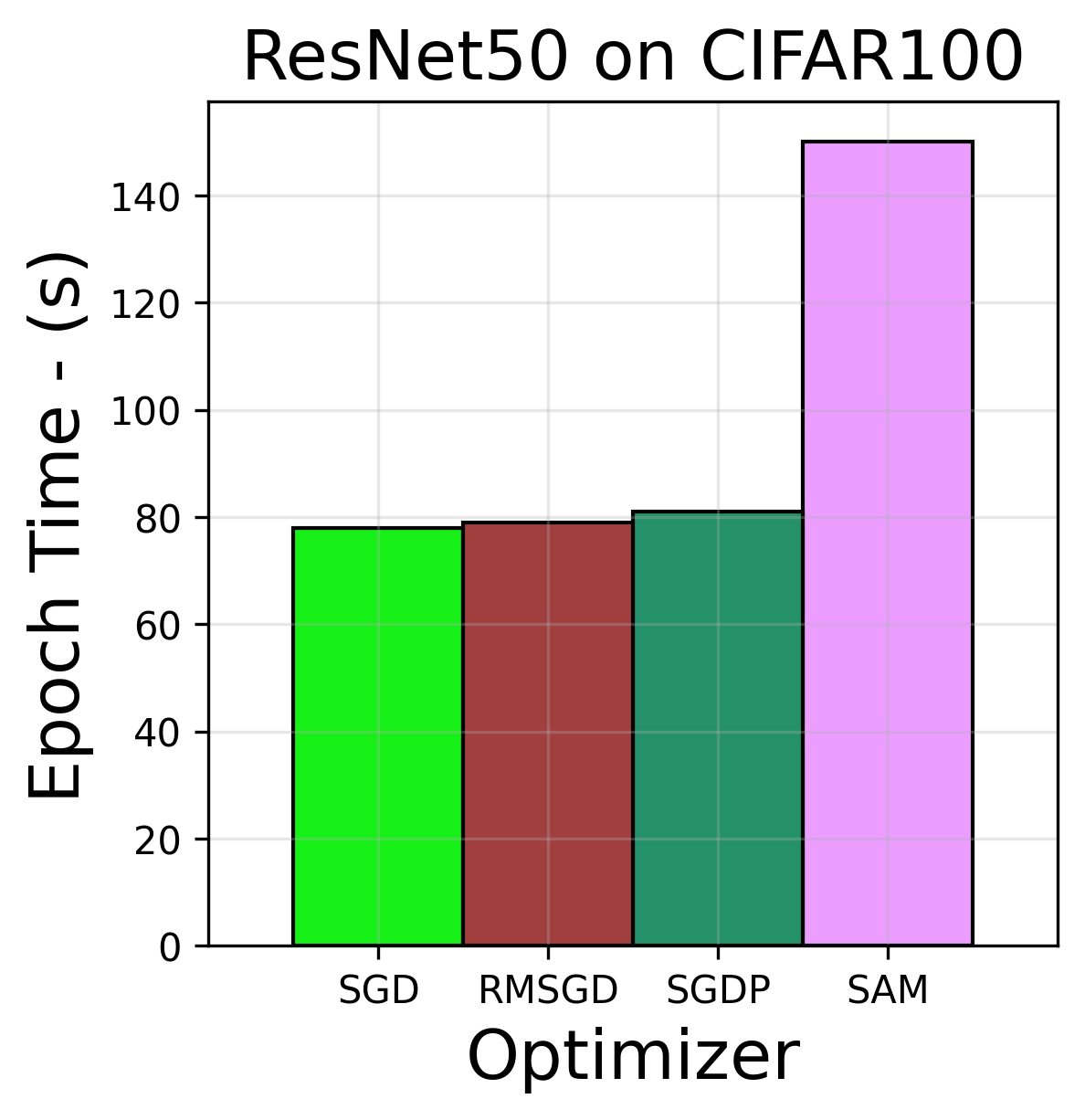}\\

\includegraphics[width=0.23\textwidth]{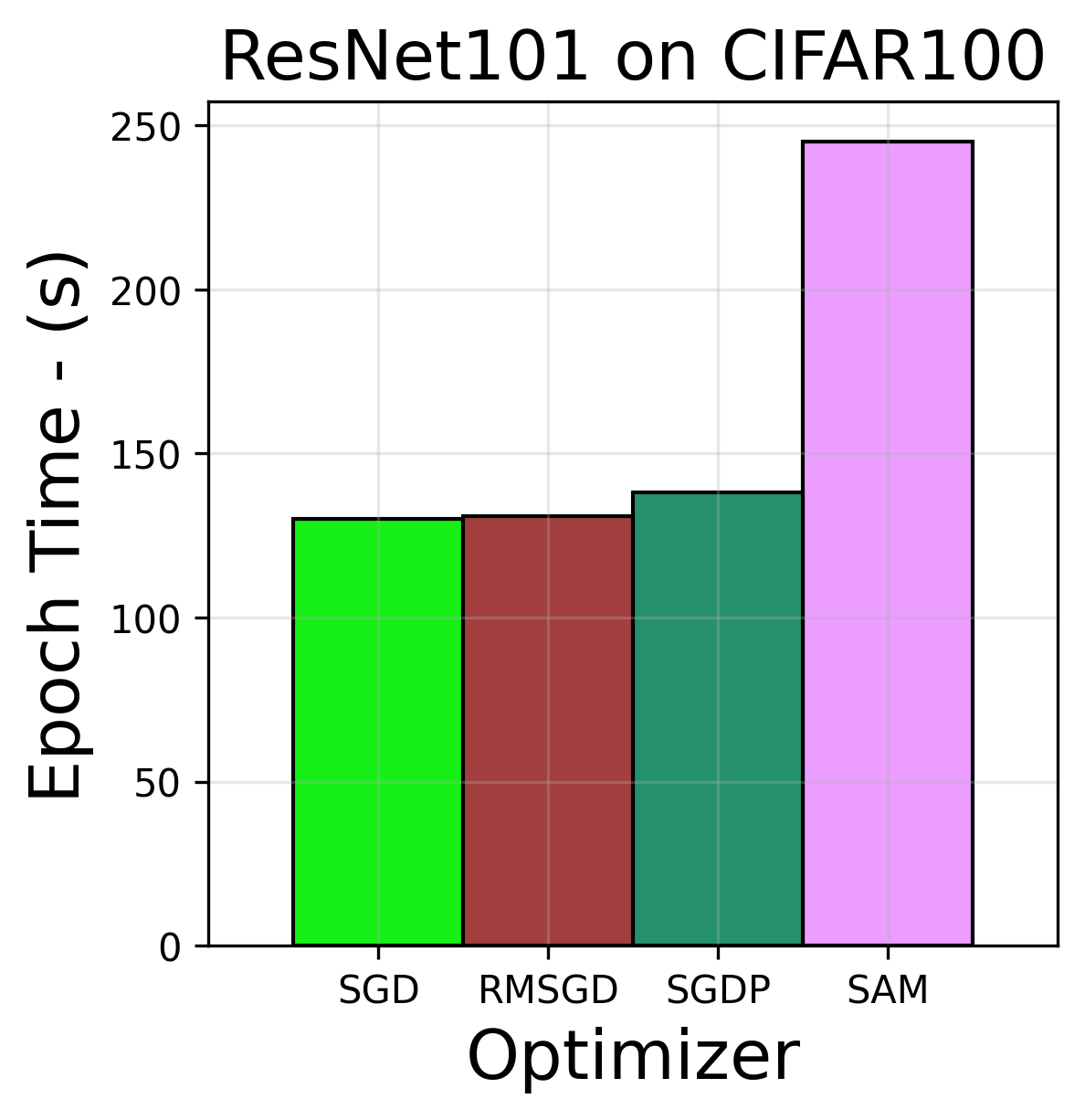}
\includegraphics[width=0.23\textwidth]{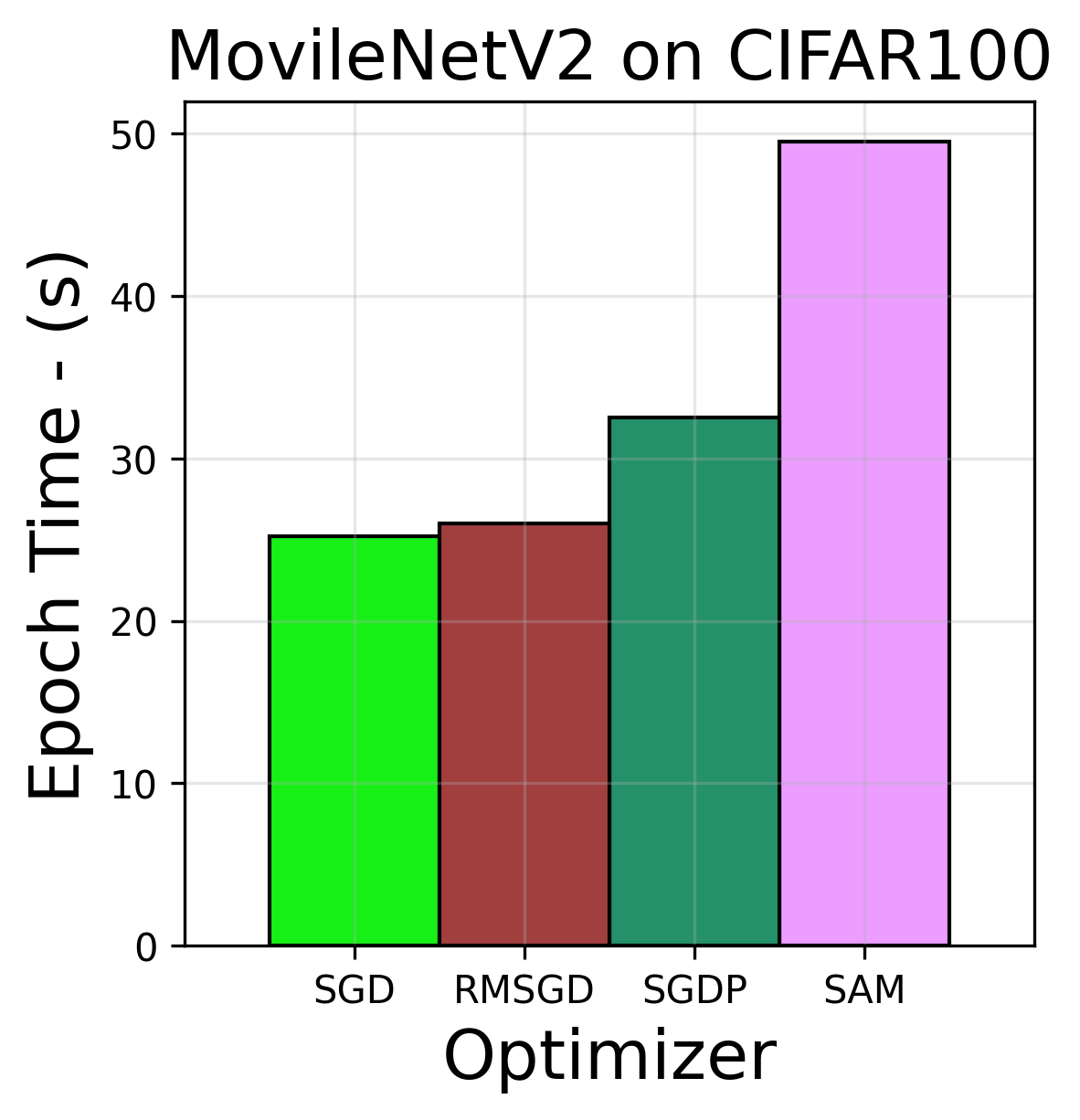}
\includegraphics[width=0.23\textwidth]{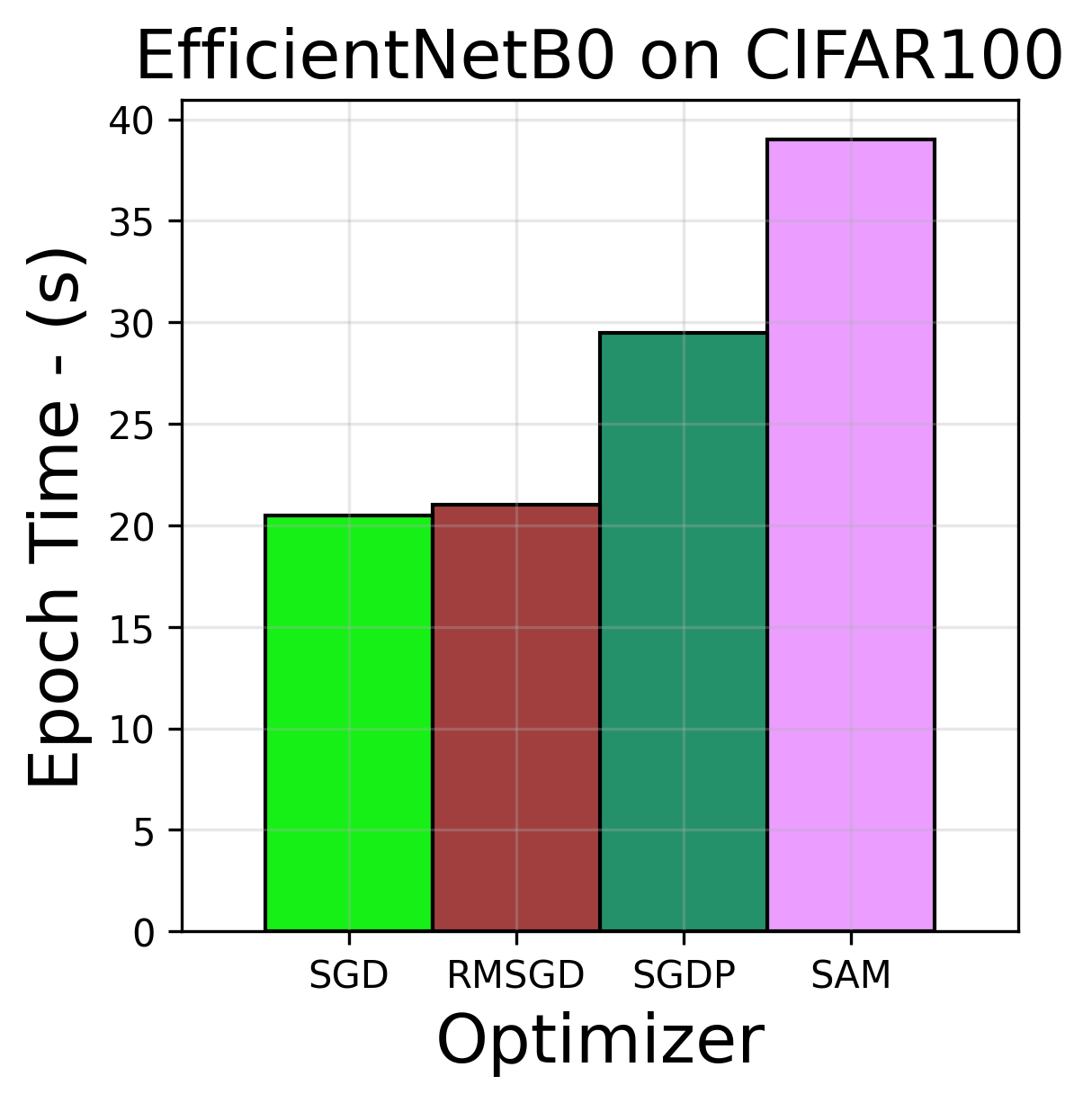}
    }
    \caption{Epoch times for various networks on  CIFAR100 using SDG, RMSGD, SGDP, and SAM.}
    \label{fig:epoch_times_c100}
\end{figure}

\clearpage
\subsection{Language Modelling}
\label{app_exp_lm}
\subsubsection{General Notes}
The Penn TreeBank dataset is a word-level language modelling dataset. We first perform comparisons to Adam, SGD, AdamW, RAdam, and AdaBelief on the $3$-layer LSTM, and then only Adam, SGD, and AdaBelief on the $1$- and $2$-layer LSTM as these optimizers proved most competitive. The model architecture is a 3-layer LSTM. The emebdding size is $400$ and the number of hidden units per layer is $1150$.
\subsubsection{Hyper-Parameters}
Note we tuned hyper-parameters on the $3$-layer LSTM and then applied the same configuration to the $1$- and $2$-layer LSTM.  For each optimizer, a step decay scheduling of step size $75$ epochs and decay rate $0.1$ was used. Each optimizer had fixed weight decay of $\num{1e-6}$. For SGD and RMSGD, momentum of $0.9$ was used. Note for \AlgName, we used $\beta=0.9$ due to the smaller max epoch number of $200$. A batch size of $80$ was used, with gradient clipping of $0.25$. Each optimizer's learning rate was tuned using a grid search method on the 3-layer LSTM, and the same hyper-parameters were applied to the 1- and 2-layer LSTM. The grids are as follows, with the selected learning rate in bold:
\begin{itemize}
    \item RMSGD: $\{0.1, 2, 5, 10, \mathbf{15}, 30\}$
    \item SGD: $\{0.1, 2, 5, 10, 15, \mathbf{30}\}$
    \item SGDP: $\{0.1, 2, 5, 10, 15, \mathbf{30}\}$
    \item SAM: $N/A$
    \item AdaBelief: $\{0.00005, 0.0001, 0.0003, 0.001, \mathbf{0.01}, 0.1\}$
    \item AdamW: $\{0.00005, 0.0001, 0.0003, \mathbf{0.001}, 0.01 0.1\}$
    \item RAdam: $\{0.00005, 0.0001, 0.0003, \mathbf{0.001}, 0.01 0.1\}$
    \item Adam: $\{0.00005, 0.0001, 0.0003, 0.001, \mathbf{0.01}, 0.1\}$
\end{itemize}

\subsection{Generative Adversarial Network}
\label{app_exp_gan}
As noted in the main draft, we used the popular Wasserstein GAN (WGAN) \cite{arjovsky2017wasserstein}.
\subsubsection{Hyper-Parameters}
All hyper-parameters were left as their default values, except for momentum in SGD, \AlgName, and SGDP, which used a momentum rate of $0.9$. Note for \AlgName, we used $\beta=0.95$ due to the smaller max epoch number of $100$. Learning rate was the only hyper-parameter we tuned, and we performed a learning rate grid search over the full $100$ epochs of training. The learning rate grids are as follows, with the selected learning rate in bold:
\begin{itemize}
    \item RMSGD: $\{0.001, 0.01, 0.03, 0.1, \mathbf{0.5}, 2\}$
    \item SGD: $\{0.001, 0.01, 0.03, \mathbf{0.1}, 0.5, 2\}$
    \item SAM: $\{0.0001, 0.001, 0.01, 0.1, \mathbf{0.5}, 1.0\}$
    \item AdaBelief: $\{0.00005, 0.0001, \mathbf{0.0003}, 0.001, 0.01, 0.1\}$
    \item RAdam: $\{0.00005, 0.0001, \mathbf{0.0003}, 0.001, 0.01, 0.1\}$
    \item Adam: $\{0.00005, 0.0001, \mathbf{0.0003}, 0.001, 0.01, 0.1\}$
    \item RMSProp: $\{0.00005, 0.0001, \mathbf{0.0003}, 0.001, 0.01, 0.1\}$
    \item AdaBound: $\{0.00005, 0.0001, \mathbf{0.0003}, 0.001, 0.01, 0.1\}$
\end{itemize}

\end{document}